# Smart Fashion: A Review of AI Applications in the Fashion & Apparel Industry

Seyed Omid Mohammadi*

University of Tehran, College of Engineering, S.OmidMohammadi@alumni.ut.ac.ir

Ahmad Kalhor

University of Tehran, College of Engineering, AKalhor@ut.ac.ir

The fashion industry is on the verge of an unprecedented change. The implementation of machine learning, computer vision, and artificial intelligence (AI) in fashion applications is opening lots of new opportunities for this industry. This paper provides a comprehensive survey on this matter, categorizing more than 580 related articles into 22 well-defined fashion-related tasks. Such structured task-based multi-label classification of fashion research articles provides researchers with explicit research directions and facilitates their access to the related studies, improving the visibility of studies simultaneously. For each task, a time chart is provided to analyze the progress through the years. Furthermore, we provide a list of 86 public fashion datasets accompanied by a list of suggested applications and additional information for each.

## 1 INTRODUCTION

Artificial intelligence brings many benefits to the fashion industry's retailers and customers alike [1]. That is why more and more studies are dedicated to AI applications in the fashion industry every year, and AI will soon reshape this industry into smart fashion. As studies in this field proliferate, more branches and leaves are added to this enormous tree. It is a vast hierarchy that sometimes makes it hard to spot some novel ideas and deprive them of well-deserved attention. That is why, unlike previous review articles, we try to include as many examples as possible and not only state-of-the-art methods. Hopefully, this will increase the visibility of studies in each area, leading to better and more accurate future contributions.

* Corresponding author

Multiple survey and review articles cover the newest developments in smart fashion. We can separate these studies into two groups. Most of them belong to group one, which are application-based surveys focusing on a single application covering state-of-the-art and novel methods for that specific application. We cover these studies separately, each in their appropriate category. Survey articles in the second group have a broader focus, covering different applications. In 2014, [2] provided a short study of three components of the styling task. In 2018, [3] talked about computational fashion and the collision of fashion and multimedia technologies, providing a list of companies in the fashion industry and the applications they are currently working on. In 2019, [4] also reviewed previous research studies focusing on three groups of fashion applications, plus datasets and industry applications. In 2020, there are [5], a bibliometric survey, and [6], a fantastic review of the state-of-the-art methods in each application. Finally, the latest work is a comprehensive survey in 2021, which includes 232 significant studies in 4 main topics and 12 sub-categories [7].

Our focus is not only on significant works in the field but also on covering any relevant contribution. This way, we bring attention to possible unseen potentials, and also we can analyze the progress of smart fashion through the years with a broader range. We choose articles published in 2010-2020 (with some earlier/later exceptions), which leads to a massive number of 587 relevant studies in total. We categorized all these articles into multiple application classes and sub-classes with a multi-label scheme, meaning that one piece might contribute to various applications. These categories are shown in Figure 1. We assign each article to an application category only if it explicitly reports relevant results for that application.

The main contributions of our article are as follows:

- We provide a survey of AI applications in the fashion and apparel industry, and the scope of our work is more than twice the size of the most comprehensive study to date.
- We introduce more than 22 applications and list all relevant studies for each application separately with a multi-label scheme.
- We list 86 public fashion datasets along with the structural information and a list of suggested applications for each dataset. It is the most comprehensive public fashion dataset list to our knowledge, and we believe that it can help many researchers in the future as a quick reference.
- Every application category comes with a time chart of 2010-2020 articles. Thus, it helps analyze the progress speed of research in each category separately.
- We also provide a co-occurrence table for categories that summarizes how these applications are related to one another.

Sec. 2 reviews articles in each application category. It includes ten main categories and a total of 22 applications, as shown in Figure 1, each with a short introduction along with a list of relevant articles and a time chart showing the popularity trends and the progression of each application through the years. Sec. 3 summarizes public fashion datasets. In Sec. 4, we discuss further details of the future of AI in the fashion industry and draw our conclusions in Sec. 5.

## 2 APPLICATIONS

Here, we follow an application-based grouping of articles. Following the taxonomy of [3], [6] grouped these applications into three classes: 1) Low-Level fashion recognition, 2) Mid-Level fashion understanding, and 3) High-Level fashion applications. The categorization we provide here is based on the main focus of each study. Thus, bear in mind that there exist overlaps between these categories. Higher-level applications might consist of mid-level or multiple low-level applications, e.g., try-on applications might also cover parsing, labeling, classification, detection, etc. Each application comes with a summarized table of articles. Due to space limitations, we introduce the articles in a single-line format using the first author's name, publication date, technical keywords, results (wherever possible). These technical keywords try to summarize used methods and are not the same as the article's keywords. They provide rich, compact and simplified information about each article. Additionally, we use "Application Notes" to add a short but straightforward application detail to each article.

The nature of tables requires us to use the abbreviated form of words, including accuracy (Acc), precision (Prec), recall (Rec), mean (m), True Positive (TP), Human Studies/Score (HS), and other common technical words. Researchers should be aware when consulting these tables that different studies experiment on various datasets under different circumstances. We also use specific terms to talk about multiple fashion image types; Figure 2 introduces some examples of these terms. "Item" and "Title" refer to professional catalog images of one fashion article with a white or neutral background, while "Model" refers to a full/half-body image of a model wearing a single or several fashion items under standard conditions. "Shop" images are professional images with a neutral background and might be "Item," "Model," or a combination of both. "Street" images are out-of-the-studio good quality pictures usually focused on one professional model. They have more sophisticated backgrounds, different lighting conditions, and minor occlusion due to various yet standard poses. "Wild" photos, on the other hand, have no constraints at all. They are user-created amateur versions of Street photos, sometimes with heavy occlusion, bad lighting, cropping, and poor overall quality.

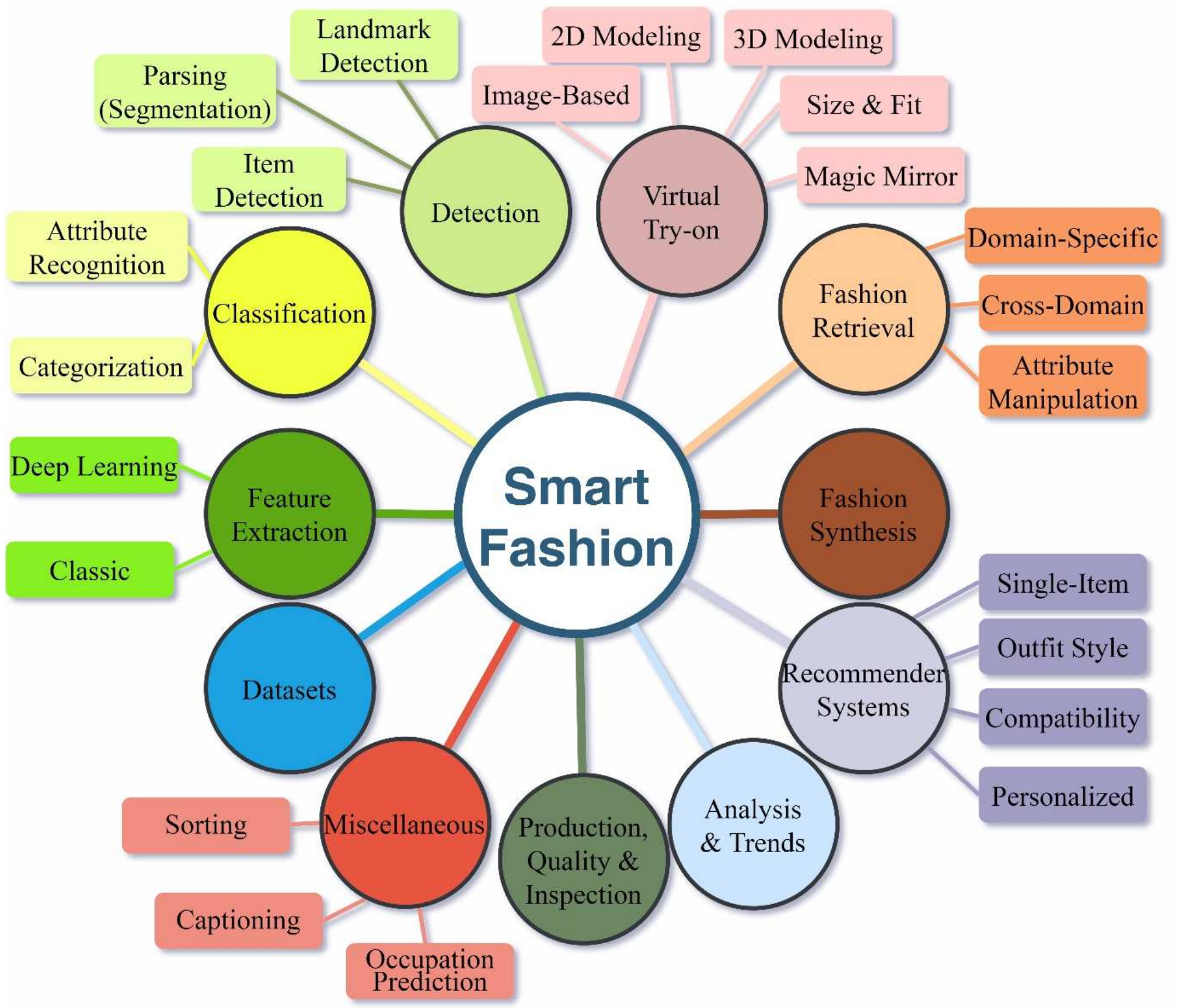

Figure 1: Diagram of smart fashion categories in this research.

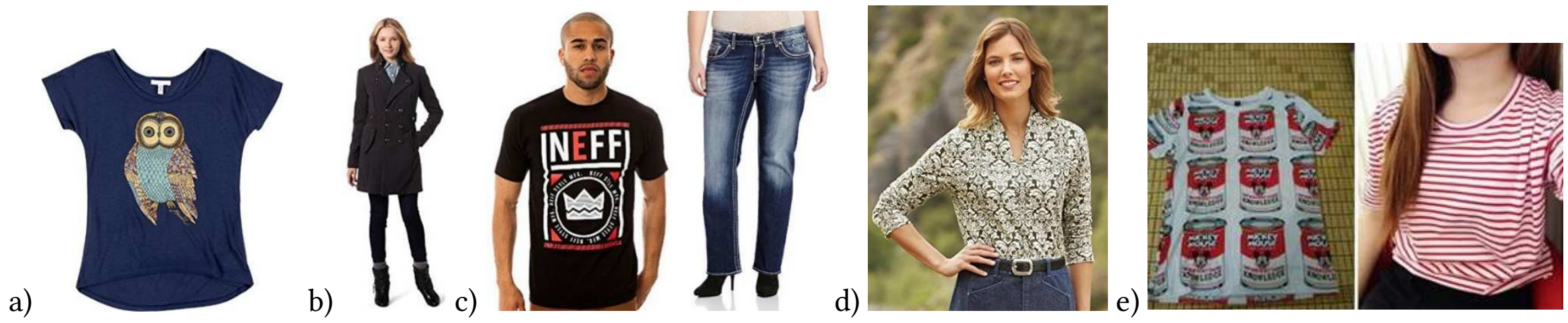

Figure 2: Examples of different types of fashion-related images from Amazon [8] and Deepfashion [9] datasets. a) Item/Title b) Full-body Model c) Half-body Model d) Street e) Wild

## 2.1 Feature Extraction

Feature extraction has the goal of learning the mathematical representation of fashion items. It is a low-level but fundamental task. As it is used in many applications, especially for item similarity, we only report some examples in this section. Feature extraction is done either through classic methods and hand-crafted features or deep learning methods.

### *2.1.1 Classic Methods*

Classic feature extraction methods and image processing techniques were widely used before the rise of artificial neural networks. Examples of these methods are Color Histograms, Local Binary Patterns (LBP), Histogram of Oriented Gradients (HOG), Scale-Invariant Feature Transformation (SIFT), and many more.

We can use any of these methods or a combination of them. For example, in 2009 [10] used LBP, HOG, and color histogram for its smart mirror fashion recommender. Yang et al. [11] used a combination of HOG, SIFT, DCT, and color histogram for clothing recognition in surveillance videos. We can also use these image processing techniques on images to preprocess them before using neural networks, e.g. [12] in 2019, applied Haar-Cascade and Difference of Gaussian (DoG) on image inputs of their inception based deep CNN to build a recommender.

### *2.1.2 Deep Learning Methods*

With the emergence of artificial neural networks and deep convolutional neural networks, researchers shifted their focus to these networks for representation learning tasks. These networks soon replaced the high-effort task of feature engineering. Deep learning methods also allow us to learn fine-grained features. Some examples are: Fashion DNA [13], Fashion Style in 128 single-precision floats [14], Style2Vec [15], and [16] utilizing weakly annotated fashion images.

## 2.2 Classification

Classification is the task of systemically arranging items into groups. We further break this task into 1) Categorization and 2) Attribute Recognition. The terminology might seem a little confusing, but we need to separate these two. Although they might have some overlaps, they are two different problems with different levels of complexity.

### *2.2.1 Categorization*

This article uses the term categorization to imply a form of classification based on a shared set of qualities and rules. Categorization is a subjective grouping of fashion items. This task focuses on predicting only the main category of a fashion item (shirt, dress, pants, etc.). As each item can only exist in one class in a set of categories, this task is a single-label prediction most of the time.

Table 1: Articles Related to Categorization

| No | Article Reference | Year | Technical Keywords/Claimed Results | Application Notes |
|---|---|---|---|---|
| 1 | Yang [11] | 2011 | Linear SVM, HOG, BOW, DCT, 80% Rec | Surveillance videos clothing recognition |
| 2 | Hidayati [17] | 2012 | Classic, F-score of 92.25% on >1000 images | Genre classification, Style elements |
| 3 | Willimon [18] | 2013 | Classic, L-C-S-H, Mid-level layers, 90% TP | Laundry items classification, 3 Categories |
| 4 | Kalantidis [19] | 2013 | SIFT, LBP, Multi-Probe LSH Index, 54% Avg Prec | For recommendation task |
| 5 | Kiapour [20] | 2014 | Classic, Between/Within-Class, above 70% Acc | Style classification of a whole outfit |
| 6 | Tong Xiao [21] | 2015 | CNN, Probabilistic graphical model | Noisy labels |
| 7 | Lao [22] | 2015 | AlexNet, 50.2% Acc for clothing style | Style recognition |
| 8 | Vittayakorn [23] | 2015 | Semantic parse, SVM, KNN, 27.8/57.8/12.9% Acc | Year/Season/Brand classification |
| 9 | Yamazaki [24] | 2015 | Classic, Gabor Filters, Feature description | Clothes sorting, Bundled clothing |
| 10 | Surakarin [25] | 2015 | Classic, LDP, SURF, Bag of features, SVM | Seven categories of clothing |
| 11 | Z. Liu [9] | 2016 | VGG-16, FashionNet, Landmark, 82.58% Top-3 Acc | Benchmark, DeepFashion |
| 12 | Patki [26] | 2016 | New architecture better than VGG16, 41.1% Acc | Street photos |
| 13 | Arora [27] | 2016 | Modified VGG16, 92% and 71% Acc | Catalogue and street photos |
| 14 | R. Li [28] | 2016 | ELM, AE-ELM, Feature fusion, MLP | Efficiency and time comparison with MLP |
| 15 | Simo-Serra [14] | 2016 | CNN, 128 Floats, Triplet ranking, VGG-16, 61.5% IOU | Style Class., Street, Weak data, Features |
| 16 | Sun [29] | 2017 | SVM, LBP, SI, TSD, and BSP features | Clothes sorting, Bundled, Single-Shot |
| 17 | Bhatnagar [30] | 2017 | CNN, 92.54% Acc | On Fashion-MNIST |
| 18 | Qian [31] | 2017 | Seg., ASPP, CRF, Faster R-CNN, VGG-16, 88.9% mAcc | Pattern classification, Street photos |
| 19 | Chen [32] | 2017 | CNN, Distributed computing, 59% Acc | Multiple architectures & datasets |
| 20 | Corbiere [33] | 2017 | ResNet50, Bag-of-words, 86.30% Top-3 Acc | Weakly Annotated Data |
| 21 | Inoue [34] | 2017 | CNN, Multi-task label cleaning, 64.62% mAP | Multiple items |
| 22 | Dong [35] | 2017 | Multi-task curriculum transfer, 65.96% Prec | Street photos, Detection |
| 23 | X. Zhang [36] | 2017 | CNN, Alexnet, Avg. AUC of 81.2% on 3 datasets | Multiple items category detection |
| 24 | Lee [15] | 2017 | CNN, Style2Vec, VGG, 61.13% Acc | Style Class., Representation learning |
| 25 | Takagi [37] | 2017 | CNN, VGG, Xception, Inception, ResNet50, 72% mAcc | Style Classification, Street |
| 26 | Gu [38] | 2017 | QuadNet, SVM, 65.37/42.80% Acc, 49.92% Prec | Season/Style/Garment, Street |
| 27 | Veit [39] | 2017 | Conditional similarity, CNN, Triplet, 53.67% Acc@1 | Brand classification, Similarity learning |
| 28 | Bedeli [40] | 2018 | AlexNet, 75.3% Acc in Surveillance data | Forensics, Surveillance camera, Logos |
| 29 | Verma [41] | 2018 | StyleNet, CNN, Attention, ST-LSTM, 68.38% mAP | Multiple items, On Fashion144K |
| 30 | Zhang [42] | 2018 | Graph-based DCNN, CNN, VGG-16, ~85% mAcc | Style recognition |
| 31 | Dong [43] | 2018 | VGG-Net, Spatial pyramid pooling, 76.78% Acc | Style recognition |
| 32 | Schindler [44] | 2018 | CNN, comparing five networks, VGG16 | Item, Gender classification, Person Detect. |
| 33 | Kuang [45] | 2018 | Hierarchical deep learning, Avg 85.63% Acc | Hierarchical classification |
| 34 | T. Nawaz [46] | 2018 | CNN, RmsProp, 89.22% Acc | Traditional clothing |
| 35 | Wazarkar [47] | 2018 | Linear convolution, Matching points, 71.4% TP | On Fashion 10K dataset |
| 36 | Bhatnagar [48] | 2018 | Compact bilinear CNN, 84.97% Top-3 Acc | Weak annotations |
| 37 | K-Gorripati [49] | 2018 | CNN, VGG16, 83% Acc | For recommendation task |
| 38 | Hidayati [50] | 2018 | Classic, SVM, Face detection, 88.40% mean F | 16 clothing genres for upper/under-wear |
| 39 | Wang [51] | 2018 | VGG16, Fashion grammar, BCRNN, 90.99% Top-3 Acc | Landmark-driven attention and detection |
| 40 | Ye [52] | 2019 | Hard-Aware BackPropagation, GAN, 90.93% Top-3 Acc | Insufficient training data |
| 41 | F. Wang [53] | 2019 | CNN, Region Proposal Strategy, 91.7% Acc | Cashmere/Wool, Textile |
| 42 | Seo [54] | 2019 | Hierarchical CNN, VGG-Net, 93.33% Acc | On Fashion-MNIST |
| 43 | P. Li [55] | 2019 | Two-stream multi-task network, 93.01% Top-3 Acc | Landmark-driven |
| 44 | Madulid [56] | 2019 | CNN, Inception, 96.2% Acc, 0.981 Rec, 1 Prec | Seven categories of clothing |
| 45 | Umaashankar [57] | 2019 | Benchmark, ResNet34, 0.92 micro f-score | Benchmark, Atlas |

| No | Article Reference | Year | Technical Keywords/Claimed Results | Application Notes |
|---|---|---|---|---|
| 46 | Guo [58] | 2019 | CNN, Incpetion-BN, 88.2% Top-3 Acc | Benchmark, iMaterialist |
| 47 | J. Liu [59] | 2019 | Feature map upsampling, 91.16% Top-3 Acc | Landmark-Aware attention |
| 48 | Asiroglu [12] | 2019 | CNN, Inception, Haar-cascade, DoG, 86%/86%/98% Acc | Color/Gender/Pattern classification |
| 49 | Tuinhof [60] | 2019 | CNN, AlexNet, BN-Inception, 87%/80% Acc | Category/Texture, For recommendation |
| 50 | Tran [61] | 2019 | YOLO, Resnet18, 75.66% mAcc | Street images, 33 classes, For retrieval task |
| 51 | Park [62] | 2019 | CNN, SEResNeXt50, ~88.42% Top-3 Acc | Benchmark, Multiple methods |
| 52 | Stan [63] | 2019 | CNN, AlexNet, 83% Acc | For recommendation task |
| 53 | Ma [64] | 2019 | Bi-LSTM, ResNet-18, 47.88% Occ. & 73.95% Cat. Acc | Social media photos, Occasion & Category |
| 54 | Hidayati [65] | 2019 | Local features, Skin color, SURF, SVM, 73.15% F1 | Genre classification, Street images |
| 55 | Alotaibi [66] | 2020 | Autoencoder, DeepAutoDNN, 93.4% Acc | On Fashion-MNIST |
| 56 | M. Nasir [67] | 2020 | CNN, Custom 17-layer, 97.9% Acc | Comic superheroes classification |
| 57 | Verma [68] | 2020 | Faster RCNN, MobileNet MTL, 73.5% Acc | For cold-start problem in recommenders |
| 58 | Shajini [69] | 2020 | Attentive CNN, VGG-16, 91.02% Top-3 Acc | Landmark-Driven |
| 59 | Z. Wang [70] | 2020 | CNN, Noise attention, A2NL, 6.4% test error | Attention-aware noisy label learning |
| 60 | J. Liu [71] | 2020 | Random Forest, VGG-IE, 93.97% Acc | On Fashion-MNIST |
| 61 | Rame [72] | 2020 | VGG-16, Attention, Illumination correction. | Color regression, Main and multiple colors |
| 62 | Ziegler [73] | 2020 | Category aware attention, 78.63% Top-3 Acc | Clothes sorting, Robotics, In-lab images |
| 63 | Jain [74] | 2020 | Data Mining, Compare 3 Methods, 86% Acc | Benchmark, Data mining techniques |
| 64 | Truong [75] | 2020 | ResNet, Soft attention, 98.55% Acc | Relation (what worn by who?) |
| 65 | Iqbal Hussain [76] | 2020 | ResNet-50, VGG-16, Rotation, 99.30% Acc | Fabric weave classification, Texture |
| 66 | Shubathra [77] | 2020 | Acc of 90.4% MLP, 93.3% CNN, 97.1% ELM | On Fashion-MNIST, Benchmark |
| 67 | Fengzi [78] | 2020 | VGG-16, InceptionV3, 88.6% Acc for article type | Master/Sub categories, Gender |
| 68 | Y. Zhang [79] | 2020 | TS-FashionNet, 89.94% Top-3 Acc | Landmark-Aware attention |
| 69 | Tian [80] | 2021 | Faster RCNN, Multi-grained Branches | Category grouping |

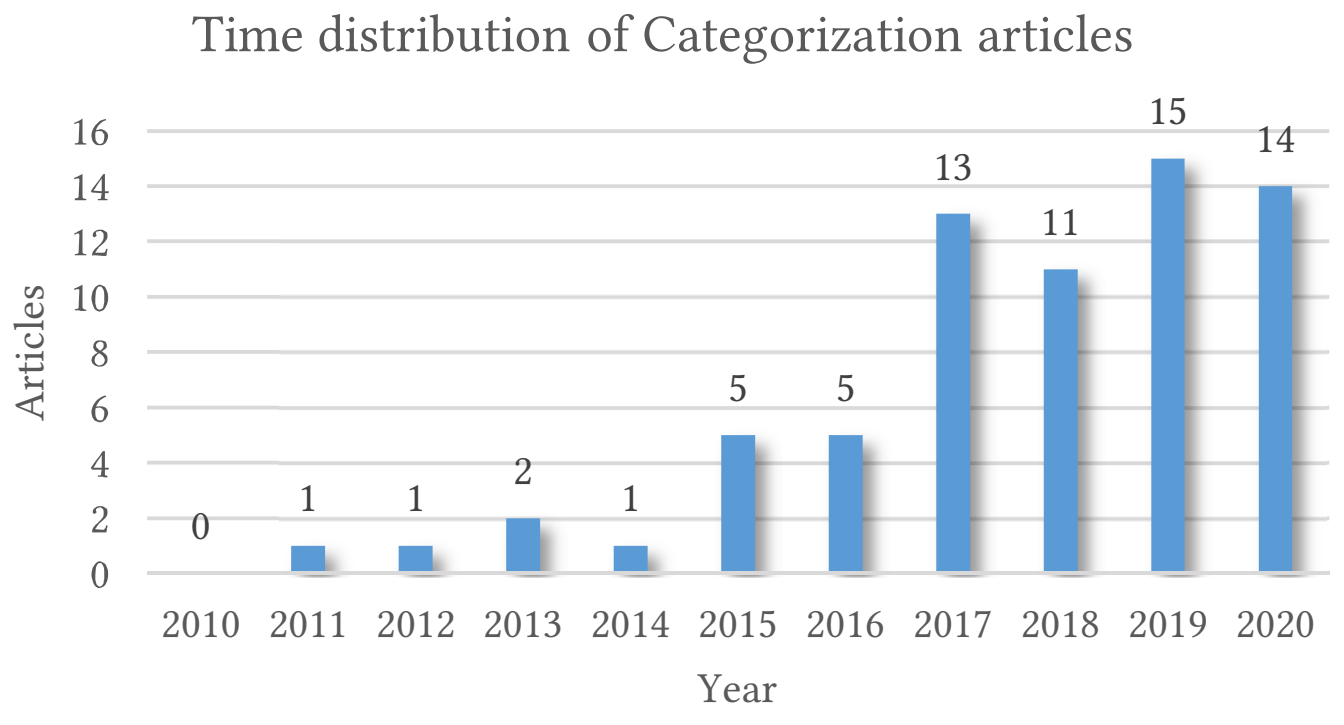

Figure 3: Time Analysis of Categorization Articles

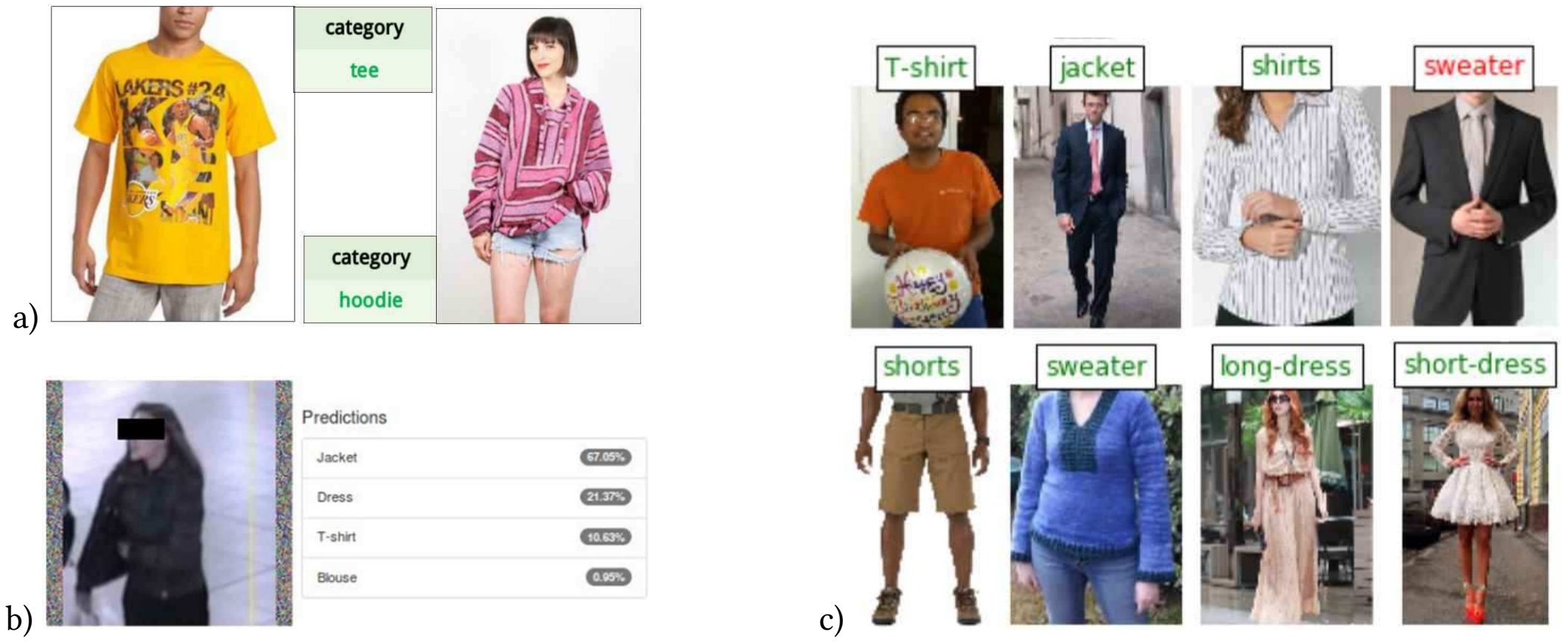

Figure 4: Categorization using a) Shop images [79] b) Surveillance camera footage [40] c) Street images [27]

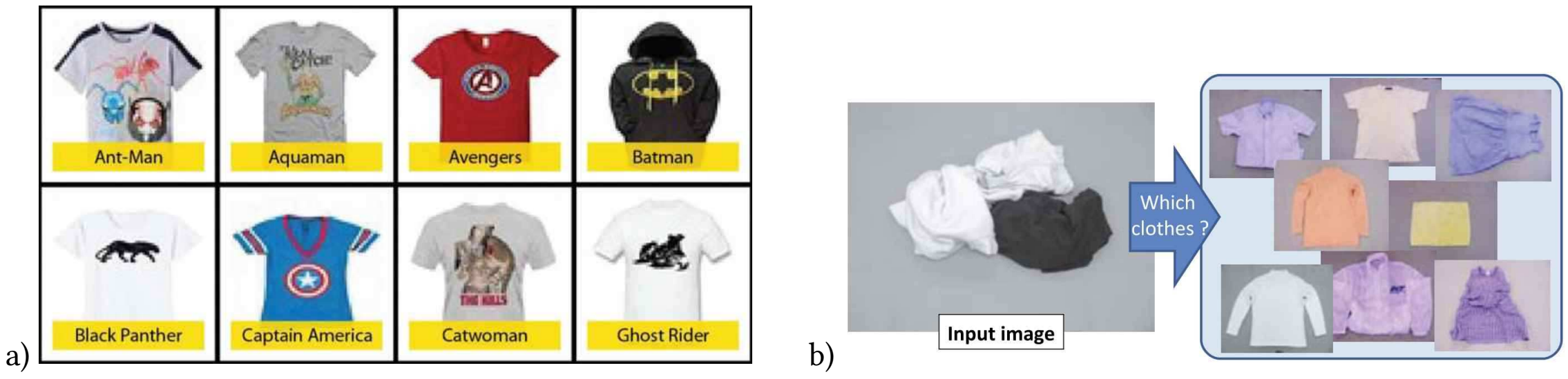

Figure 5: Categorization of a) Comic superheroes [67] and b) Bundled clothing [24]

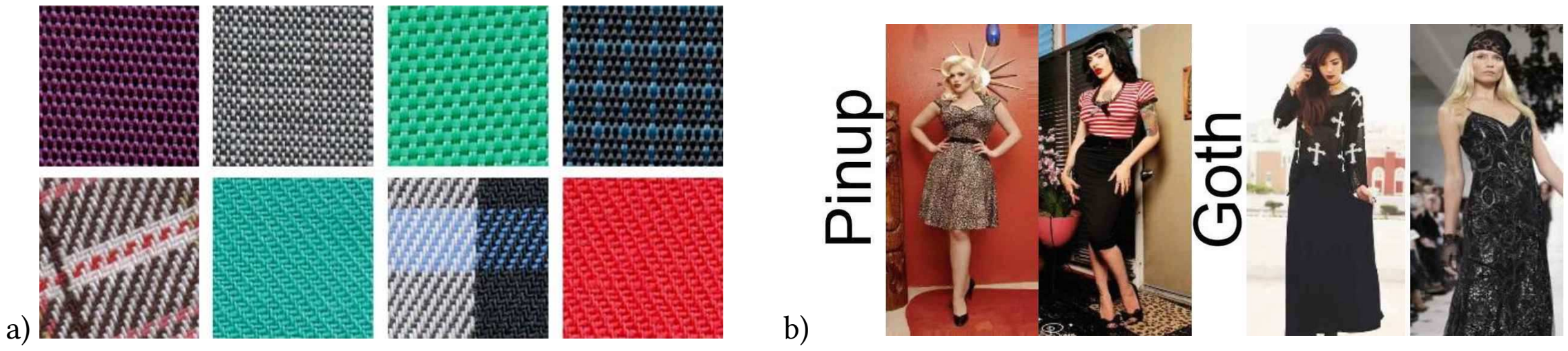

Figure 6: Categorization of a) Woven fabric [76] and b) Style [20]

### *2.2.2 Attribute Recognition*

Attributes are each item's characteristics and objective qualities. Each item can have multiple attributes; that is why it is usually a multi-label task. For example, a dress (category) can have color, pattern, material, price, texture, style, etc., as attributes. Attribute recognition is a broader task than categorization; thus, it might predict the category as well.

Table 2: Articles Related to Attribute Recognition

| No | Article Reference | Year | Technical Keywords/Claimed Results | Application Notes |
|---|---|---|---|---|
| 1 | Bourdev [81] | 2011 | Classic, HOG, SVM, Poselets, Avg 65.18% Prec | Street photos, different viewpoints |
| 2 | Chen [82] | 2012 | Classic, Pose-adaptive, CRF, four combined features | Street photos, Pose estimation, Gender |
| 3 | S. Liu [83] | 2012 | Classic, SVM, HOG, LBP, Color Histogram | Magic closet |
| 4 | Di [84] | 2013 | Classic, SIFT, LBP, HOG, GIST, bag-of-words, SVM | Coat and jacket style |
| 5 | Bossard [85] | 2013 | Transfer forest, SURF, HOG, LBP, SSD, Avg 41.38% Acc | Natural scenes |
| 6 | Huang [86] | 2015 | DARN, NIN, MLPConv, SVR | Attribute-aware cross-domain retrieval |
| 7 | Q. Chen [87] | 2015 | Deep domain adaptation, R-CNN, NIN model | Street photos |
| 8 | Lao [22] | 2015 | AlexNet, 74.5% Acc across all labels | For clothing retrieval task |
| 9 | Yamaguchi [88] | 2015 | CRF, Localization, AlexNet, 67.8% F1 | Considers inter-label correlation |
| 10 | X. Chen [89] | 2015 | Classic, Latent SVM, HOG, DPM detector | In-Lab & Internet photos, Kinect |
| 11 | K. Chen [90] | 2015 | Classic, SIFT, Pose estimation, SVM, 62.6% Acc | Fashion shows & street photos |
| 12 | Z. Liu [9] | 2016 | VGG-16, FashionNet, Landmark, 45.52% Top-3 Acc | Benchmark, DeepFashion |
| 13 | K. Liu [91] | 2016 | VGG-16, Decision fusion, 13.2% labels over 0.8 F1 | View-Invariant, Catalog images, MVC |
| 14 | Patki [26] | 2016 | CNN, ZCA whitening, Avg 84.35% Acc | Street photos |
| 15 | Sha [92] | 2016 | Classic, Color matrix, ULBP, PHOG, Fourier, GIST | For recommendation task |
| 16 | Vaccaro [93] | 2016 | Polylingual topic model, Gibbs sampling, MALLET | Elements of fashion style |
| 17 | Sun [94] | 2016 | Classic features, Pose detection, LBP, PCA | Part-based clothing image annotation |
| 18 | Vittayakorn [95] | 2016 | CNN, Neural activations, KL divergence, 60% mAP | Street images |
| 19 | Z. Li [96] | 2016 | Domain-Adaptive Dict. Learning, K-SVD, PCA | Cross-Domain, Style recognition |
| 20 | R. Li [97] | 2017 | Multi-Weight CNN, Multi-Task, 55.23%mAP | Real-world clothing images |
| 21 | Corbiere [33] | 2017 | ResNet50, bag-of-words, 23.10% Top-3 Acc | Weakly Annotated Data |
| 22 | Dong [35] | 2017 | Multi-Task Curriculum Transfer, 64.35% mAP | Cross-Domain, Street photos, Detection |
| 23 | K. Chen [98] | 2017 | Pose estimation, SIFT, SVM, CRF, 62.6% Acc | Popularity of each attribute |
| 24 | Hsiao [99] | 2017 | Polylingual LDA, Topic model, 53% Avg AP | Unsupervised, Street images |
| 25 | Ly [100] | 2018 | Multi-task learning, LMTL-IDPS, 54.70% Avg Rec | Inner-group correlations, Imbalanced data |
| 26 | Liao [101] | 2018 | EI tree, BLSTM, ResNet50, bi-directional Ranking Loss | Interpretable |
| 27 | S. Zhang [102] | 2018 | Triplet DCNN, fast R-CNN, VGG-16, Avg 87.93% mAP | Video (Fashion shows) |
| 28 | Lee [103] | 2018 | PAFs, SIFT, LBP, HSV, CNN, SVM | CNN and SVM comparison |
| 29 | Zheng [104] | 2018 | Polygon-RNN++, ResNet-50, Xception-65, 45.0% F1 | Benchmark, ModaNet, Polygon, Color |
| 30 | Zakizadeh [105] | 2018 | Bilinear VGG-16, Pairwise ranking loss | Fine-grained attribute recognition |
| 31 | Deng [106] | 2018 | CNN, Color histogram, LBP, 77.38% Avg Acc | For a recommender application |
| 32 | Cardoso [107] | 2018 | VGG-16, Multi-modal Fusion, RNN, 85.58% Avg Acc | ASOS fashion e-commerce retailer |
| 33 | Hidayati [50] | 2018 | Classic, SVM, Face detection, 94.24% Avg F-score | Full-body images, 12 style elements |
| 34 | W. Wang [51] | 2018 | VGG-16, Fashion grammar, BCRNN, 51.53% Top-3 Acc | Landmark-Driven, Detection |
| 35 | Ye [52] | 2019 | Hard-aware BackPropagation, GAN, 52.82% Top-3 Acc | Insufficient training data |
| 36 | Yang [108] | 2019 | Tree-based models, GBDT, CNN, MLP | Interpretable, For mix-and-match |
| 37 | R. Li [109] | 2019 | Multi-task, Multi-weight, Multi-label, CNN | Imbalance, Benchmark 3 methods |
| 38 | He [110] | 2019 | DenseNet161, Separate networks, 97.72% mAP | 2018 FashionAI Global Challenge |
| 39 | P. Li [55] | 2019 | Two-stream multi-task network, 59.83% Top-3 Acc | Landmark-Driven |
| 40 | Zou [111] | 2019 | AttributeNet, Hierarchy, ~86% Acc | Benchmark, FashionAI |
| 41 | J. Liu [59] | 2019 | Feature map upsampling, 54.69% Top-3 Acc | Landmark-Aware attention |
| 42 | Adhikari [112] | 2019 | ResNet-34, Branch network, 77.58% Avg Acc | Progressive attribute learning |
| 43 | Stan [63] | 2019 | CNN for each Cat., AlexNet, Two-stage, Avg 80.58% Acc | For a recommender system |
| 44 | Ma [64] | 2019 | CNN, Bi-LSTM, ResNet-18, 69.59% Acc | Social media photos |
| 45 | Q-Ferreira [113] | 2019 | OpenPose, VSAM, VGG-16, mean 49.22% Top-3 Acc | Pose-guided attention |
| 46 | S. Zhang [114] | 2020 | CNN, TAN, Resnet101, VGG-16, 69.72% mAvgP | Cross-Domain, Task-aware Attention |
| 47 | X. Liu [115] | 2020 | ResNet50, Landmark, 99.81% Top-5 Acc | MMFashion Toolbox |
| 48 | Chun [116] | 2020 | SAC, Grad-CAM, CNN, 81.02% Avg Acc | Self-attention mechanism |

| No | Article Reference | Year | Technical Keywords/Claimed Results | Application Notes |
|---|---|---|---|---|
| 49 | Verma [68] | 2020 | Faster RCNN, MobileNet MTL, 89.1% Acc | For recommendation task |
| 50 | Shajini [69] | 2020 | Attentive CNN, VGG-16, 51.89% Top-3 Acc | Landmark-Driven |
| 51 | Z. Wang [70] | 2020 | CNN, Noise Attention, A2NL, 34.8% test error | Attention-aware noisy label learning |
| 52 | Yue [117] | 2020 | Design Issue Graphs, DCNN, 75.15% F1 | Style recognition |
| 53 | Su [118] | 2020 | Inception-ResNet-v2, 46.0% AP | For retrieval task |
| 54 | Park [119] | 2020 | Machine learning, Hierarchical classification | Attribute classification system |
| 55 | Xiang [120] | 2020 | RCNN, ResNet-50, L-Softmax, 89.02% Prec | Attributes + Bounding box detection |
| 56 | Y. Zhang [79] | 2020 | TS-FashionNet, Two-Stream, 50.58% Top-3 Acc | Landmark-Aware attention |
| 57 | Shi [121] | 2020 | Faster R-CNN, Segmentation, 75% Acc | Trend Analysis, Fashion show Videos |
| 58 | Mohammadi [122] | 2021 | ResNet50, Shallow net, 44.4% IOU, 73.1% Prec, 48.4% Rec | For recommendation task |

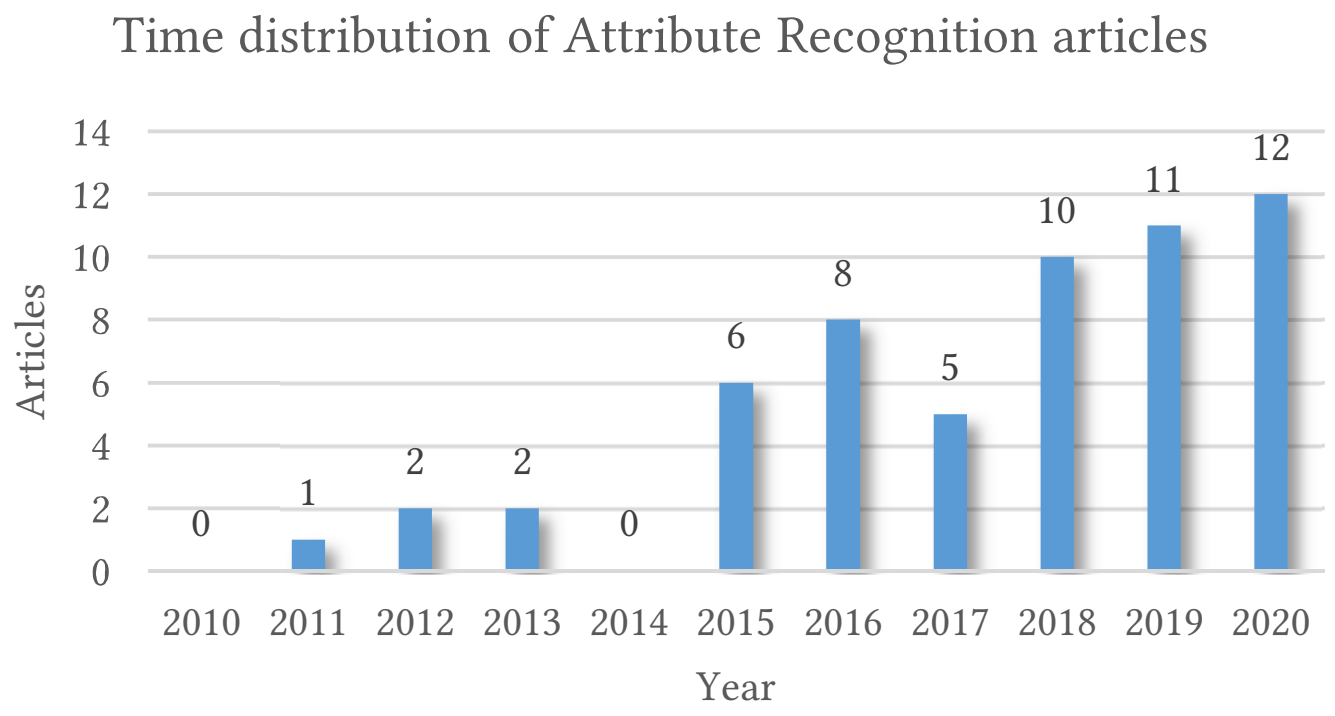

Figure 7: Time Analysis of Attribute Recognition Articles

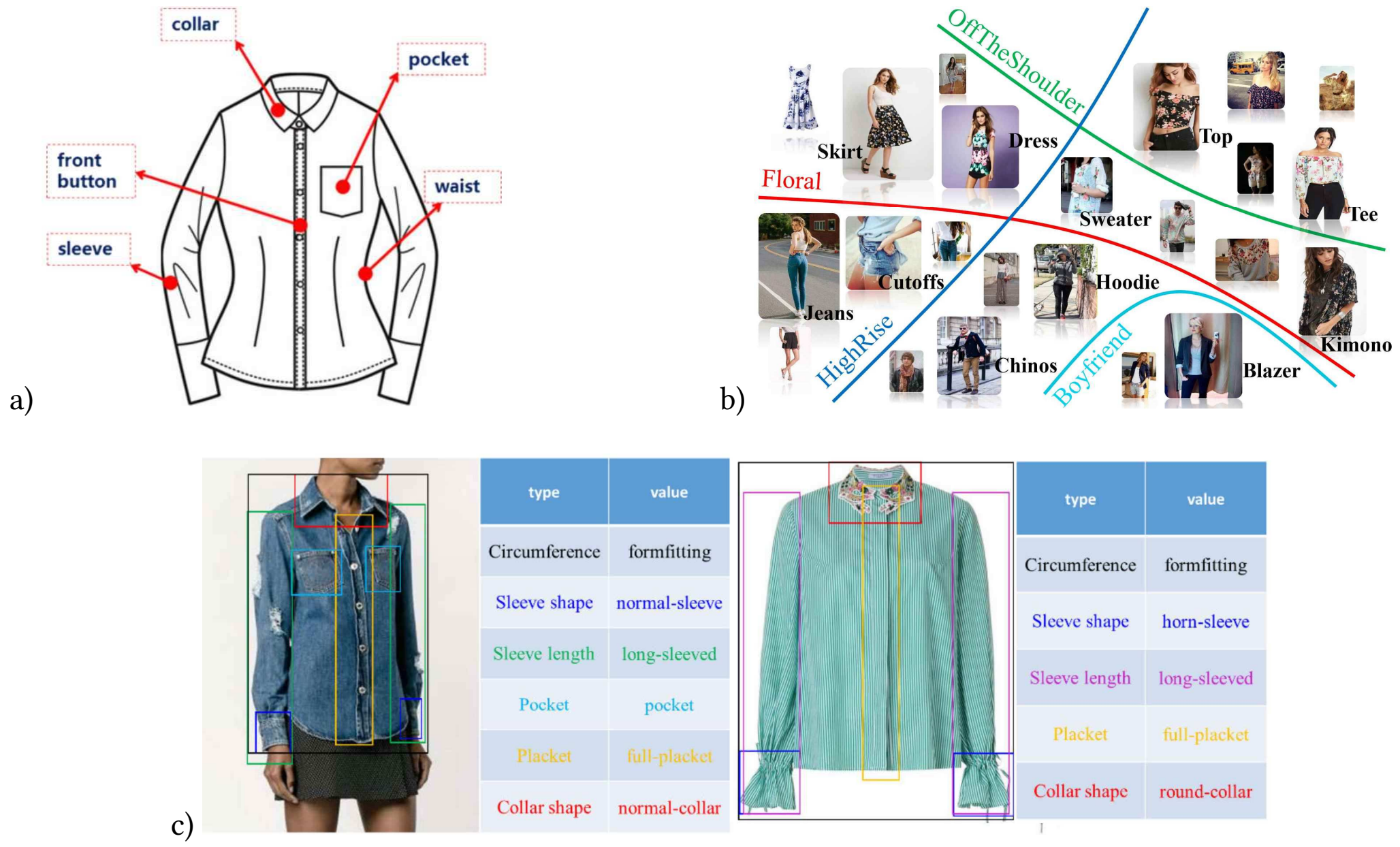

Figure 8: a) Attributes of a woman shirt [120] b) Deepfashion attributes overlap [123] c) Attributes detection [120]

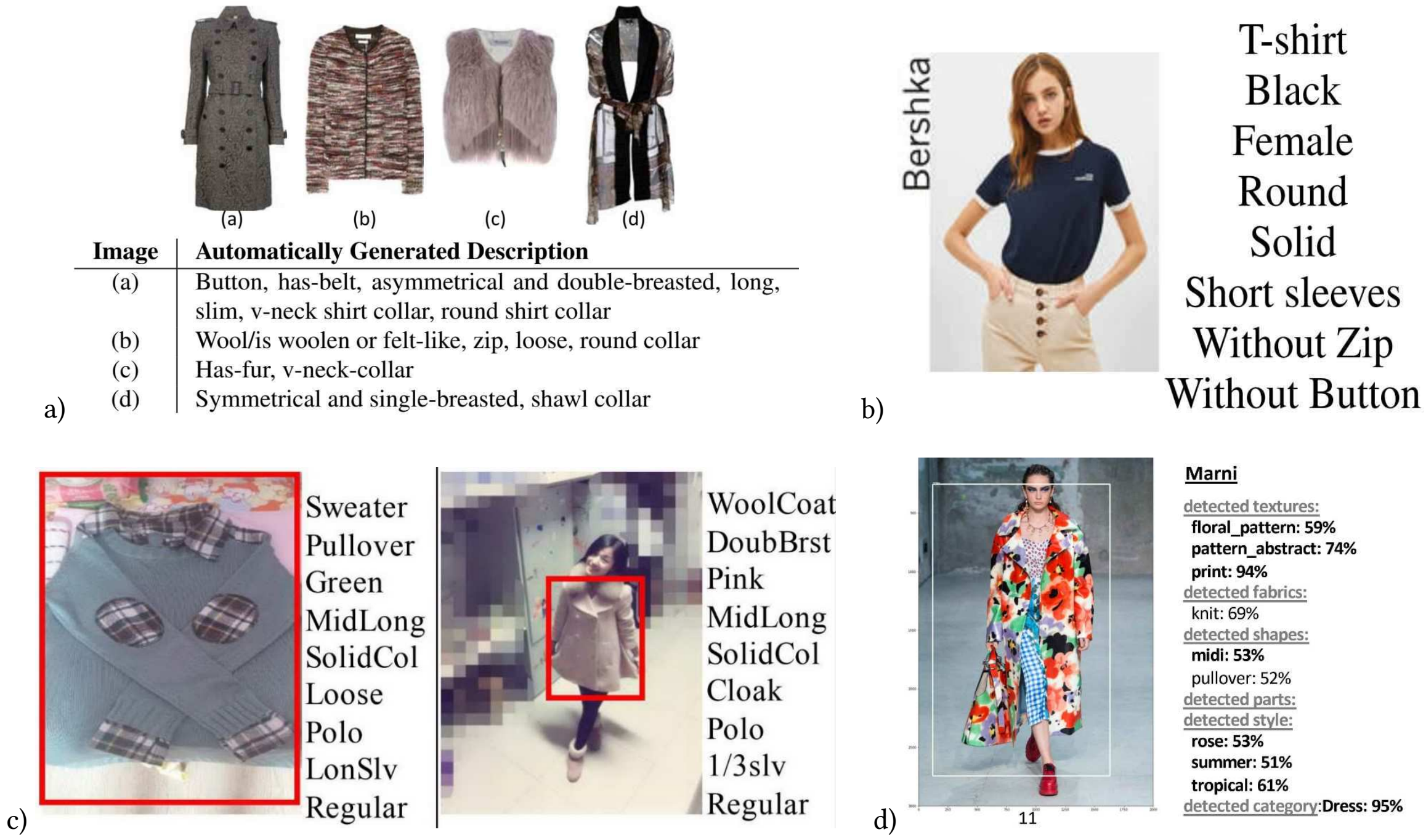

Figure 9: Attribute recognition on a) Item images [84] b) Model image [116] c) Wild images (with detection) [35] d) Fashion show videos [121]

### 2.3 Detection

Detection tasks aim to pinpoint a target's location in images and are used in many higher-level applications. For example, it might be the location of a fashion item or just the informationally-rich areas of the picture. Thus, we break this task into three sub-categories: 1) Item Detection, 2) Parsing or Segmentation, and 3) Landmark Detection.

#### *2.3.1 Item Detection*

Item detection task focuses on finding fashion items in images/videos and usually outputs one or multiple bounding boxes containing the items.

Table 3: Articles Related to Item Detection

| No | Article Reference | Year | Technical Keywords/Claimed Results | Application Notes |
|---|---|---|---|---|
| 1 | Iwata [124] | 2011 | Face detection algorithm | Top/Bottom, Full-body Fashion magazines |
| 2 | S. Liu [83] | 2012 | Part-based detection scheme | Upper/Lower body, Magic closet |
| 3 | Bossard [85] | 2013 | Face detection, Calvin upper body detector | Natural scenes |
| 4 | Lao [22] | 2015 | R-CNN, Selective Search, 93.4% Val Acc | Street photos |
| 5 | Chen [89] | 2015 | Classic, Deformable-part-model, latent SVM | Component detection, Kinect |
| 6 | Qian [31] | 2017 | Region-based FCN, R-CNN, SSD, 83.4% mAP | Multiple items, Street photos |
| 7 | Shankar [125] | 2017 | VGG-16, Faster R-CNN, Avg 68.2% mAP | For the recommendation task, Wild images |
| 8 | Dong [35] | 2017 | Multi-Task Curriculum Transfer, Faster R-CNN | Street photos |
| 9 | Y. Liu [126] | 2018 | R-CNN body detection, 92.01% Avg Acc | Upper/Lower body, compare with part-based |
| 10 | S. Zhang [102] | 2018 | Fast R-CNN, VGG-16, 92.80% mAP | Video (Fashion shows) |
| 11 | Zheng [104] | 2018 | Faster R-CNN, SSD, YOLO, 82% mAP | Benchmark 3 methods, ModaNet |
| 12 | Manandhar [127] | 2018 | Faster R-CNN, RPN, 96% logo Acc, 98% item Acc | Item & Brand logo joint detection |
| 13 | Ramesh [128] | 2018 | Faster R-CNN, Inception, ResNet-V2, 84.01% mAP | 4 Methods comparison |
| 14 | Ge [129] | 2019 | Mask R-CNN, Match R-CNN, 66.7% AP box | Benchmark, DeepFashion2, Street images |
| 15 | Tran [61] | 2019 | SSD 512, YOLO V3 300/416, ResNet50, 72%mAP | Street photos, For retrieval task |
| 16 | Sidnev [130] | 2019 | CenterNet, DeepMark, Hourglass, 72.3% AP box | DeepFashion2 Challenge, Multi items |
| 17 | H. Zhang [131] | 2020 | Faster R-CNN, SSD, YOLO V2, 97.99% mAP | Benchmark multiple methods |
| 18 | S. Zhang [114] | 2020 | SSD, VGG-16, 92.92% mAP | Street photos, Single item |
| 19 | X. Liu [115] | 2020 | MaskRCNN, ResNet50-FPN, 59.9% AP box | MMFashion Toolbox |
| 20 | Ji [132] | 2020 | Adaptive training sample selection, 72.8% AP | Wild image, For retrieval task |
| 21 | Ravi [133] | 2020 | Pose detection, Mask RCNN, 78% mAP | Full frontal images, Multiple items |
| 22 | Sidnev [134] | 2020 | CenterNet, DeepMark++, 73.7% mAP box | Real-time, Smartphone use |
| 23 | Tian [80] | 2021 | RCNN Multi-grained branches, 69.02% AP | Category grouping |
| 24 | Kim [135] | 2021 | EfficientDet, CoordConv, 68.6% mAP box | Multiple items, Efficient time, Light power |

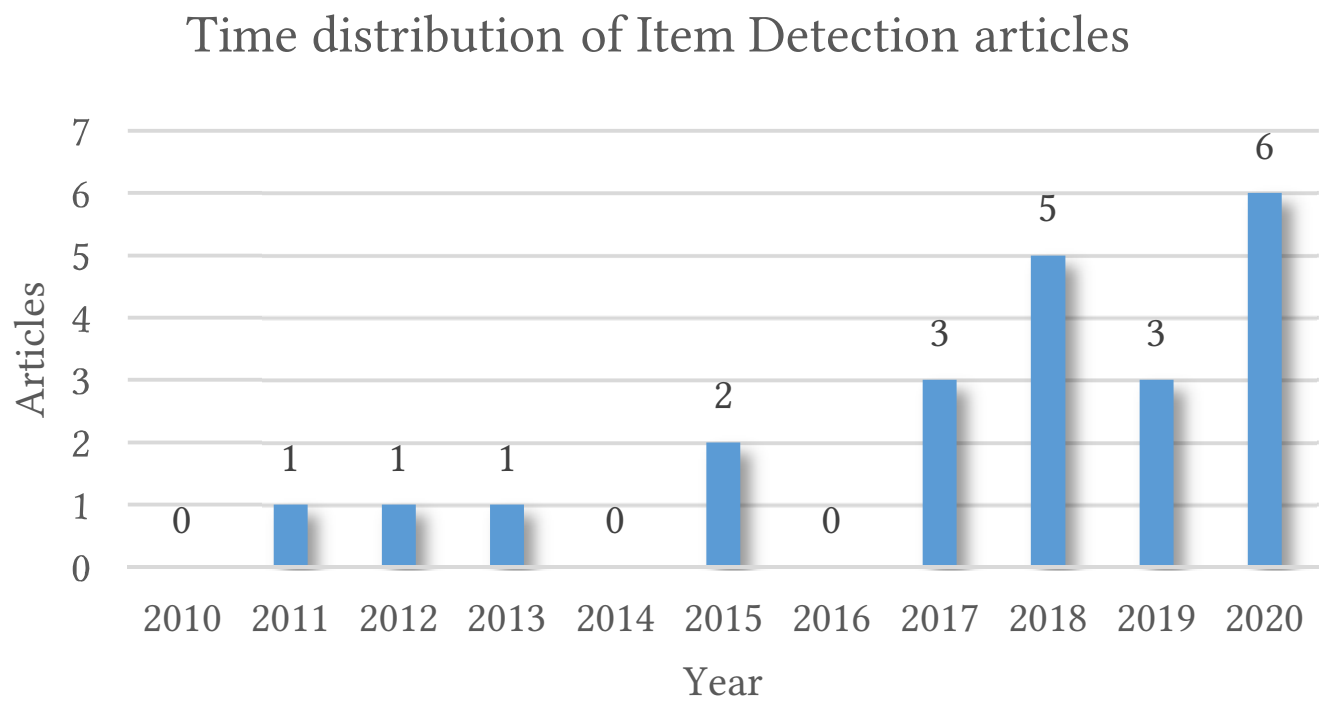

Figure 10: Time Analysis of Item Detection Articles

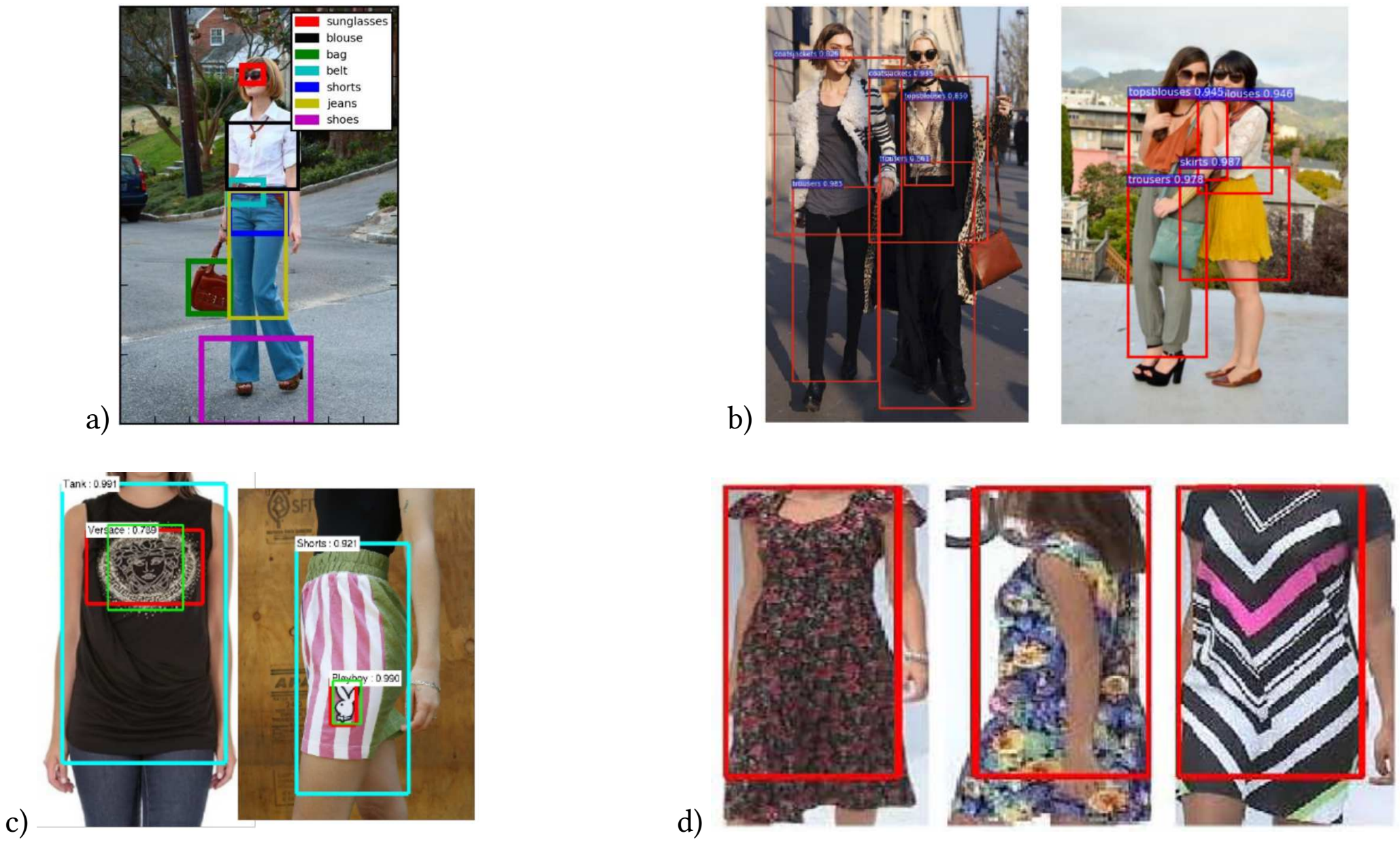

Figure 11: Item detection for a) Multiple items, a single person [22] b) Multiple items, multi-person [31] c) Single item, logo detection [127] d) Single item, fashion show videos [102]

#### 2.3.2 Parsing (Segmentation)

Fashion parsing is the semantic segmentation of clothing items, and each segment comes with a category label. The main difference between item detection and parsing is that the former generates only a bounding box around the object. In contrast, in parsing, we label fashion articles on a pixel level which is a much more complex task, especially for fashion items due to factors like human pose, occlusion, deformation, etc. Table 4 lists related articles. Abbreviations such as DR (Detection Rate) and NE (Normalized Error) are used whenever necessary.

Table 4: Articles Related to Parsing

| No | Article Reference | Year | Technical Keywords/Claimed Results | Application Notes |
|---|---|---|---|---|
| 1 | M. Yang [11] | 2011 | Region growing, Face Det., Canny Edge Det., Voronoi | Surveillance videos, Background removal |
| 2 | Wang [136] | 2011 | Layout model, Clothing & blocking forest, 92.8% Acc | Multi-person, Occlusion relation |
| 3 | Yamaguchi [137] | 2012 | CRF, Pose estimation, MAP, Superpixel, 89% Acc | Street photos |
| 4 | Kalantidis [19] | 2013 | Articulated pose Est., Graph-based, AGM, 80.2% Acc | For recommendation task, Street images |
| 5 | Dong [138] | 2013 | Deformable Mixture, Parselets, 84.6% mIOU | Street photos |
| 6 | Yamaguchi [139] | 2013 | Global; NN; and Transferred parse, 84.68% Acc | A combination of 3 methods |
| 7 | W. Yang [140] | 2014 | Exemplar-SVM, Graph Cuts, MRF, 90.29% Acc | Image co-segmentation, Street photos |
| 8 | S. Liu [141] | 2014 | Pose estimation, MRF-based, Superpixel, 42.1% IOU | Weak color-category labels |
| 9 | Yamaguchi [142] | 2014 | Pose Est., Transferred parse, NN parse, 84.68% Acc | Similar styles retrieval for parsing |
| 10 | Kiapour [20] | 2014 | Pose estimation, Unrestricted parsing | Parsing for style indicator |
| 11 | S. Liu [143] | 2015 | Active learning, Pose Est., SIFT, Superpixel, 88.92% Acc | Video co-parsing, Multi-person |
| 12 | Simo-Serra [144] | 2015 | CRF, CPMC, clothelets, 84.88% Acc | Street photos |
| 13 | Liang [145] | 2015 | Active template regression, CNN, 91.11% Acc | Street photos |
| 14 | Liang [146] | 2015 | Co-CNN, Within-super-pixel smoothing, 97.06% Acc | Street photos, Chictopia10k |
| 15 | S. Liu [147] | 2015 | Quasi-parametric, Matching-CNN, KNN, 89.57% Acc | Street photos |
| 16 | Z. Li [96] | 2016 | FCN, Hierarchical superpixel merging, exemplar-SVM | Cross-Domain, For retrieval task |
| 17 | Qian [31] | 2017 | FCN, ASPP, CRF, FasterR-CNN, DeepLab, 59.66% mPA | Street photos |
| 18 | Tangseng [148] | 2017 | FCN+Side-branch, CRF, VGG-16, 92.39% Acc | Street photos |
| 19 | Xia [149] | 2017 | Part & Pose FCNs, FCRF, Pose Est., 64.39% mIOU | Multi-Person, Human part, Wild photos |
| 20 | J. Li [150] | 2017 | MH-Parser, Graph-GAN, 37.01% PCP Top-5% Overlaps | Multi-Person, Human parsing, Wild |
| 21 | Gong [151] | 2017 | DeepLabV2, SSL, FCN-8s, SegNet, 84.53% Acc | Benchmark, different methods |
| 22 | Zheng [104] | 2018 | FCN, CRFasRNN, DeepLabV3+, 51.14% Avg IOU | Benchmark, ModaNet |
| 23 | Gong [152] | 2018 | Part grouping network, Deeplab-v2, 68.40% Avg IOU | Benchmark, CIHP, Multi-person |
| 24 | Zhou [153] | 2018 | ATEN, Parsing-RCNN, convGRU, 37.9% mIOU | Benchmark, VIP, Video, Multi-person |
| 25 | Liang [154] | 2018 | Joint human parsing & pose Est., 51.37% mIOU | Benchmark, LIP, Parsing+Pose Est., Wild |
| 26 | Zhao [155] | 2018 | Nested adversarial network, 34.37% PCP0.5 | Benchmark, MHP V2, Multi-person, Wild |
| 27 | Jain [156] | 2019 | Nearest neighbor, Pose distance, CRF, 85.92% Acc | Street photos |
| 28 | Ge [129] | 2019 | Mask R-CNN, Match R-CNN, 67.4% AP mask | Benchmark, DeepFashion2, Street images |
| 29 | Lasserre [157] | 2019 | CNN, U-net, ~97.8% mean Acc | Background removal, Street2Fashion2Shop |
| 30 | Griebel [158] | 2019 | Mask R-CNN, Feature pyramid network | Fashion Curation System |
| 31 | Xu [159] | 2019 | Multi-task learning, JFNet, DeepLabV3+, 84.65% mIOU | Part parsing, For 3D modeling |
| 32 | Ruan [160] | 2019 | Context embedding+Edge perceiving, 56.50% mIoU | Multi-person, 1st in 2nd LIP Challenge |
| 33 | Hidayati [65] | 2019 | Price-collecting Steiner tree | Street images, for Genre recognition |
| 34 | Gong [161] | 2019 | Graph Transfer Learning, Graphonomy, 71.14% mIOU | Universal, Multiple datasets, Multi-person |
| 35 | Wang [162] | 2019 | Compositional neural information fusion, 57.74% mIOU | Multi-person, Multiple datasets |
| 36 | X. Liu [115] | 2020 | Mask R-CNN, ResNet50-FPN, 58.4% Ap mask | MMFashion Toolbox |
| 37 | Castro [163] | 2020 | Unet, SegNet, Atrous, FCN, DenseNet, 93% Acc | Compares different models |
| 38 | Zhang [164] | 2020 | Body generation, PConvNet, Graphonomy | Fine-grained parsing (e.g. right/left sleeves) |
| 39 | Shi [121] | 2020 | Mask R-CNN, Segmentation | Trend Analysis, Fashion show Videos |
| 40 | Chou [165] | 2021 | Cloth2pose, PGN, Pose-guided parsing translator | Change parsing based on clothing & pose |

| No | Article Reference | Year | Technical Keywords/Claimed Results | Application Notes |
|---|---|---|---|---|
| 41 | Lewis [166] | 2021 | Pose-conditioned StyleGAN2, AdaIN | Change parsing based on clothing & pose |

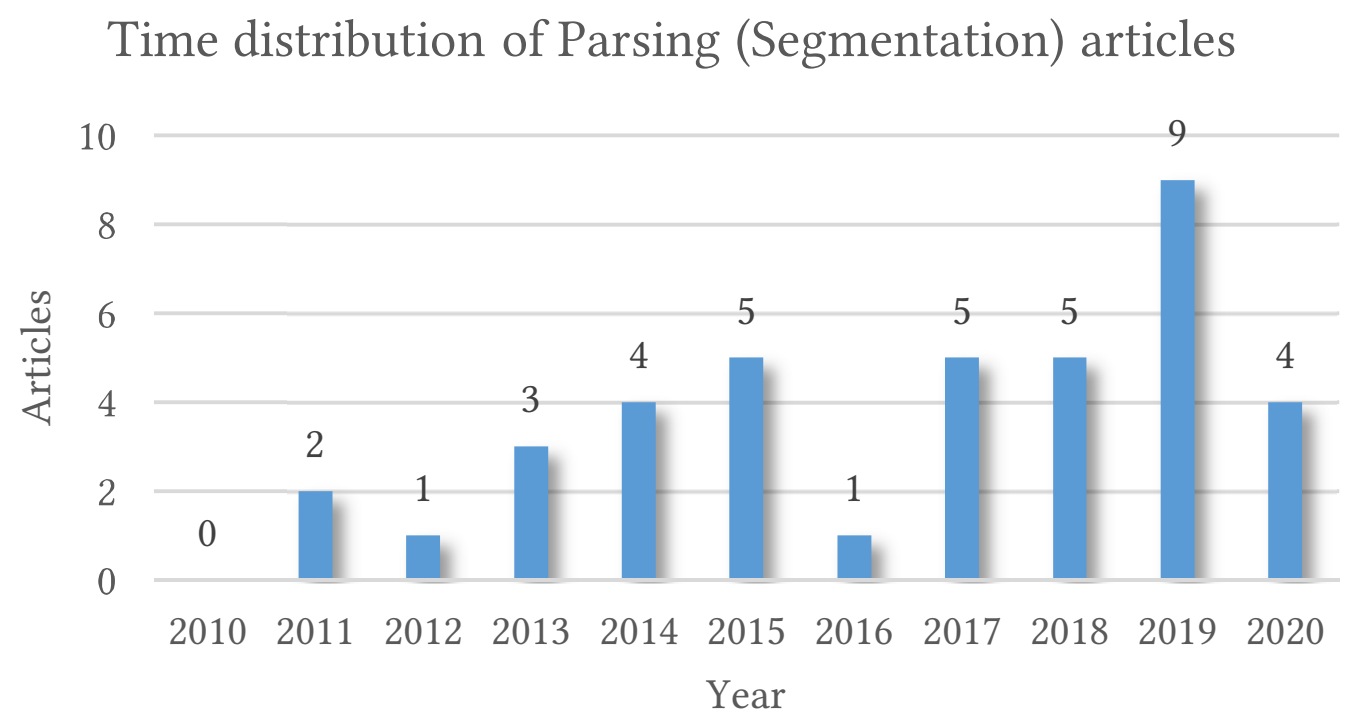

Figure 12: Time Analysis of Parsing (Segmentation) Articles

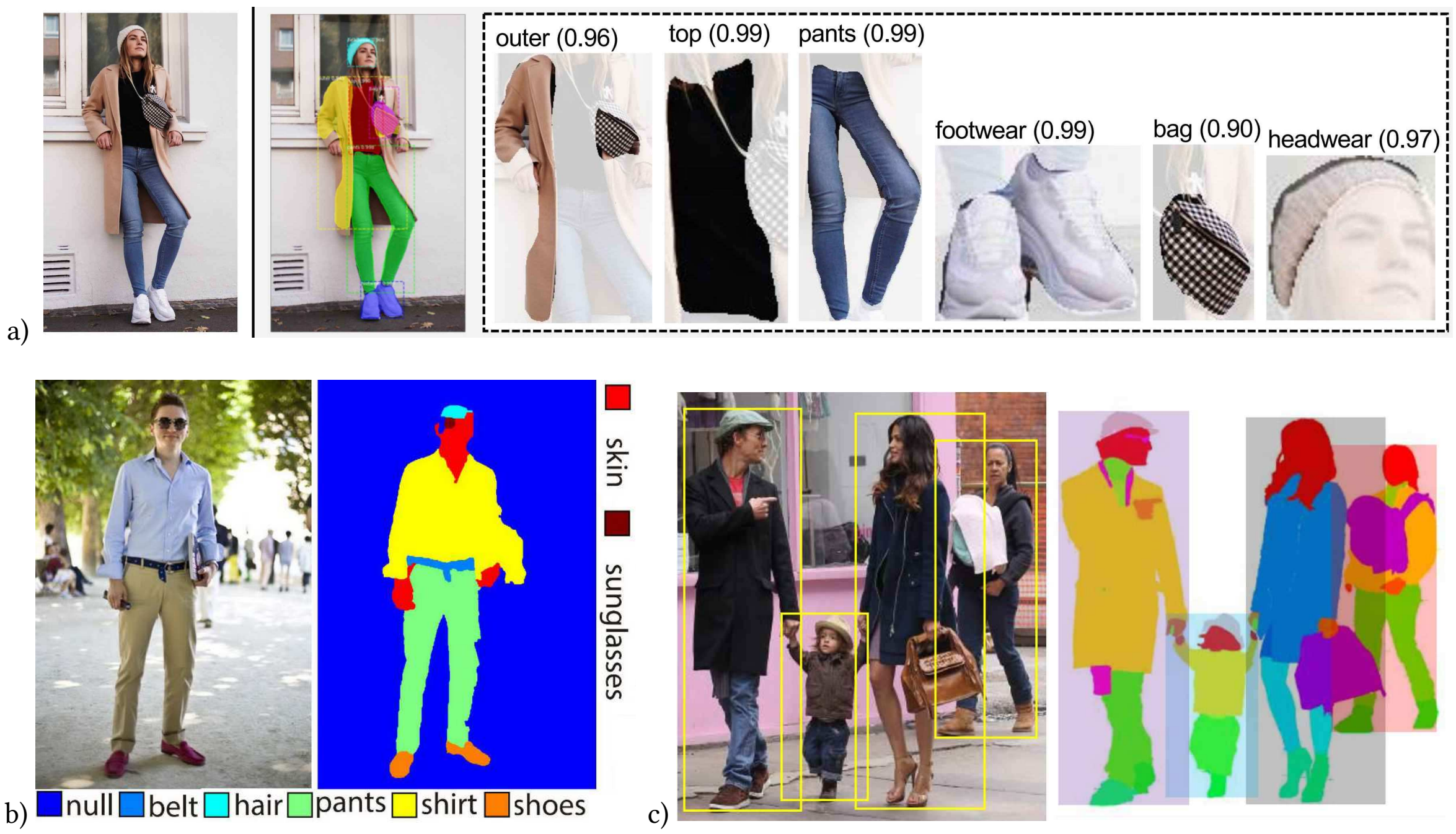

Figure 13: Semantic segmentation examples. a) Detected garments with probabilities [158] b) Parsing street images [140] c) Multi-human parsing [155]

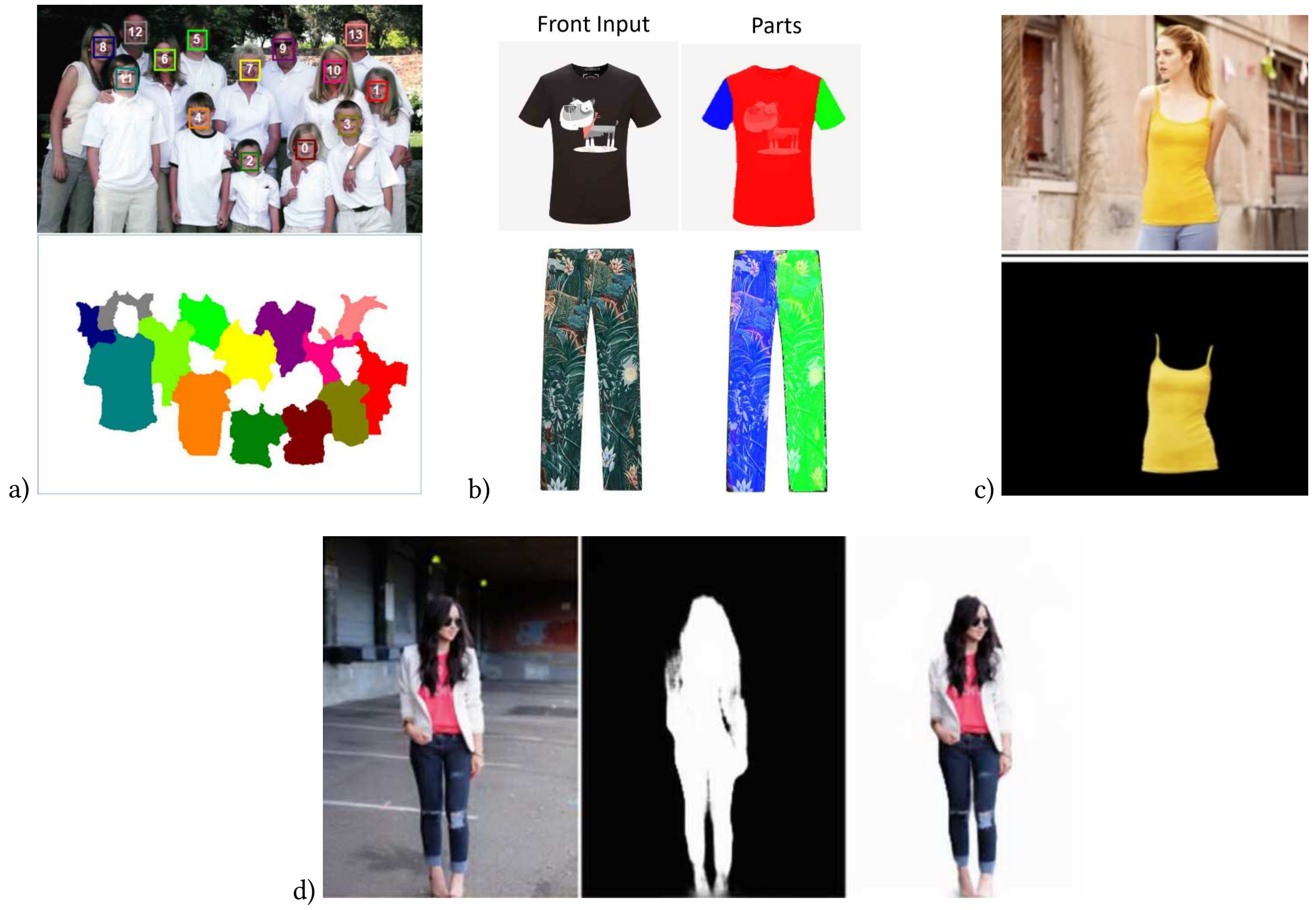

Figure 14: a) Parsing in crowded scenes with heavy occlusion [136] b) Part parsing [159] c) Clothing background removal [65] d) Model background removal [157]

### *2.3.3 Landmark Detection*

First introduced in 2016 [9], landmark detection aims to find key points of fashion items. For example, landmarks for upper-body items can be left/right collar end, left/right sleeve end, etc. These landmarks also implicitly contain bounding boxes, and landmark pooling proved to enhance performance in certain applications [115].

Table 5: Articles Related to Landmark Detection

| No | Article Reference | Year | Technical Keywords/Claimed Results | Application Notes |
|---|---|---|---|---|
| 1 | Z. Liu [9] | 2016 | VGG-16, FashionNet, Landmark visibility, ~80% DR | Benchmark, DeepFashion |
| 2 | Z. Liu [167] | 2016 | VGG-16, Pseudo-labels, Network Cascade, 78.6% mDR | Benchmark, FLD, Wild images |
| 3 | Yan [168] | 2017 | Hierarchical recurrent spatial transformer, 73.8% mDR | Wild images |
| 4 | Chou [169] | 2018 | Key-point detection, CPM, Gaussian peak heatmap | For shoe try-on task |
| 5 | Wang [51] | 2018 | VGG-16, Fashion grammar, BCRNN 58.3% mDR | Wild images |
| 6 | Li [55] | 2019 | Two-stream multi-task network, Hourglass, 0.0467 NE | For classification On DeepFashion dataset |
| 7 | Ge [129] | 2019 | Mask R-CNN, Match R-CNN, 56.3% AP pt | Benchmark, DeepFashion2, Street images |
| 8 | J. Liu [59] | 2019 | Feature map upsampling, Gaussian filter, 0.0474 NE | For fashion analysis |
| 9 | Xu [159] | 2019 | Convolutional pose machines, ResNet-101, 0.0265 mNE | For 3D modeling |
| 10 | Sidnev [130] | 2019 | CenterNet, DeepMark, Hourglass, 53.2% mAP pt | DeepFashion2 Challenge, Multi items |

| No | Article Reference | Year | Technical Keywords/Claimed Results | Application Notes |
|---|---|---|---|---|
| 11 | Lee [170] | 2019 | VGG-16, Global-local embedding, 0.0393 Avg NE | Wild images |
| 12 | Chen [171] | 2019 | Dual attention feature enhance, FPN, 0.0342 Avg NE | On DeepFashion, FLD, and DeepFashion2 |
| 13 | X. Liu [115] | 2020 | Mask R-CNN, ResNet50-FPN, ~78% DR < 30 Pix. Distance | MMFashion Toolbox |
| 14 | Lin [172] | 2020 | Homogeneity, Aggregation, Fine-tuning, 58.9% AP | 1st in the DeepFashion2 Challenge 2020 |
| 15 | Shajini [69] | 2020 | VGG-16, Multiscale, SDC, 0.0425 Avg NE | For attribute detection |
| 16 | Sidnev [134] | 2020 | CenterNet, DeepMark++, Hourglass, 59.1% mAP pt | Key-point grouping, Real-time |
| 17 | Sidnev [173] | 2020 | CenterNet, Hourglass, Clustering, 59.2% mAP | Key-point grouping |
| 18 | Ziegler [73] | 2020 | Feature map upsampling, Gaussian filter, 0.1047 Avg NE | Sorting, Robotics, In-lab & catalog images |
| 19 | Bu [174] | 2020 | Multi-depth dilated Net., B-OHKM, 0.0221 Avg NE | Street photos |
| 20 | Lai [175] | 2020 | Cascaded pyramid network | Multi-Person, For try-on task |
| 21 | Roy [176] | 2020 | Human/Fashion correlation layer | Human+Fashion landmarks, For try-on |
| 22 | Xie [177] | 2020 | Pose estimation, MSPN | For try-on task |
| 23 | Kim [135] | 2021 | EfficientDet, BiFPN, CoordConv, 45.0% mAP | Multiple items, Efficient time |

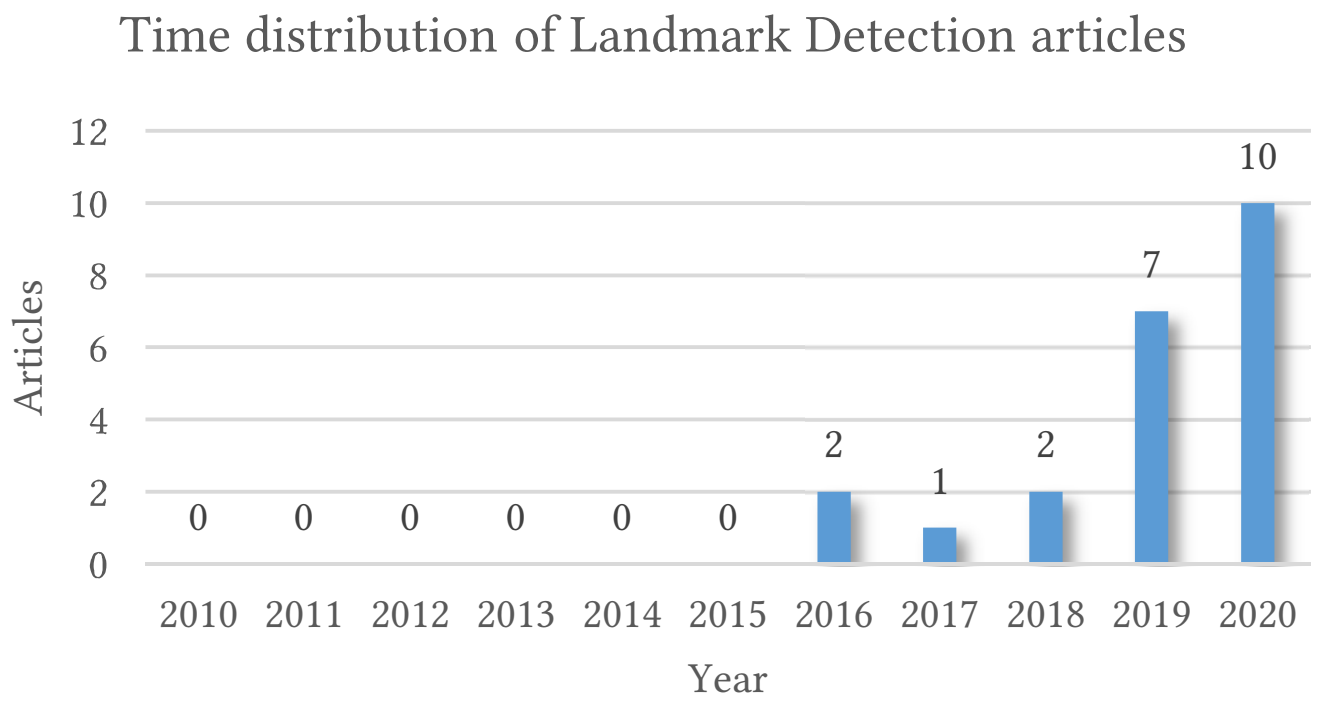

Figure 15: Time Analysis of Landmark Detection Articles

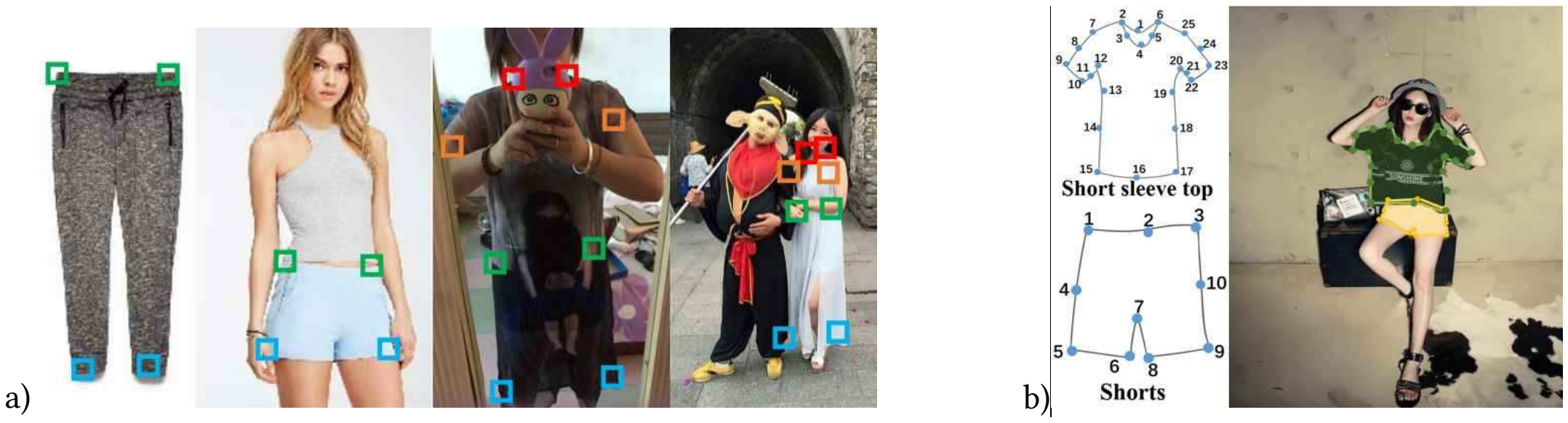

Figure 16: a) Single-item landmark detection for Item, Model, and wild images [9] b) Multi-item landmark examples [129]

a) 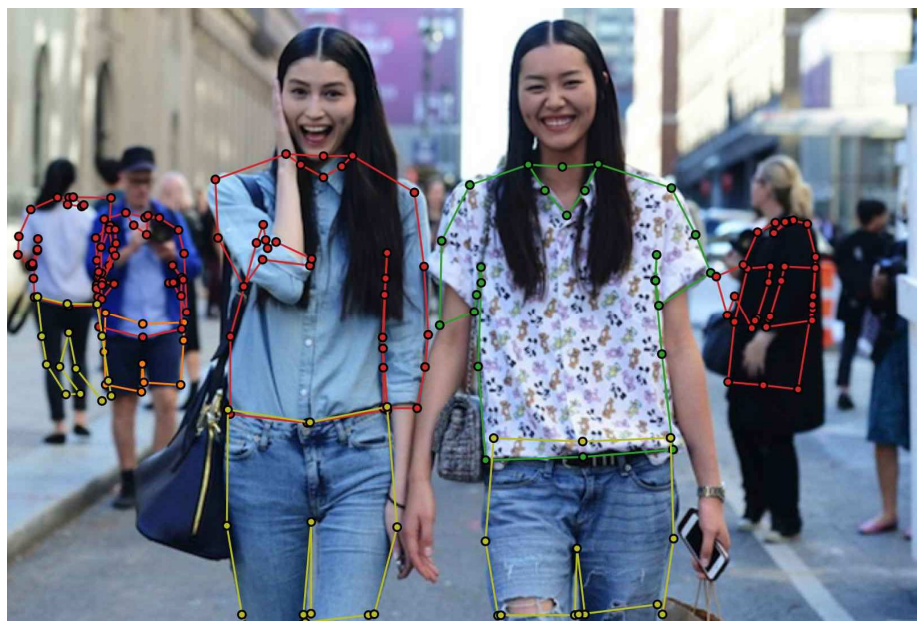 b) 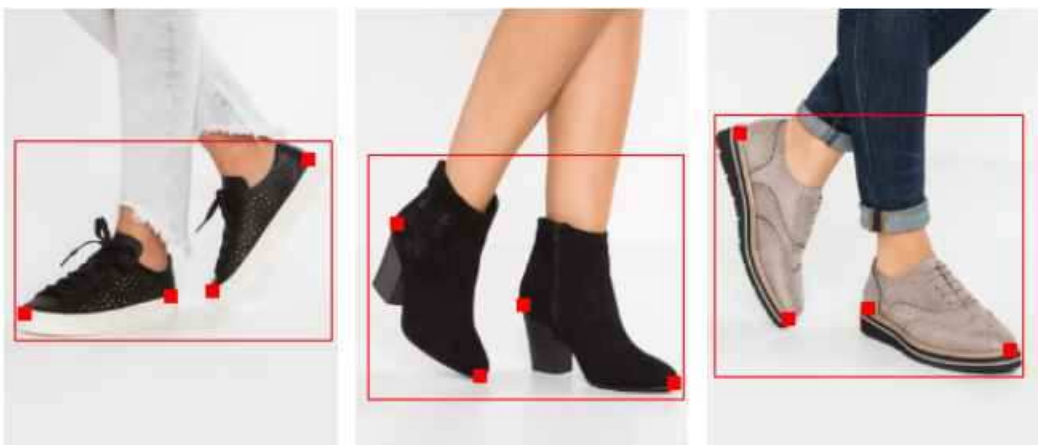

Figure 17: a) Multi-item and multi-person landmark detection in the wild [134] b) Shoes key-points examples [169]

### 2.4 Virtual Try-on

Virtual try-on is a highly active field, primarily due to its potential applications in the online fashion retail industry and also offline intelligent software packages used in clothes stores. We separate virtual try-on into five sub-categories: 1) Image-Based Try-On, 2) 2D Modeling, 3) 3D Modeling, 4) Size & Fit, and 5) Magic Mirror. Remember that the image-based try-on task is also 2-dimensional, but it does not change the input image, just the clothing items. Reference [178] is a 2020 taxonomical survey on virtual try-on systems with GAN.

#### *2.4.1 Image-Based Try-On*

Image-based try-on systems usually take one image as input and change fashion items present in the photo according to the user's need. The changes only take effect on specific regions of the input image, and the rest remains intact. There are also makeup transfer and hairstyle suggestion applications that we only report a few examples of and do not fully cover in this article. Image-based try-on systems typically take two inputs, one reference image, one target outfit, and transfer the outfit to the reference image. In Table 6, we try to report the exact type of this transfer using dual-keywords (Target-Reference) in the "Application Notes" section. These systems transfer qualities of the "Target" to the "Reference" image; for example, Model-Model designs transfer clothing from one human model image to another image with the human model present, whereas Title-Model systems need an in-shop catalog image of the desired outfit as target. Studies use different evaluation metrics such as Inception Score (IS), Human Score (HS), Structural Similarity (SSIM), and various others.

Table 6: Articles Related to Image-Based Try-On

| No | Article Reference | Year | Technical Keywords/Claimed Results | Application Notes |
|---|---|---|---|---|
| 1 | W. Yang [179] | 2012 | Active shape model, Matting, Statistical learning | Hairstyle, Recommender |
| 2 | Hauswiesner [180] | 2013 | Image-based visual hull rendering, IBVH | Model-Model, In-lab images |
| 3 | Liu [181] | 2014 | Dual linear transformation, Guided filter, Alpha blending | Hairstyle & makeup, Recommender |
| 4 | S. Yang [182] | 2016 | Joint material-pose optimization, Pose Est. | Model-Model, Optimized virtual 3D outfit |
| 5 | Jetchev [183] | 2017 | Conditional Analogy GAN, PatchGAN | Title-Model, Upper body |
| 6 | Zhu [184] | 2017 | FashionGAN, Segmentation, Text-to-image, 82.6% mAP | Text-Model, Text-Guided, Upper body |
| 7 | Han [185] | 2018 | VITON, Multi-task Encoder-Decoder, TPS, 2.514 IS | Title-Model, Upper body, Also wild |
| 8 | Chou [169] | 2018 | Pose Invariant, PIVTONS, PatchGAN, Key-points | Shoe try-on, Title-Model |
| 9 | Raj [186] | 2018 | Segmentation, Dual-path U-net, DRAGAN, SwapNet | Model-Model, Swap clothes, Pose |

| No | Article Reference | Year | Technical Keywords/Claimed Results | Application Notes |
|---|---|---|---|---|
| 10 | B. Wang [187] | 2018 | Characteristic-Preserving, CP-VTON, 84.5% HS | Title-Model, Upper body |
| 11 | Chen [188] | 2018 | CAGAN, LIP-SSL, Transform, 90.3% HS | Title-Model, Upper body |
| 12 | Zanfir [189] | 2018 | 3D pose & Shape, DMHS, SMPL, HAS, 4.13 IS | Model-Model, Swap clothes |
| 13 | Han [190] | 2019 | FiNet, Parser, Pose Est., Encoder-decoder, VGG-19 | Model-Model, Transfer using inpainting |
| 14 | Lomov [191] | 2019 | Pix2pix, CGAN, Perceptual loss functions, 3.098 IS | Title-Model, Upper body |
| 15 | Wu [192] | 2019 | Pose alignment, Texture refinement, M2E-TON, 83.7% HS | Model-Model |
| 16 | Ayush [193] | 2019 | Auxiliary learning, Human segmentation, 0.712 SSIM | Title-Model, Preserves characteristics |
| 17 | Yildirim [194] | 2019 | Modified Conditional Style GAN, 9.63 FID | Model-Model, Color transfer, High-Res. |
| 18 | Issenhuth [195] | 2019 | Warping U-net, WUTON, Geometric Trans., 0.101 LPIPS | Title-Model, Handles masked images |
| 19 | L. Yu [196] | 2019 | Inpainting-based, I-VTON, TIN, Triplet, 2.729 IS | Model-Model, Selective article transfer |
| 20 | Honda [197] | 2019 | LA-VITON, Geometric Matching, SNGAN, 78.78% HS | Title-Model |
| 21 | Kikuchi [198] | 2019 | Spatial Transformer, ST-GAN, 32% IOU@0.75 | Glasses, Title-Model |
| 22 | Pumarola [199] | 2019 | Unsupervised, Memory, GAN, Segmentation, 3.94 IS | Image-to-Video clothing transfer |
| 23 | Honda [200] | 2019 | VITON-GAN, LIP, GMM, TOM | Title-Model, Occlusion |
| 24 | R. Yu [201] | 2019 | Feature Preservation, VTNFP, Segmentation, 77.38% HS | Title-Model |
| 25 | Han [202] | 2019 | Flow-based GAN, ClothFlow, Pyramid Net., 0.803 SSIM | Title-Model, Also Pose-Guided |
| 26 | Sun [203] | 2019 | Structural consistency, Mask R-CNN, U-net, GAN | Title-Model, Less missing body parts |
| 27 | Ayush [204] | 2019 | Multi-Scale Patch Adversarial Loss, 2.558 IS | Title-Model |
| 28 | Kubo [205] | 2019 | UV mapping, UVTON, DensePose, 59.38% HS | Title-Model |
| 29 | Zhang [206] | 2019 | Disentangled Representation, DMT, GAN, 0.992 SSIM | Makeup transfer, Four modes, Model-Model |
| 30 | W. Liu [207] | 2019 | Liquid warping GAN, Denoising Conv. auto-encoder | Model-Model, Also In-Lab images, Detailed |
| 31 | Pandey [208] | 2020 | Poly-GAN, Three stages in one network, 2.790 IS | Title-Model |
| 32 | Z. Yu [209] | 2020 | Unsupervised apparel simulation GAN, AS-GAN | Street-Street, CCTV, For person ReID task |
| 33 | Issenhuth [210] | 2020 | Student-teacher paradigm, Parser-Free, STN, 3.154 IS | Title-Model |
| 34 | Raffiee [211] | 2020 | GarmentGAN, Semantic parser, SPADE-style, 2.774 IS | Title-Model |
| 35 | Jeong [212] | 2020 | Graphonomy, SEAN, ResBlK, SEBlK, 0.865 SSIM | Model-Model, Selective article transfer |
| 36 | Minar [213] | 2020 | 3D model-based, CloTH-VTON, U-Net, 3.111 IS | Title-Model, 3D cloth reconstruction |
| 37 | H. Yang [214] | 2020 | Content Generating & Preserving, ACGPN, 2.829 IS | Title-Model, Detail preservation |
| 38 | Hashmi [215] | 2020 | Neural Body Fit, GAN, RPN, STN, 76.62% Acc | User custom try-on |
| 39 | Neuberger [216] | 2020 | O-VITON, pix2pixHD, Segmentation, cGAN, 3.61 IS | Multiple Models-Model, Multi-item try-on |
| 40 | Lai [175] | 2020 | Key-points matching, KP-VTON, Mask R-CNN, 2.80 IS | Title-Model |
| 41 | Roy [176] | 2020 | Landmark Guided, LGVTON, TPS, cGAN, 2.71 IS | Model-Model |
| 42 | Xie [177] | 2020 | Landmark-Guided, LG-VTON, MSPN, TPS, 2.885 IS | Title-Model |
| 43 | Jandial [217] | 2020 | SieveNet, Coarse-to-Fine Warping, TPS, 2.82 IS | Title-Model, Robust |
| 44 | Song [218] | 2020 | Shape-Preserving, SP-VITON, DensePose, 2.656 IS | Title-Model |
| 45 | Li [219] | 2020 | U-Net, Shape Matching, Cascade Loss, 7.04 FID | Title-Model, Chooses best Title-Model pair |
| 46 | K. Wang [220] | 2020 | Unpaired shape transformer, AdaIN, 66.42 SSIM | Title-Model, Try-on/Take-off |
| 47 | Fincato [221] | 2020 | Geometric Transformation, VITON-GT, 2.76 IS | Title-Model |
| 48 | Minar [222] | 2020 | 3D Model-based, SMPL, TPS, CP-VTON | Title-Model, 3D cloth reconstruction |
| 49 | Minar [223] | 2020 | CP-VTON+, CNN geometric matching, 3.1048 IS | Title-Model, Shape & texture preserving |
| 50 | Kips [224] | 2020 | Color Aware, CA-GAN, PatchGAN | Makeup transfer, Model-Model |
| 51 | Men [225] | 2020 | Attribute-Decomposed GAN, U-Net, AdaIN, VGG | Model-Model, Selective article transfer |
| 52 | Lewis [166] | 2021 | Pose-conditioned StyleGAN2, VOGUE, AdaIN, 32.21 FID | Model-Model, Selective article transfer |
| 53 | Minar [226] | 2021 | 3D deformation, CloTH-VTON+, Segmentation, 2.787 IS | Title-Model, Method comparison |

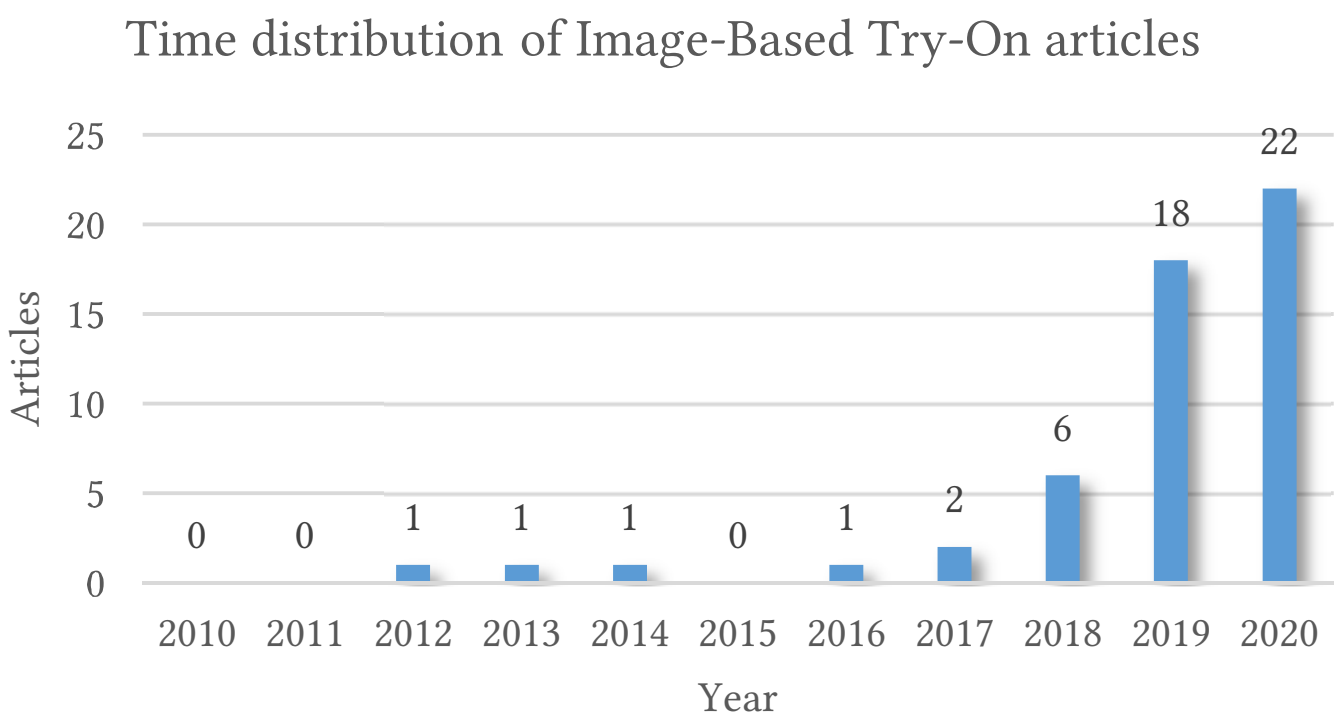

Figure 18: Time Analysis of Image-Based Try-On Articles

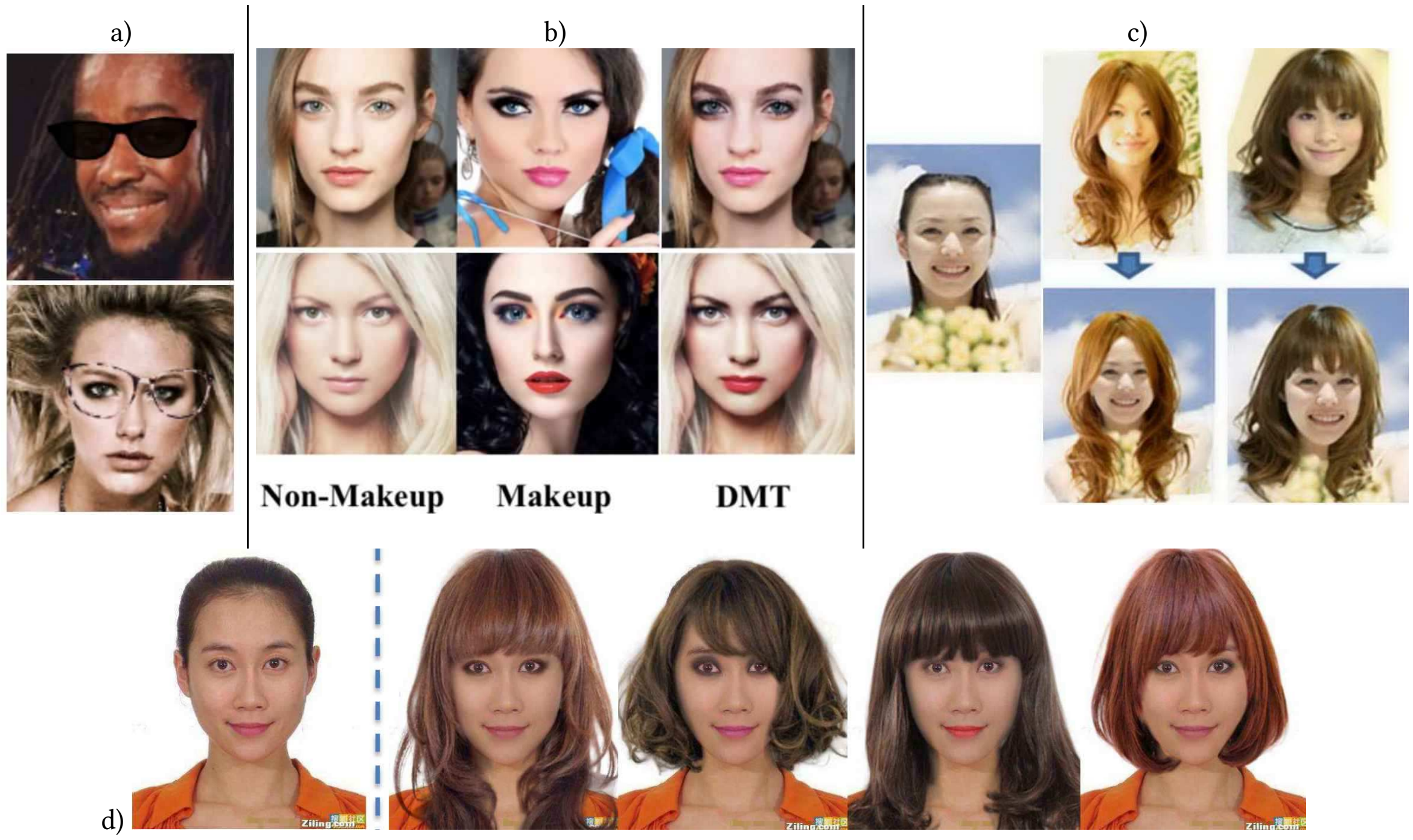

Figure 19: a) Glasses try-on [198] b) Makeup transfer [206] c) Hairstyle transfer [179] d) Hairstyle and makeup effects [181]

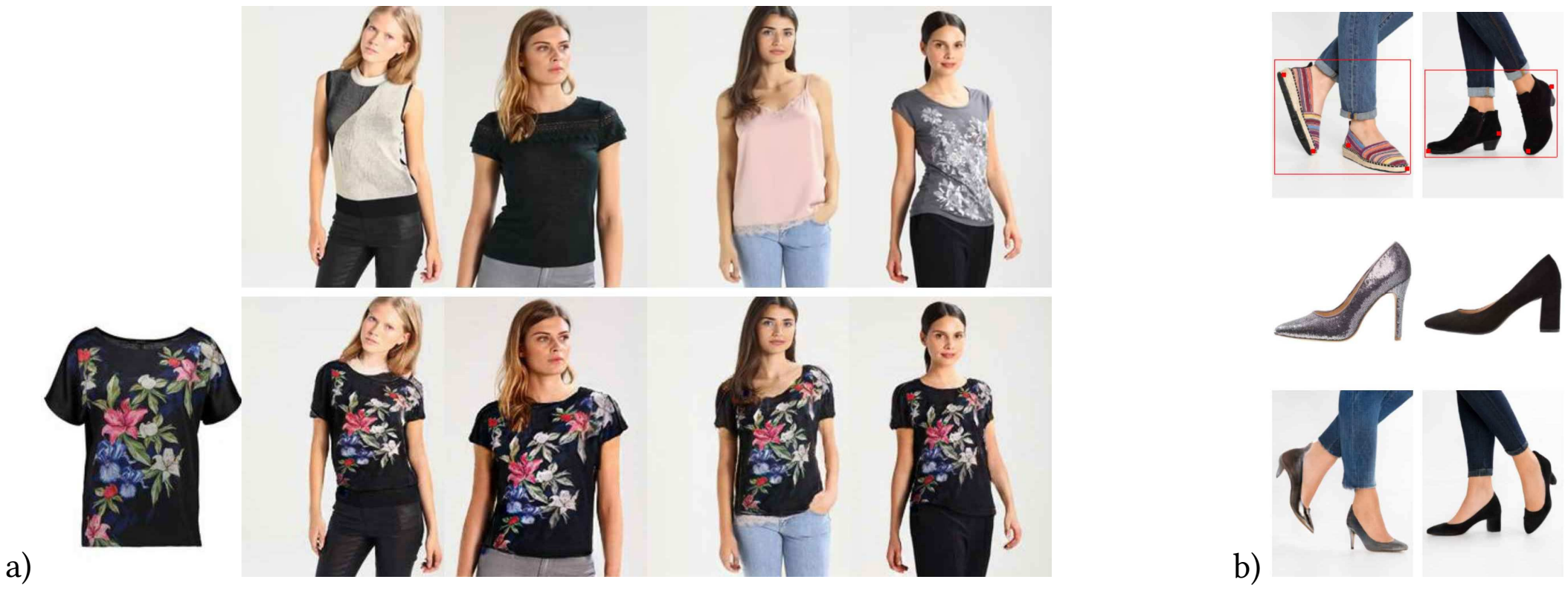

Figure 20: a) Item-Model try-on [210] b) Item-Model shoes try-on [169]

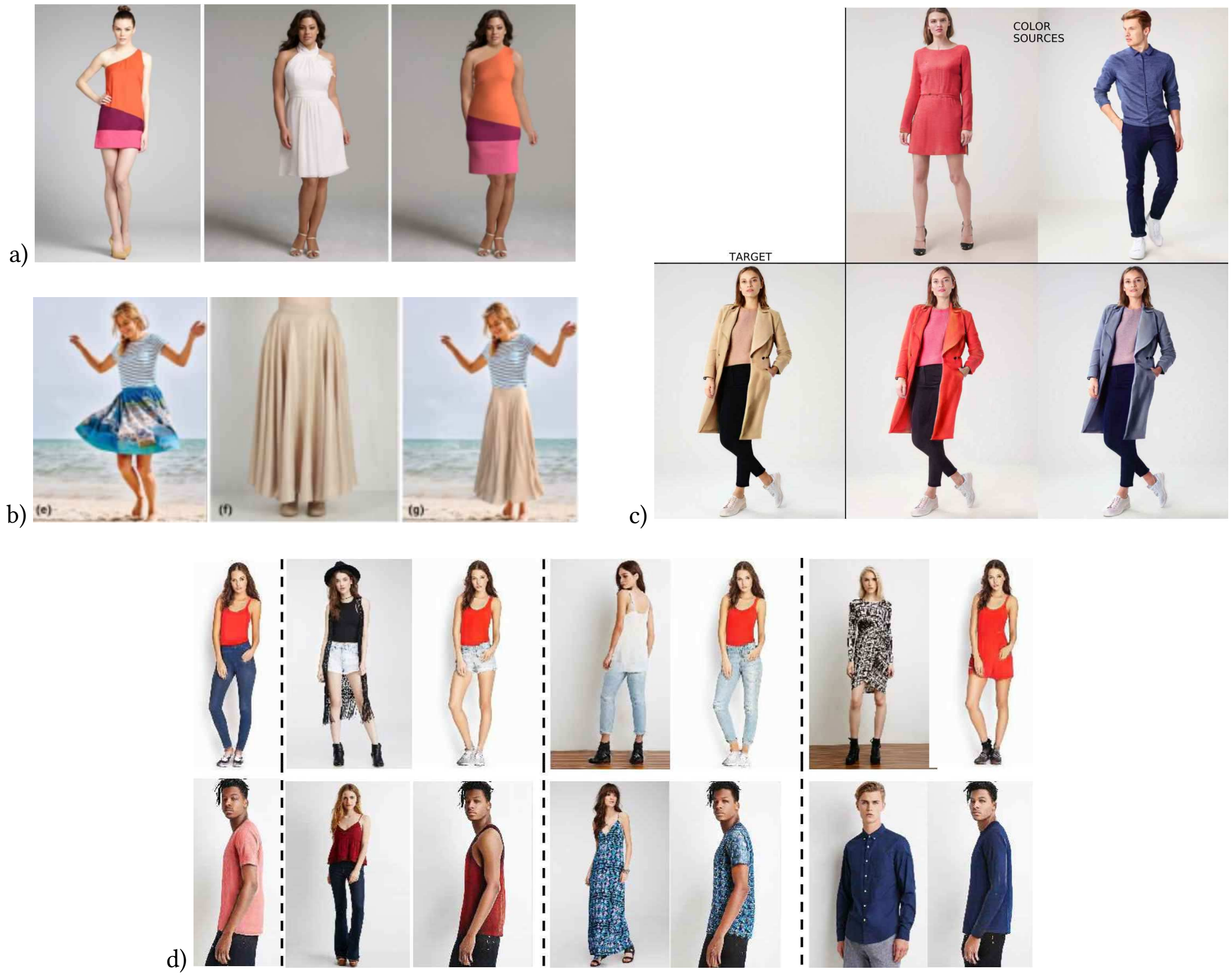

Figure 21: Model-Model try-on examples. a, b) Model and street try-on [182] c) Color transfer [194] d) Controllable try-on (selective article) [225]

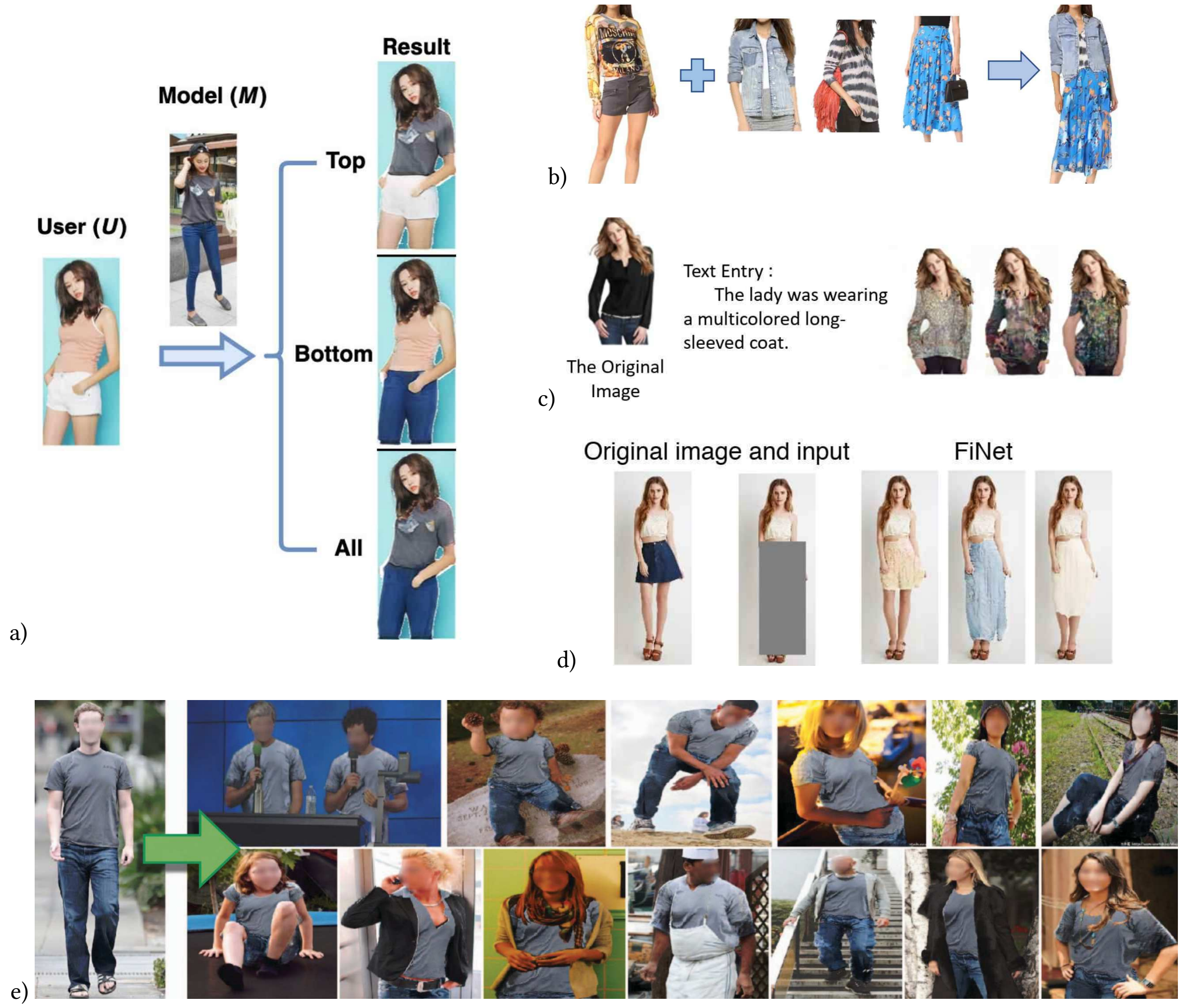

Figure 22: a) Street-Model controllable try-on [196] b) Multiple items try-on [216] c) Text-guided try-on [184] d) Fashion image inpainting [190] e) Image to video try-on [199]

### *2.4.2 2D Modeling*

2D modeling is also image-based, with one main distinction. Here, the input image completely changes, and the output is a new 2-dimensional model of the original image. 2D modeling can be the synthesis of the same image from a different angle, pose-guided image synthesis of a person with a different pose (known as pose transformation), or even a graphical/cartoon model or an avatar of the input image. Most systems we label as 2D modeling are pose-guided try-on systems. There also exist pose-transfer systems that might not focus on fashion; however, their proposed methods can be implemented in 2D modeling try-on systems.

Table 7: Articles Related to 2D Modeling

| No | Article Reference | Year | Technical Keywords/Claimed Results | Application Notes |
|---|---|---|---|---|
| 1 | Ma [227] | 2017 | $PG^2$, U-Net-like, Conditional DCGAN, 3.090 IS | Human pose transfer |
| 2 | Raj [186] | 2018 | Dual-path U-Net, DRAGAN, SwapNet, 3.04 IS | Pose-guided, Swap clothes |
| 3 | Esser [228] | 2018 | Conditional variational U-Net, VGG19, 3.087 IS | Shape/Pose-guided person generator |
| 4 | Siarohin [229] | 2018 | Deformable skip connections, Def-GAN, 3.439 IS | Human pose transfer |
| 5 | Ma [230] | 2018 | Disentangled representation, U-Net, $PG^2$, 3.228 IS | Foreground/Background/Pose manipulation |
| 6 | Zanfir [189] | 2018 | 3D pose & shape, DMHS, SMPL, HAS, Layout warping | Appearance transfer, Model-Model, Pose |
| 7 | Qian [231] | 2018 | Pose-Normalization, PN-GAN | Pose transfer, For ReID task, CCTV |
| 8 | Dong [232] | 2018 | Soft-Gated Warping-GAN, Parsing, 3.314 IS | Human pose transfer |
| 9 | Si [233] | 2018 | Hourglass, CRF-RNN, 3D joints, 0.72 SSIM | Human pose transfer, In-lab images |
| 10 | Balakrishnan [234] | 2018 | Segmentation, U-Net, 0.863 SSIM | Pose transfer, Has problems with the face |
| 11 | Pumarola [235] | 2018 | Unsupervised, Conditioned bidirectional GAN, 2.97 IS | Human pose transfer, Unsupervised |
| 12 | Yildirim [194] | 2019 | Modified conditional style GAN, 9.63 FID | Try-on multiple items, Pose-guided |
| 13 | Hsieh [236] | 2019 | Conditional GAN, Fit-Me, Four stages, 3.336 IS | Pose-guided try-on |
| 14 | Dong [237] | 2019 | Flow-navigated warping, FW-GAN, CGAN, 6.57 FID | Video virtual try-on |
| 15 | Dong [238] | 2019 | MG-VTON, Conditional parsing, Warp-GAN, 3.368 IS | Multi-pose guided virtual try-on |
| 16 | Zheng [239] | 2019 | Attentive bidirectional GAN, 0.7541 SSIM | Pose-guided try-on |
| 17 | Han [202] | 2019 | Flow-based GAN, ClothFlow, Pyramid Net., 3.88 IS | Human pose transfer |
| 18 | Y. Li [240] | 2019 | Dual-path U-Net, Pixel warping, PatchGAN, 3.338 IS | Human pose transfer |
| 19 | Albahar [241] | 2019 | Bi-directional feature transformation, 3.22 IS | Human pose transfer |
| 20 | W. Liu [207] | 2019 | Liquid warping GAN, HMR, 3.419 SSIM | Pose/Outfit transfer, Motion, In-lab images |
| 21 | Sun [242] | 2019 | Bi-directional Conv. LSTM, U-Net, 3.006 IS | Human pose transfer |
| 22 | Zhu [243] | 2019 | HPE, Pose attention, VGG-19, 3.209 IS | Human pose transfer, Shape consistency |
| 23 | Song [244] | 2019 | Semantic parsing transformation, E2E, 3.441 IS | Human pose transfer, Unsupervised |
| 24 | Zhou [245] | 2019 | Multi-modal, LSTM, Attentional upsampling, 4.209 IS | Text-guided pose & appearance transfer |
| 25 | Jeong [212] | 2020 | Graphonomy, SEAN, ResBlK, SEBlK | Try-on and also human pose transfer |
| 26 | Hsieh [246] | 2020 | Parsing, CIHP, pix2pix, cGAN, U-Net, 3.191 IS | Pose-guided try-on, Good detail generation |
| 27 | Tsunashima [247] | 2020 | Unsupervised, Disentangled representation, UVIRT | Try-on using consumer clothing images |
| 28 | Men [225] | 2020 | Attribute-decomposed GAN, U-Net, AdaIN, 3.364 IS | Controllable person image generator, Pose |
| 29 | Ren [248] | 2020 | Differentiable global-flow local-attention, 10.573 FID | Human pose transfer |
| 30 | Huang [249] | 2020 | Appearance-aware pose stylizer, AdaNorm, 3.295 IS | Human pose transfer |
| 31 | Wang [250] | 2020 | Spatially adaptive instance Norm., SPAdaIn ResBlock | 3D Mesh pose transfer |
| 32 | J. Liu [251] | 2020 | Dense local descriptors, Autoencoder, 0.959 SSIM | Human pose transfer, Try-on, Video |
| 33 | Gao [252] | 2020 | Semantic-aware attentive transfer, LGR, 3.855 IS | Recapture, Pose+Body shape+Style, Video |
| 34 | K. Li [253] | 2020 | Pose-guided non-local attention, PoNA, GAN, 3.338 IS | Human pose transfer |
| 35 | Kuppa [254] | 2021 | DensePose, CP-VTON, GELU, ReLU, U-Net | Video virtual try-on |
| 36 | Chou [165] | 2021 | Template-free, TF-TIS, Parsing, cGAN, 3.077 IS | Pose-guided try-on, Good detail generation |
| 37 | Lewis [166] | 2021 | Pose-conditioned StyleGAN2, VOGUE, AdaIN, 32.21 FID | High-resolution pose transfer |

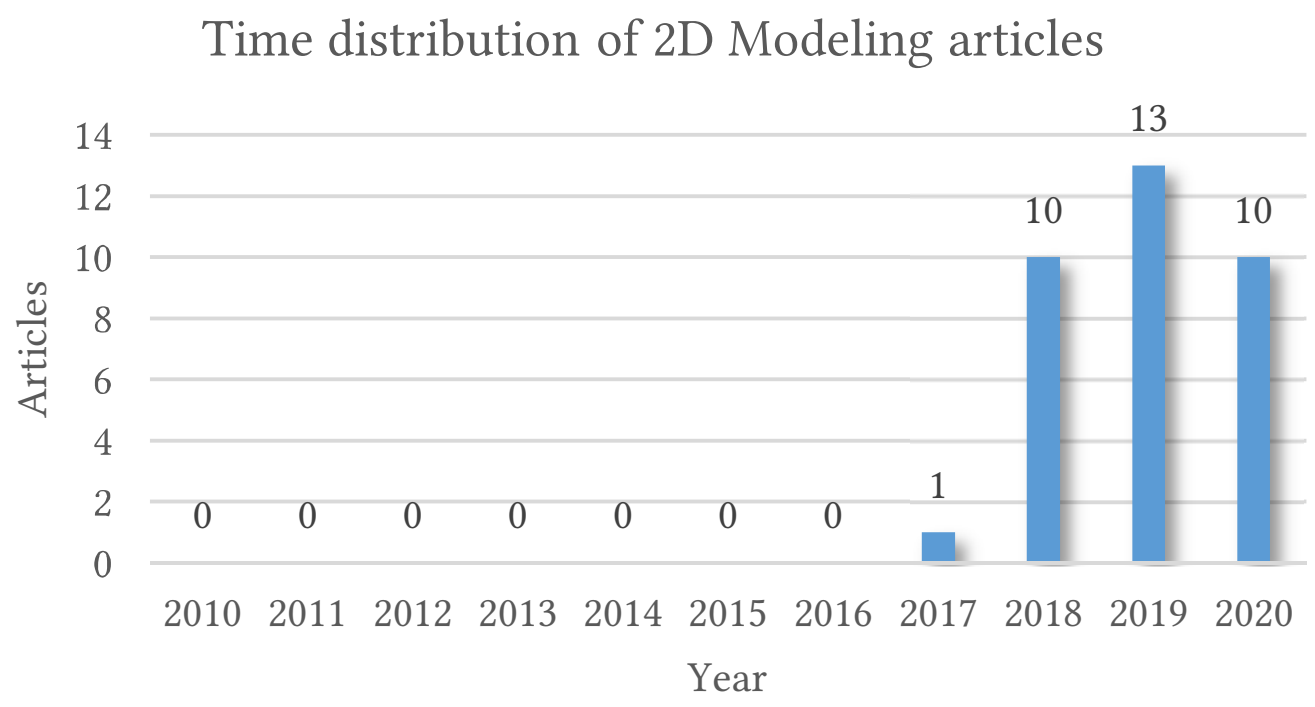

Figure 23: Time Analysis of 2D Modeling Articles

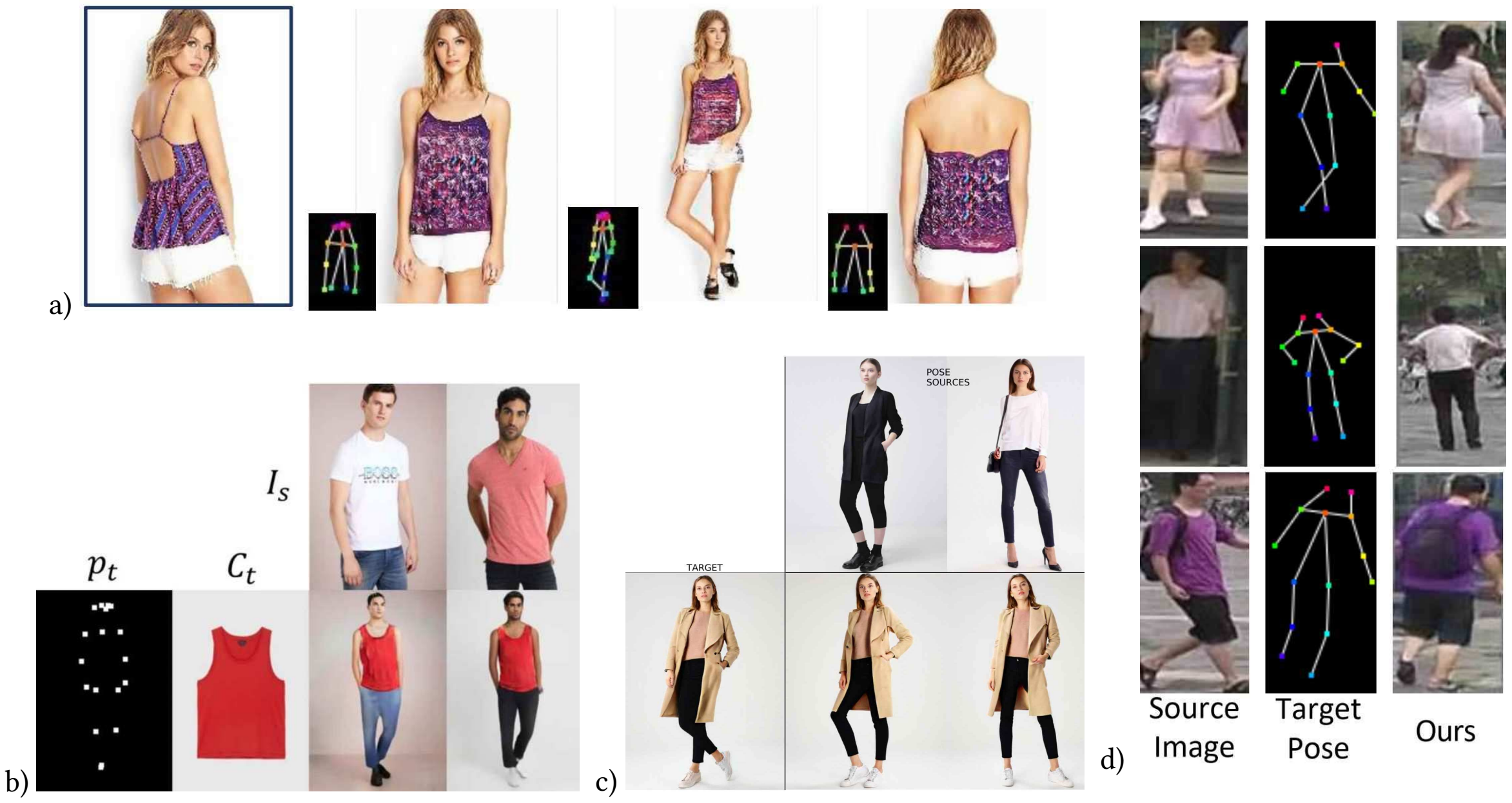

Figure 24: a) Pose transfer on Model images [249] b) Pose-guided try-on [246] c) Model-Model pose transfer [194] d) Pose transfer in the wild [248]

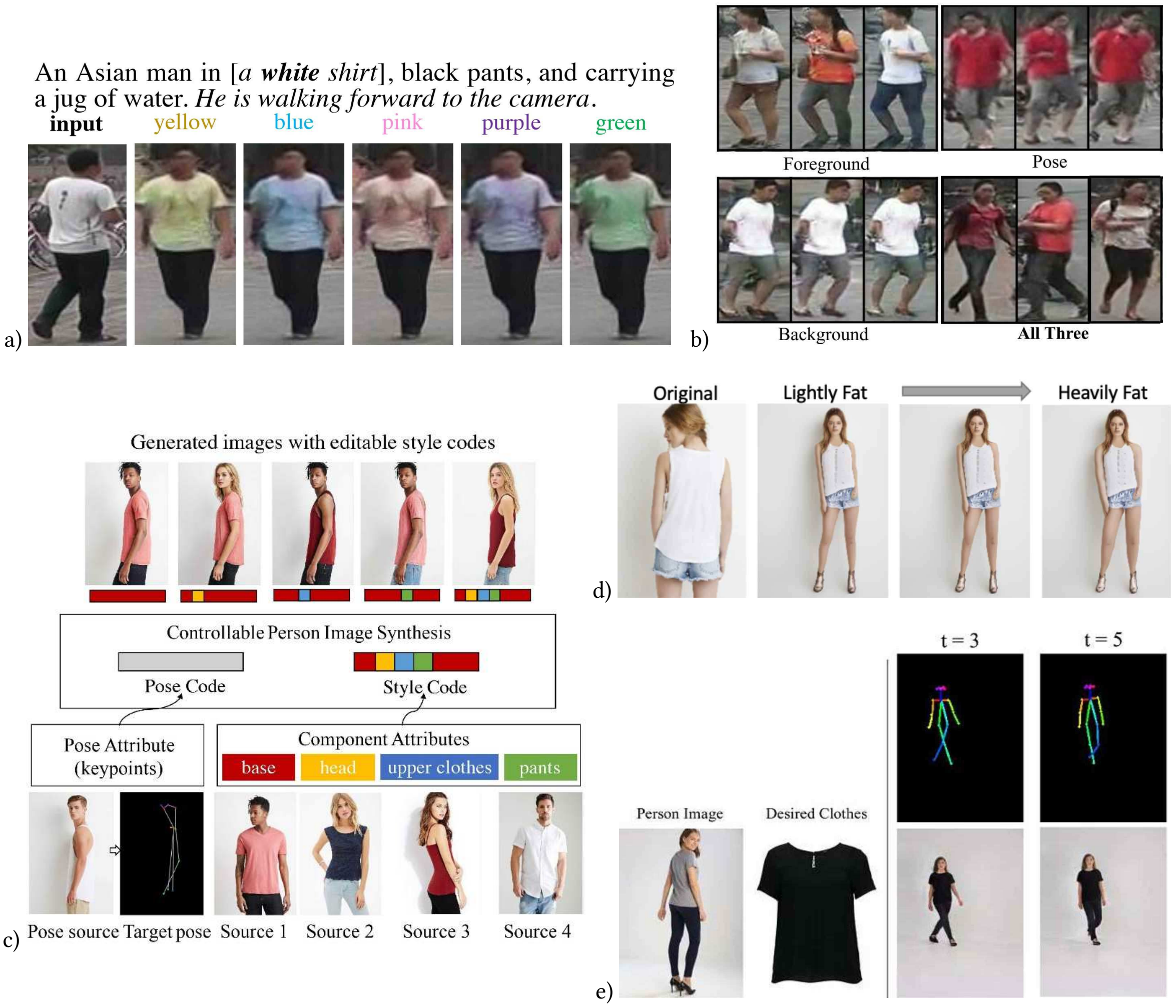

Figure 25: a) Text-guided appearance and view transfer [245] b) Disentangled person image generation [230] c) Controllable person image generation [225] d) Image recapture (Pose, body shape, style) [252] e) Virtual try-on video [237]

#### 2.4.3 3D Modeling

3D modeling applications include try-on and also 3D garment modeling. Some studies focus on 3D body scanning and geometry or texture modeling of garments, while others focus on 3D modeling and physical simulation from a 2D input image. 3D modeling of clothed humans is a highly active field, not only for fashion purposes but also partly due to its applications in the huge movie and animation industry and gaming graphics. We use dual keywords (Input-Output) in the "Application Notes" column of Table 8 to categorize systems whenever possible. For example, "Image-3D Body" shows a system that generates 3D body models from 2D images.

Table 8: Articles Related to 3D Modeling

| No | Article Reference | Year | Technical Keywords | Application Notes |
|---|---|---|---|---|
| 1 | D'Apuzzo [255] | 2007 | 3D scanning, Overview of methods, Companies | 3D scanning in the apparel industry |
| 2 | D'Apuzzo [256] | 2009 | 3D scanning, Overview of methods, Application | 3D scanning in the apparel industry |
| 3 | Y. Liu [257] | 2010 | Approaches, Human modeling, Garment design, Draping | Survey, CAD methods in 3D garment design |
| 4 | Robson [258] | 2011 | Context-aware, Geometric modeling | Sketch-3D garment |
| 5 | Yuan [259] | 2011 | Face coordinates, Occlusion, Mixed reality | 3D virtual glasses try-on |
| 6 | Niswar [260] | 2011 | Face reconstruction, Tracking, Fitting | Glasses, Head & face 3D modeling |
| 7 | Miguel [261] | 2012 | Data-driven estimation, Nonlinear models | Cloth simulation models, Deformations |
| 8 | Guan [262] | 2012 | DRAPE, Physics-based, Deformation gradients | Dress 3D bodies, Any shape, Any pose |
| 9 | Yasseen [263] | 2013 | Quad meshes, Discrete coons patches | Sketch-3D garment, Design |
| 10 | X. Chen [264] | 2013 | Deformable model, SCAPE, IK algorithm | Image-3D body, Clothed/Naked |
| 11 | Zhou [265] | 2013 | Pose estimation, Body shape, Shape-from-shading | Full body image-3D garment |
| 12 | Ionescu [266] | 2013 | KNN, KRR, Regression, Fourier embedding, LinKRR | Benchmark, Image-Multiple 3D bodies |
| 13 | S. Wang [267] | 2014 | Parametric feature model, 3D scanner, Key points, VHM | Human 3D model, Feature-based |
| 14 | Y. Yang [268] | 2014 | RGB-D camera, Object tracking, ICP, PCA | 3D Footwear try-on, Live video |
| 15 | X. Chen [89] | 2015 | Depth camera, KinectFusion, 3D deformable template | Image-3D garment, In-lab, Kinect |
| 16 | Guan [269] | 2016 | Review, A section on 3D Try-on, Various methods | Apparel virtual try-on with CAD system |
| 17 | S. Yang [182] | 2016 | Physics-based, Parameter Est., Semantic parsing, Shape | Image-3D garment, Single image |
| 18 | Pons-Moll [270] | 2017 | ClothCap, Multi-part 3D model, Segmentation | 4D Clothing capture & retargeting, Motion |
| 19 | Daněřek [271] | 2017 | Mocap sequence, CNN, 3D vertex displacement | Image-3D garment, Single image |
| 20 | Zhang [272] | 2017 | 3D scans, Parametric model, SMPL, Single-frame | Clothed 3D scans-Naked 3D body, Accurate |
| 21 | Hong [273] | 2018 | 3D Scanning, Rule-based model, Sensory descriptors | 3D-to-2D garment design, Scoliosis |
| 22 | Daanen [274] | 2018 | Measures, Devices, Processing, Virtual fit | An overview on 3D body scanning |
| 23 | T. Wang [275] | 2018 | Shared shape space, PCA, Siamese network | Sketch-3D garment, Design, Retarget |
| 24 | Lähner [276] | 2018 | cGAN, DeepWrinkles, Pose Est., PCA, LSTM | 4D scans-3D garment, Accurate, Realistic |
| 25 | Alldieck [277] | 2018 | Pose reconstruction, Unposed canonical frame | Video-3D clothed body |
| 26 | Bhatnagar [278] | 2019 | Multi-garment, MGN, SMPL, Vertex based PCA | 3D try-on from multiple video frames (8) |
| 27 | Gundogdu [279] | 2019 | Two-stream, GarNet, Spatial transformer network, MLP | 3D draping, 100x faster than physics-based |
| 28 | Xu [159] | 2019 | Multi-task learning, JFNet, ASPP, ResNet101, MLS | Two images-3D garment |
| 29 | Pumarola [280] | 2019 | CNN, Spherical area-preserving Param., GimNet | Image-3D clothed body |
| 30 | Lazova [281] | 2019 | DensePose, Segmentation, SMPL UV-space, cGAN | Image-3D avatar, Fully-textured |
| 31 | Alldieck [282] | 2019 | Canonical T-pose, SMPL+D, Octopus, CNN | Image-3D clothed body |
| 32 | Saito [283] | 2019 | Pixel-aligned implicit function, PIFu, Marching cube | Image-3D clothed body, High-resolution |
| 33 | Natsume [284] | 2019 | Silhouette-based, SiCloPe, Greedy sampling, GAN | Image-3D clothed body |
| 34 | Yu [285] | 2019 | SimulCap, DoubleFusion, Force-based mass-spring | Single-view 3D performance capture |
| 35 | Alldieck [286] | 2019 | SMPL, UV map, Pix2Pix, U-Net, PatchGAN | Image-3D body geometry, Detailed |
| 36 | Sattar [287] | 2019 | SMPL, Joint-based, Multi-photo optimization | Multiple images-3D body |
| 37 | Santesteban [288] | 2019 | Learning-based, Physics-based, RNN, MLP, PSD | 3D try-on clothing animation, Wrinkles, Fit |
| 38 | Shin [289] | 2019 | Deep image matting, DCNN, Recursive Conv. Net. | Realistic garment rendering for 3D try-on |
| 39 | T. Wang [290] | 2019 | Intrinsic garment space, MLP, Motion-driven Autoenc. | Garment authoring, Animation |
| 40 | W. Liu [207] | 2019 | SMPL, HMR, NMR, Liquid warping GAN | In-lab image-3D Mesh, Motion transfer |
| 41 | Huang [291] | 2020 | Semantic deformation field, Stacked hourglass, U-Net | Image-Animatable 3D body |
| 42 | Zhu [292] | 2020 | Dataset, Pose Est., Graph CNN, SMPL, Pixel2Mesh | Image-3D garment, Bench., DeepFashion3D |
| 43 | Jin [293] | 2020 | CNN, Pixel-based framework, PCA, Deformations | Pose-3D garment, Pose-guided 3D clothing |
| 44 | Vidaurre [294] | 2020 | Parametric 3D mesh, SMPL, Graph CNN, U-Net | Parametric try-on, Garment/Body/Material |
| 45 | Mir [295] | 2020 | Silhouette shape, U-Net, Pix2Surf, SMPL, GrabCut | Item image-3D clothed body, Texture |
| 46 | Caliskan [296] | 2020 | Multiple-view loss, CNN, Stacked hourglass | Image-3D body |
| 47 | Minar [213] | 2020 | CloTH-VTON, SMPL, U-Net, Shape-context matching | Item image-3D garment, Image-based try-on |
| 48 | Tiwari [297] | 2020 | SizerNet, 3D parsing, SMPL+G, Encoder-decoder | Size sensitive 3D clothing, 3D parser |

| No | Article Reference | Year | Technical Keywords | Application Notes |
|---|---|---|---|---|
| 49 | Patel [298] | 2020 | TailorNet, MLP, SMPL, PCA, Narrow bandwidth kernel | 3D clothed body, Pose/Shape/Style, Detailed |
| 50 | Ju [299] | 2020 | Cusick's drape, Two-stream NN, CLO3D simulator | Image (Static drape)-Cloth simulation |
| 51 | Ali [300] | 2020 | FoldMatch, Physics-based, Wrinkle-vector field | Garment fitting onto 3D scans, Accurate |
| 52 | Shen [301] | 2020 | cGAN, Non-rigid ICP, Voronoi diagram, SMPL | Sewing pattern image+3D body-3D garment |
| 53 | Li [302] | 2020 | Morphing salient points, MPII, Garment mapping | In-home 3D try-on App. |
| 54 | Ma [303] | 2020 | SMPL, Graph-CNN, mesh, Conditional MeshVAE-GAN | 3D clothed body-3D scans |
| 55 | Bertiche [304] | 2020 | Learning-based, PSD, Physics-based simulation, MLP | Unsupervised garment pose space Deform. |
| 56 | Minar [222] | 2020 | Pose Est., SMPL, TPS, Shape-context matching | Item image-3D garment, Image-based try-on |
| 57 | Jiang [305] | 2020 | Layered garment Rep., SMPL, MLP, PCA, ResNet-18, GAT | Image-3D clothed body |
| 58 | Bertiche [306] | 2020 | SMPL, Conditional variational Auto-enc., Graph Conv. | Image-3D clothed body |
| 59 | Su [307] | 2020 | UV-position map with mask, ParamNet, CNN | 3D scan clothed body shape & style editing |
| 60 | Bertiche [308] | 2020 | Local geometric descriptors, Graph Conv., MLP | Skinning, Deformation, Animation |
| 61 | Gundogdu [309] | 2020 | Physics-based, Curvature loss, GarNet++, MLP, KNN | 3D Clothing draping |
| 62 | Revkov [310] | 2020 | FITTIN™, 3D model of foot/shoe, Smartphone | Online 3D shoe try-on |
| 63 | Saito [311] | 2020 | Multi-level, Trainable, PIFu, CNN, MLP, pix2pixHD | Image- Detailed High-Res. 3D model |
| 64 | L. Chen [312] | 2021 | Temporally & spatially consistent Deform., CNN | Deep deformation detail synthesis |
| 65 | Wu [313] | 2021 | Sensitivity-based distance, Taylor expansion, LBS | Real-time 3D clothing, Virtual agents |
| 66 | Yoon [314] | 2021 | Semi-supervised, Neural clothes retargeting, CRNet | Image-3D garment, Retarget |
| 67 | Minar [226] | 2021 | CloTH-VTON+, SMPL, TPS, Shape-context matching | Item image-3D garment, Image-based try-on |

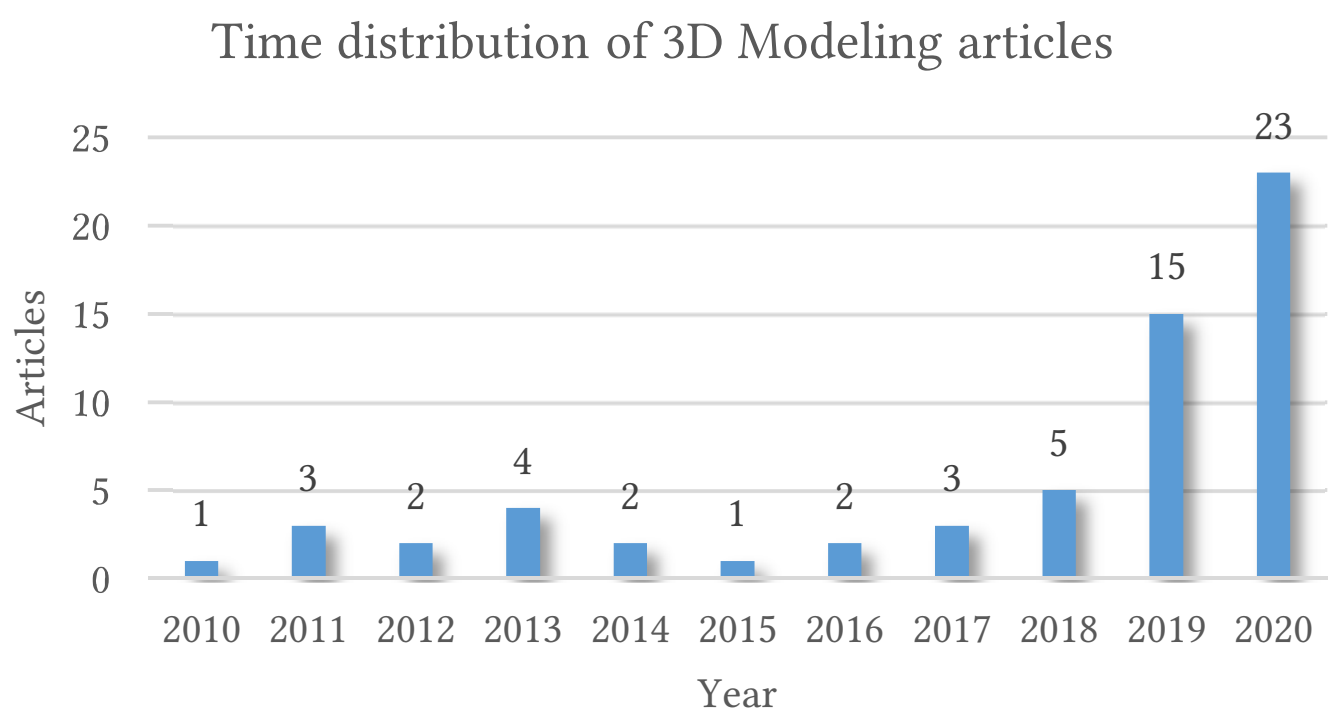

Figure 26: Time Analysis of 3D Modeling Articles

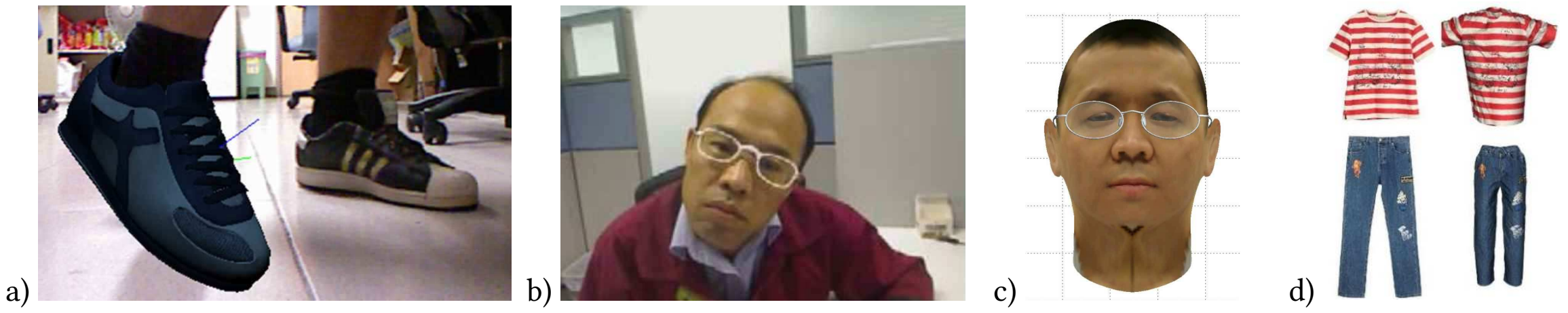

Figure 27: a) Mixed reality shoe try-on [268] b) Mixed reality glasses try-on [259] c) 3D modeling for glasses virtual try-on [260] d) Item image-3D garment [281]

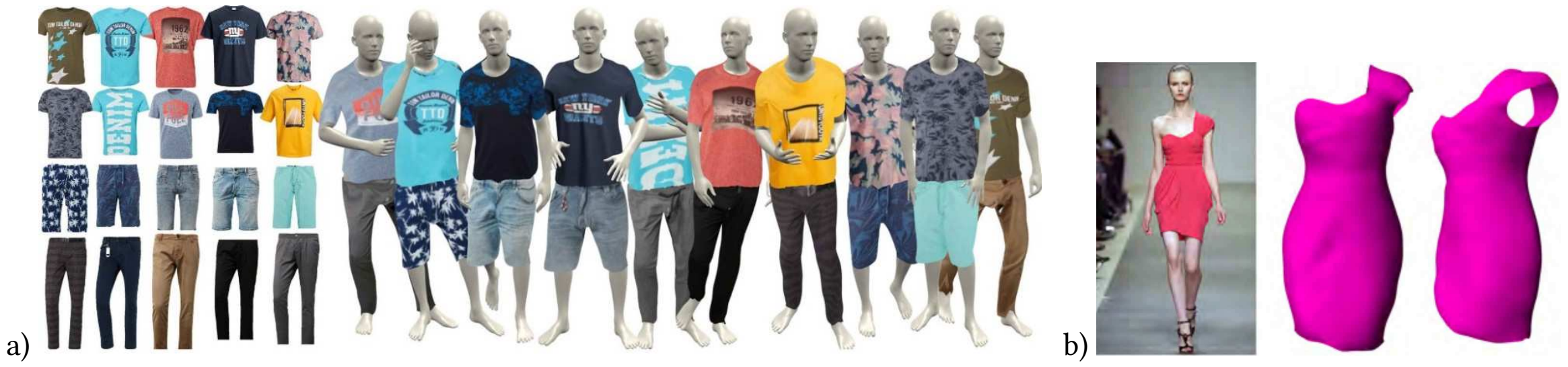

Figure 28: a) Transfer image to 3D models [295] b) Model-3D garment [265]

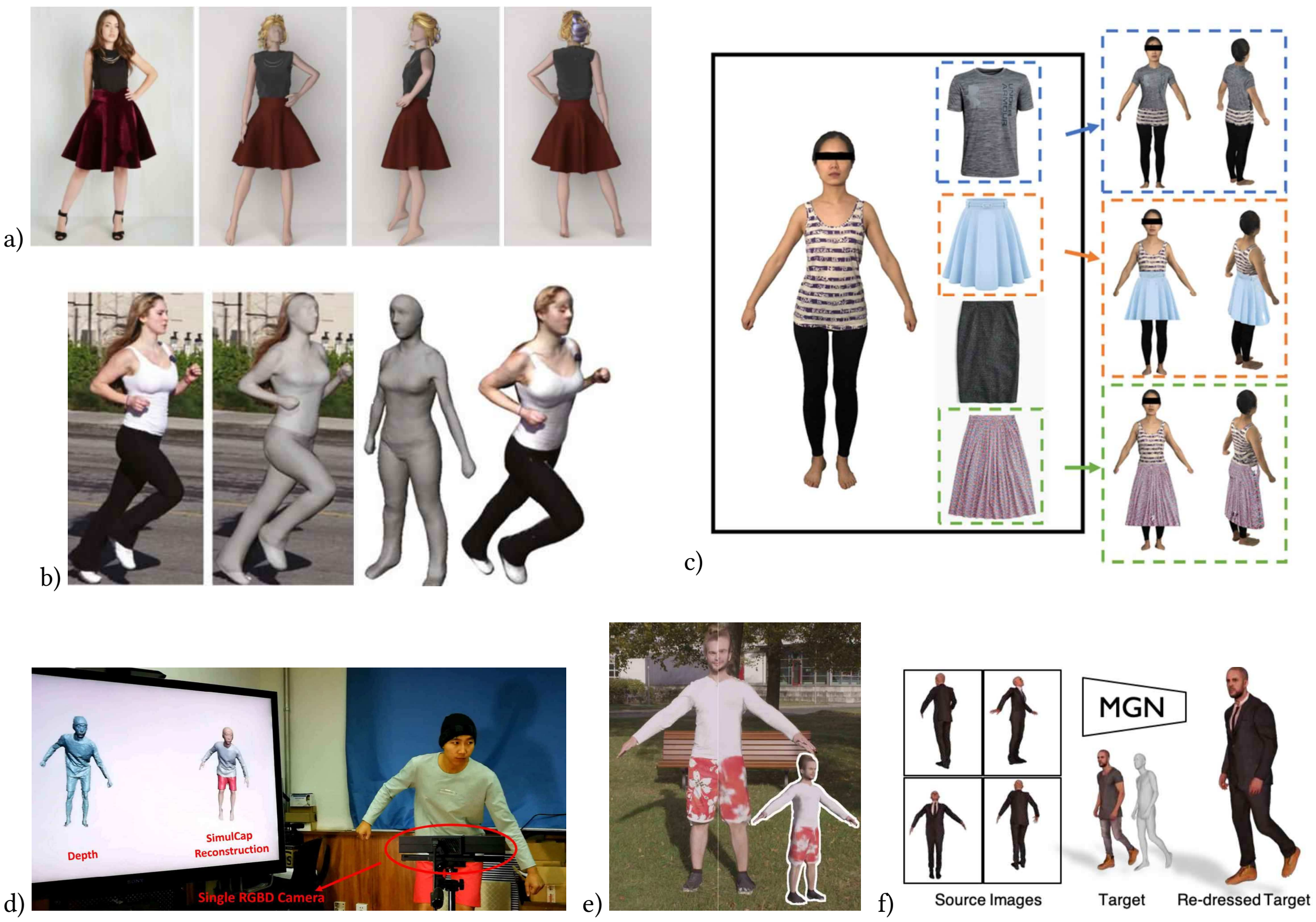

Figure 29: a) 3D garment recovery from single image [182] b) Street-3D clothed model [264] c) 3D virtual try-on [302] d) Clothed 3D model capturing using a single RGBD camera [285] e) Video-3D model [277] f) 3D garment retargeting [278]

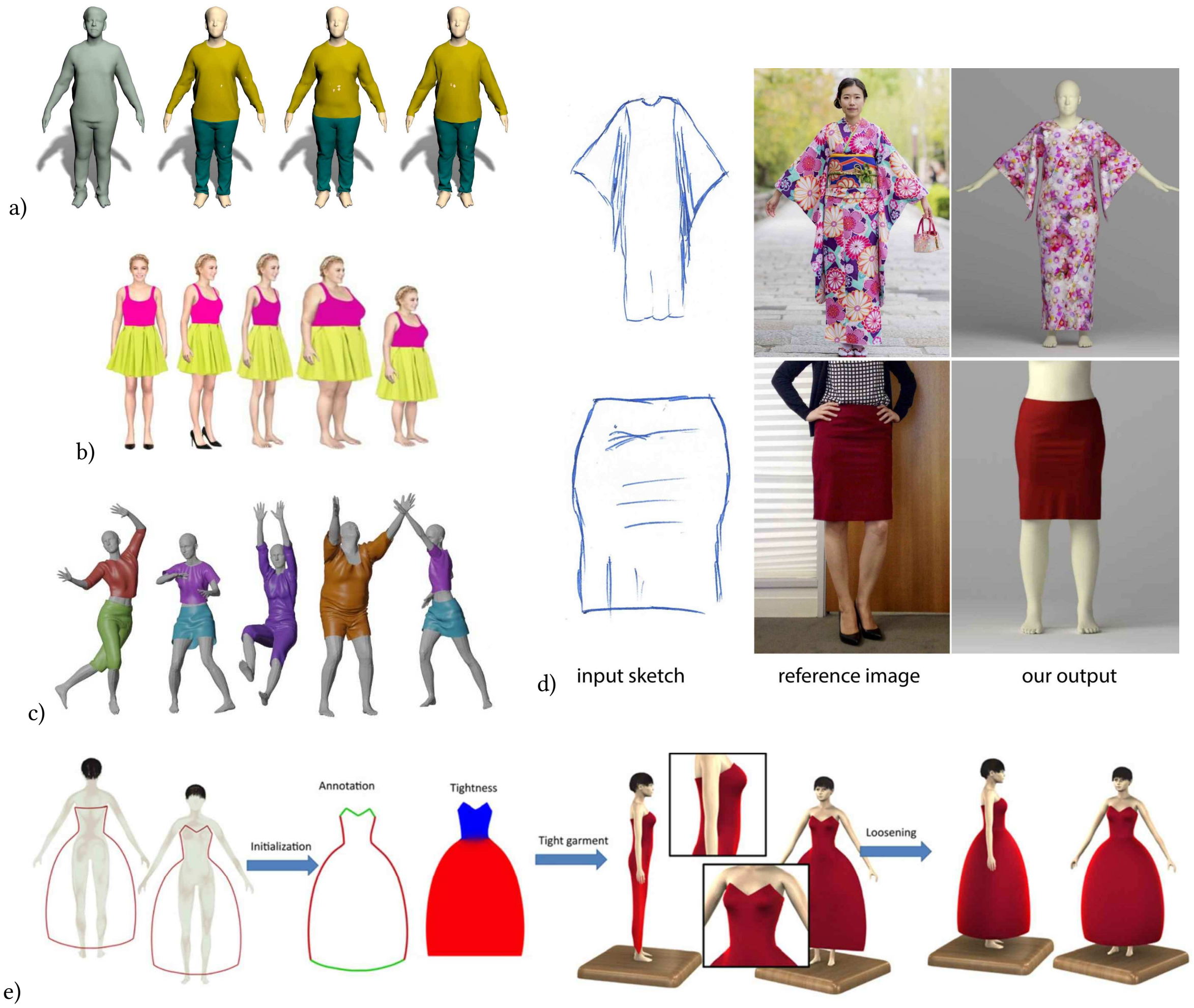

Figure 30: a) Size-sensitive 3D modeling (large, medium, and X-large from left to right) [297] b) 3D body shapes for personalized virtual try-on [289] c) 3D garment animation in different sizes and motions [304] d) Sketch-3D garment [275] e) Sketch-3D garment with regards to tightness [258]

#### *2.4.4 Size & Fit*

Choosing the right clothing size and the best fit is one of the main reasons fitting rooms exist in the real world. Technology needs to provide solutions to this problem in online apparel shops. Studies link the perception of clothing fit in women to their body image in their mind, and they showed that a good choice of clothing fit helps improve confidence and cover perceived flaws [315]. Thus, we need systems to predict the size of clothing for different individuals and fashion articles' fit based on the user's body shape and size. One of the main approaches is 3D body scanning. Digitization technologies can measure specific body parts or even generate full body measurements in seconds. Thus, we can also consider various 3D modeling methods in sec 2.4.3 for this application.

Table 9: Articles Related to Size & Fit

| No | Article Reference | Year | Technical Keywords/Claimed Results | Application Notes |
|---|---|---|---|---|
| 1 | D'Apuzzo [255] | 2007 | Measurement devices, Methods, 3D scanning | 3D scanning in the apparel industry |
| 2 | Mpampa [316] | 2010 | Statistical analysis, Iterative, LS, Classification | Sizing systems, Mass customization |
| 3 | Apeagyei [317] | 2010 | 3D scanner, CAD, Measurement extraction profile | 3D scanning, Body measurement, Methods |
| 4 | Gaur [318] | 2014 | Graph, Multi-node multi-state, Bag-of-features, SVM | Aesthetics assessment of fashion images |
| 5 | Abdulla [319] | 2017 | Gradient boosting classification, Word2vec, 81.28% Acc | Size recommendation, E-commerce |
| 6 | Hidayati [320] | 2018 | BoVW, Auxiliary visual words, Affinity propagation | Fashion Recom. for personal body shape |
| 7 | Daanen [274] | 2018 | Measures, Devices, Processing, Sizing | An overview on 3D body scanning |
| 8 | Guigourès [321] | 2018 | Hierarchical Bayesian model, Mean-field approximation | Size recommendation |
| 9 | Sheikh [322] | 2019 | Content-collaborative, SFNet, Siamese, 76.0% Acc | Size & fit prediction, E-commerce |
| 10 | Du [323] | 2019 | Agglomerative clustering, Character-LSTM, QP | Automatic size normalization |
| 11 | Sattar [287] | 2019 | SMPL, 3D model, Multi-photo optimization | Clothing preference based on body shape |
| 12 | Dong [324] | 2019 | PCW-DC, Bayesian personalized ranking, MLP | Personalized capsule wardrobe, Body shape |
| 13 | Yan [325] | 2020 | SMPL, Non-rigid iterative closest point, Non-Lin. SVR | Measurements from 3D body scans |
| 14 | Tiwari [297] | 2020 | SizerNet, 3D parsing, SMPL+G, Encoder-decoder | Size sensitive 3D clothing |
| 15 | Hsiao [326] | 2020 | Visual body-aware embedding, 3D mesh, SMPL, HMD | Fashion Recom. for personal body shape |
| 16 | Yang [327] | 2020 | Multi-view, Semantic Seg., PSPNet, Clustering, Matching | Girth measurement, Stereo images, Design |
| 17 | Li [302] | 2020 | 3D scanner, MPII, Salient anthropometric points | In-home 3D fitting room App. |
| 18 | Hu [328] | 2020 | Body PointNet, MLP, OBB Norm., Symmetric chamfer | Body shape under clothing from a 3D scan |
| 19 | Wolff [329] | 2021 | Structure sensor, Isometric bending, Var. surface cutting | 3D Custom fit garment design, Pose |
| 20 | Foysal [330] | 2021 | SURF, Box filter, Bag-of-features, k-NN, CNN, 87.50% Acc | Body shape detection, Smartphone App. |

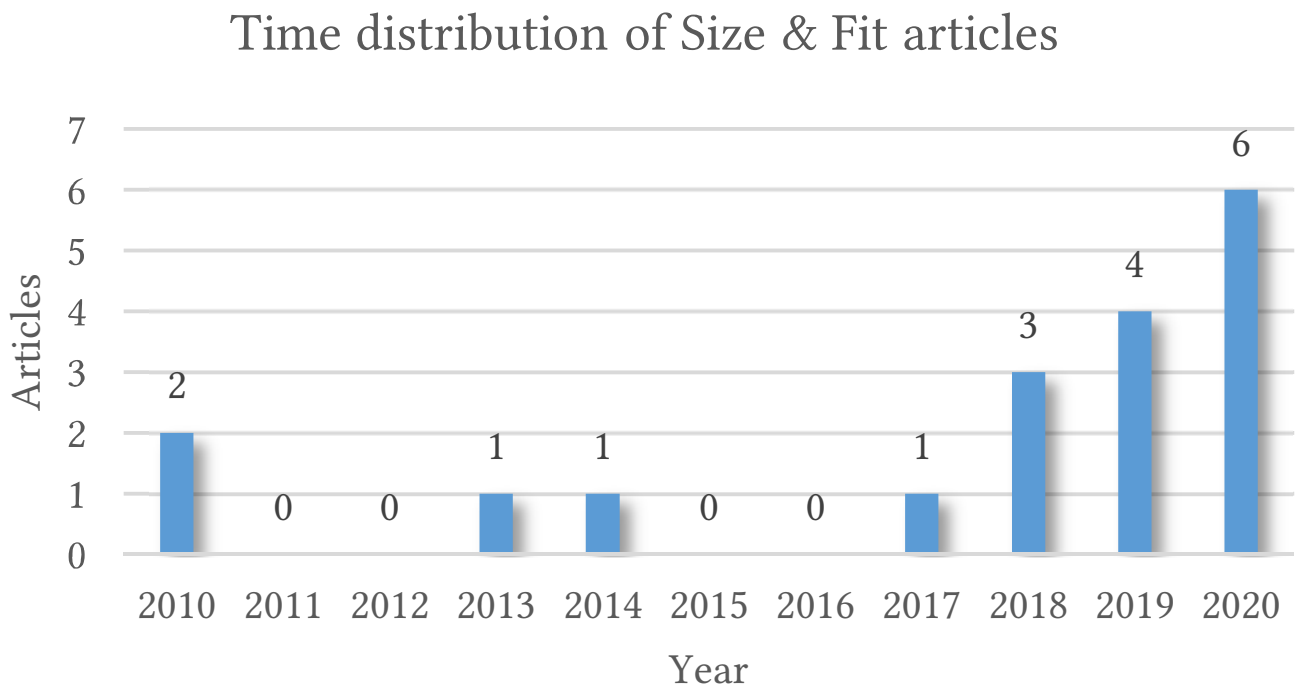

Figure 31: Time Analysis of Size & Fit Articles

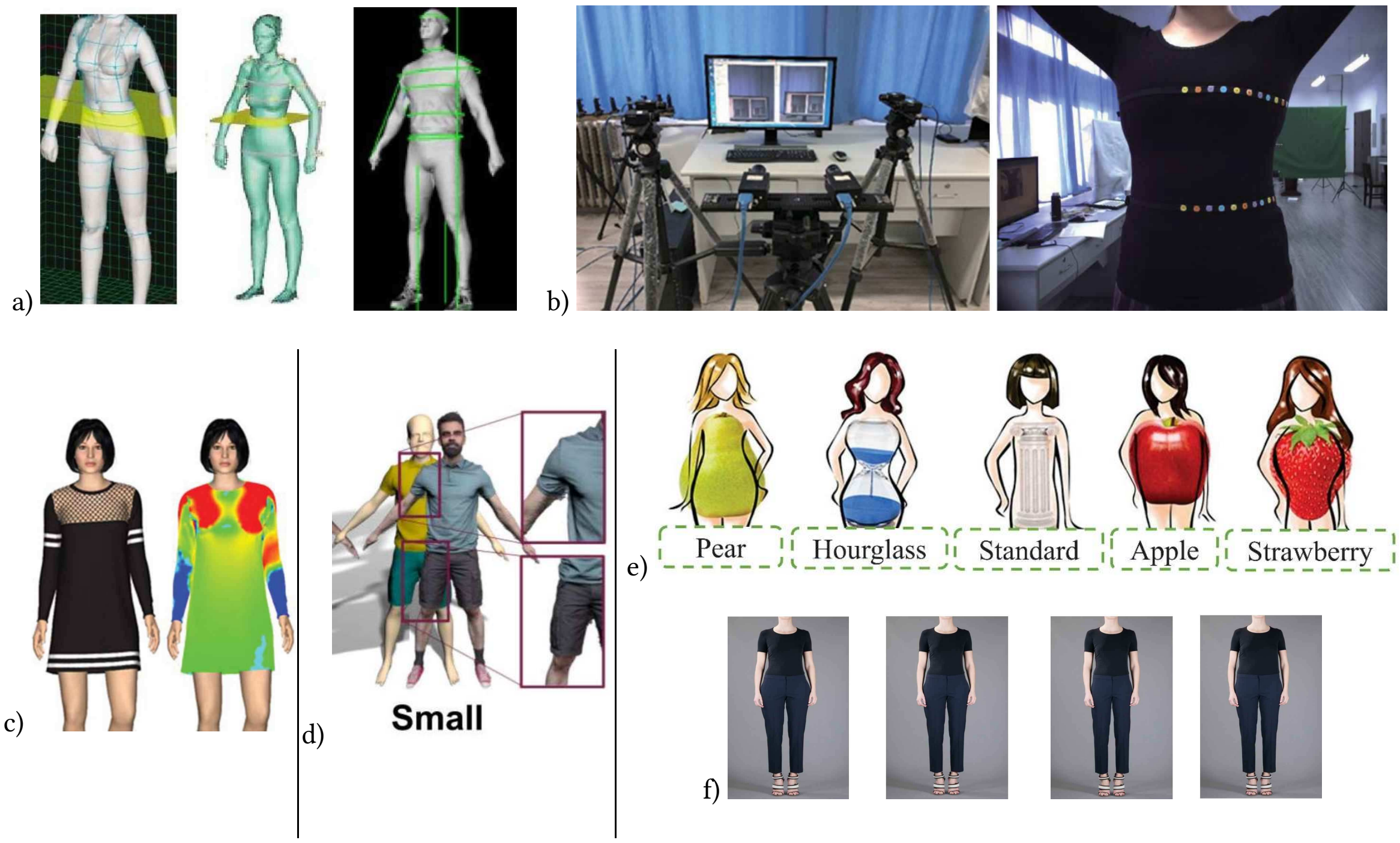

Figure 32: a) Automatic body sizes extraction [255] b) Settings for girth measurement [327] c) Tightness visualization in 3D modeling [274] d) Size-sensitive 3D models [297] e,f) Body shapes [324] and their corresponding 3D models [318]

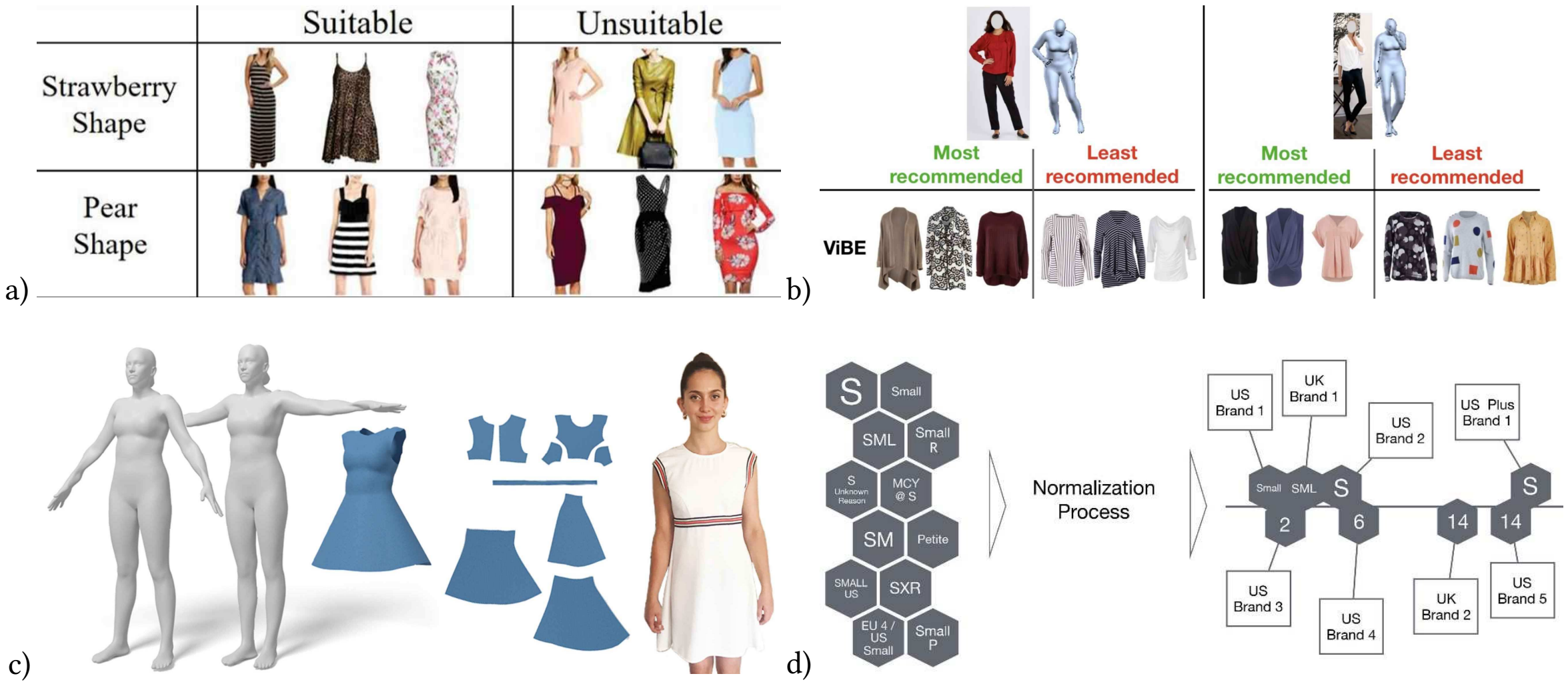

Figure 33: a) Recommendations for different body shapes [324] b) Body shape estimation and personalized recommendation [326] c) 3D custom fit garment design [329] d) Automated fashion size normalization [323]

### *2.4.5 Magic Mirror*

They were introduced in 2009 by the name of Smart Mirror [10] as a retrieval system and recommender. Then again, in 2016, by the name of Magic Mirror [331], this time as a virtual fashion consultant. In fact, magic mirrors can be much more than that. They can be the ultimate implementation of all fashion applications, including analysis, recommendation, try-on, synthesis, etc., combined with an interactive system and augmented reality. Our focus here is on studies that explain system architectures and shine some light on the hardware and schemes needed to implement magic mirrors.

Table 10: Articles Related to Magic Mirror

| No | Article Reference | Year | Technical Keywords | Application Notes |
|---|---|---|---|---|
| 1 | Chao [10] | 2009 | Smart mirror, Classic, HOG, LBP, Web camera | Style recommender |
| 2 | Yuan [259] | 2011 | Face coordinates, Occlusion, AR, Two-way mirror | Glasses try-on, Mixed reality |
| 3 | Yang [268] | 2014 | RGB-D camera, Object tracking, ICP, PCA | 3D Footwear try-on, Mixed reality, Live |
| 4 | Liu [331] | 2016 | Kinect, Bimodal deep autoencoder, Correlative label | Magic mirror, Fashion compatibility |
| 5 | Fu [332] | 2017 | Kinect, User preference, Genetic algorithm, Fashion trend | Demo, Fashion compatibility |
| 6 | Asiroglu [12] | 2019 | Embedded Linux system, Haar-cascade, DoG, CNN | Recommender, Color, Gender, Pattern |
| 7 | Boardman [333] | 2020 | Augmented reality, Virtual reality, Case study | Review, AR & VR in fashion retail |

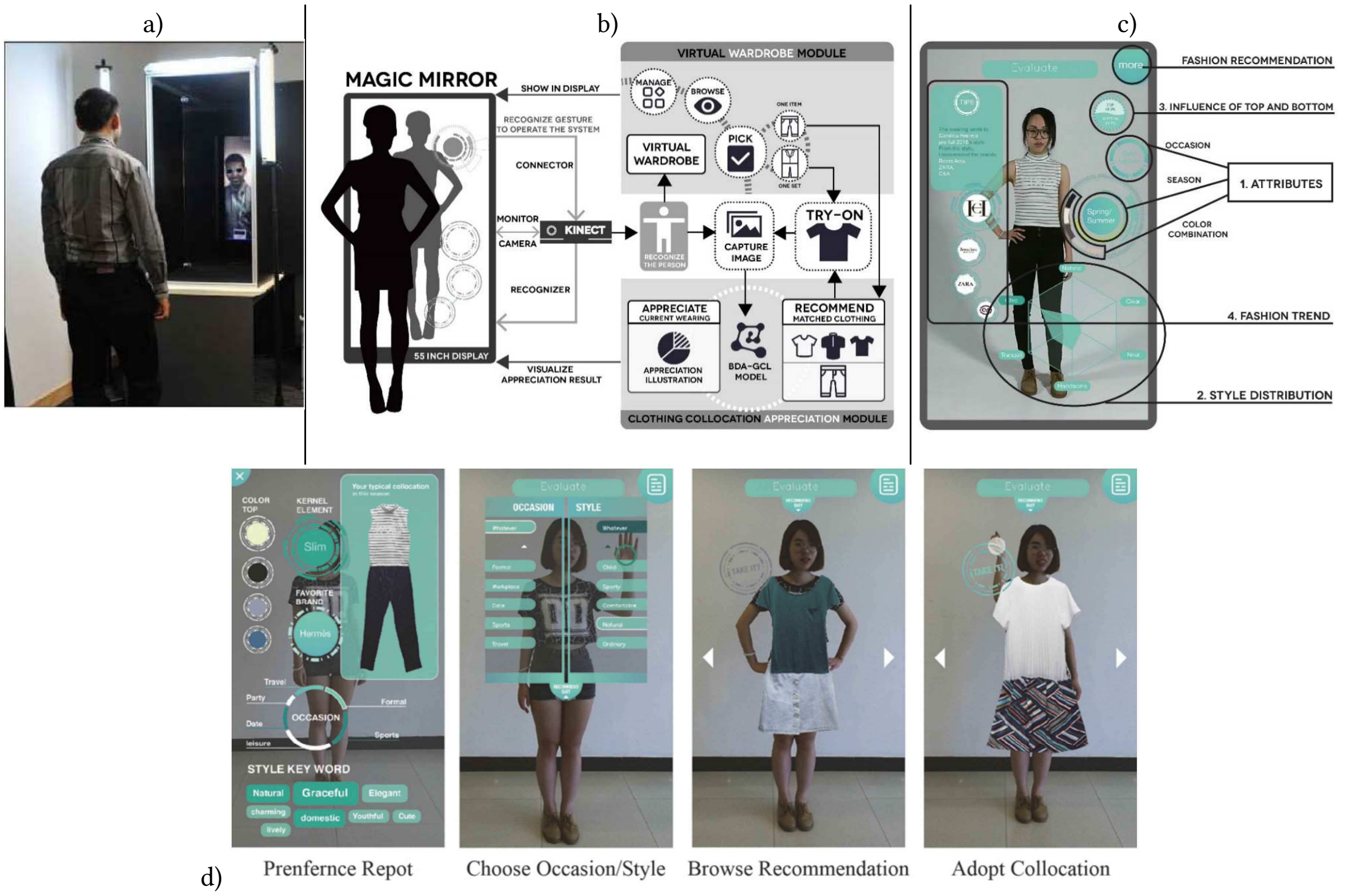

Figure 34: a) Mixed reality glasses try-on [259] b,c,d) Magic mirror schemes [331], [332]

### 2.5 Fashion Synthesis

Fashion synthesis emphasizes synthesizing new fashion item images and designs from scratch. Bear in mind that try-on applications also synthesize images, but with a different purpose. In try-on applications, the focus is on the human presence in the photo, while in fashion synthesis, the main focus is on creating novel and unseen fashion items. Comprehensive research on consumer responses to GAN-generated fashion images can be found in [334]. Various approaches exist, and different inputs are used to guide the system to generate the final output. We try to report each system's output in the "Application Notes" column of Table 11 or use dual keywords (Input-Output) wherever possible. For example, "Model-Item" shows that the system takes one fashion image with a human model and generates the fashion article's catalog image. Bear in mind that image synthesis is not the final goal of all synthesis systems, and some try to generate designs and ideas leading to the physical production of fashion items.

Table 11: Articles Related to Fashion Synthesis

| No | Article Reference | Year | Technical Keywords/Claimed Results | Application Notes |
|---|---|---|---|---|
| 1 | J. Wang [335] | 2011 | Garment surface style, 3D pattern style, Silhouette curves | Mass customization, 3D garment design |
| 2 | Mok [336] | 2013 | Interactive genetic algorithm, Parametric design, SDA | Sketch, Fashion design support system |
| 3 | Yoo [337] | 2016 | GAN, Pixel-level domain transfer, MSE, 0.21 C-SSIM | Model-Item, Item-Model, Street photos |
| 4 | J-Y. Zhu [338] | 2016 | Manifold approximation, DCGAN, AlexNet, L-BFGS-B | Attr. Manipulation, Shape, Color |
| 5 | Kang [339] | 2017 | Siamese CNN, CNN-F, GAN, LSGAN, BPR, 7.652 IS | User preference-Recommended items |
| 6 | S. Zhu [184] | 2017 | FashionGAN, Segmentation, Compositional mapping | (Text+Model)-Model, Attr. manipulation |
| 7 | Date [340] | 2017 | Segmentation, VGG-19, SVM, LBFGS | Multiple items-Item, Style transfer |
| 8 | A. Yu [341] | 2017 | Semantic Jitter, Attribute2Image, CVAE, MLP | Synthesize varying Attr. Images, Shoes |
| 9 | Lassner [342] | 2017 | ClothNet, VAE, CVAE, Image-to- image Trans. | Seg. body map-Person image, Pose, Color |
| 10 | Hong [273] | 2018 | 3D Scanning, Rule-based model, Sensory descriptors | 3D-to-2D garment design, Scoliosis |
| 11 | Kato [343] | 2018 | Case study, DeepWear, DNN, DCGAN | Models-New models, Design, Fashion show |
| 12 | J. Zhu [344] | 2018 | CNN, Nonnegative matrix factorization, VAE | Popular items-New items, Design |
| 13 | Rostamzadeh [345] | 2018 | Progressive GAN, StackGAN-v1/v2, 7.91 IS | Text-Model, Fashion-Gen, Challenge |
| 14 | Yang [346] | 2018 | Siamese, BPR, GAN, SE-Net, Inception-V3, 6.823 IS | Generates fashion collocations, Item image |
| 15 | Günel [347] | 2018 | Feature-wise linear modulation, GAN, fastText, 2.58 IS | (Text+Model)-Model, Attr. manipulation |
| 16 | Esser [228] | 2018 | Conditional variational U-Net, VGG19, 3.087 IS | Sketch-Image, Matching, For human Gen. |
| 17 | Xian [348] | 2018 | TextureGAN, VGG-19, Scribbler, Texture patch | (Sketch+Texture)-Image, Bag, Shoe, Clothes |
| 18 | Ye [52] | 2019 | Semi-supervised GAN, Hard-aware, MR-GAN, 27.28 FID | Sample generation for Hard-aware learning |
| 19 | Han [190] | 2019 | FiNet, Human parser, Encoder-decoder, 36.6% HS | Fashion image inpainting |
| 20 | Ak [349] | 2019 | Enhanced AttnGAN, Feature-wise Lin. Modul., 4.77 IS | Text-Model, Semantically consistent |
| 21 | Hsiao [350] | 2019 | Fashion++, Semantic segmentation, cGAN, VAE | Minimal edits for outfit improvement |
| 22 | Kumar [351] | 2019 | Conditional distribution, c+GAN, DCT, Faster R-CNN | Upper body image-Compatible bottom |
| 23 | Yildirim [194] | 2019 | Modified conditional style GAN, 9.63 FID | Multiple items-Clothed Model, High-Res. |
| 24 | Lin [352] | 2019 | Co-supervision, FARM, Variational transformer, DCNN | (Item+Text)-Compatible item, Recom. |
| 25 | Ping [353] | 2019 | Attribute-aware, Multi-objective AttGAN | Attribute manipulation, Color |
| 26 | Ravi [354] | 2019 | VGG-19, Style transfer CNN, Super Resolution SRCNN | (Silhouette+Pattern)-Item, Style transfer |
| 27 | Albahar [241] | 2019 | Bi-directional feature transformation, 3.22 IS | (Sketch+Texture)-Item, Image translation |
| 28 | C. Yu [355] | 2019 | Personalization, VGG-16, LSGAN, Siamese, 4.262 IS | (Item+User preference)-Compatible item |
| 29 | Ak [356] | 2019 | CNN, AMGAN, Class activation mapping, 79.48% mAcc | Attribute manipulation |
| 30 | H. Zhang [131] | 2020 | Category-supervised GAN, cGAN, Patch-GAN, pix2pix | Model-Item, Take-off, For retrieval |
| 31 | Chen [357] | 2020 | TailorGAN, Encoder-decoder, Self-attention mask | (Ref. item+Attr. item)-Item, Attr. editing |
| 32 | Kınlı [358] | 2020 | Dilated partial Conv., U-Net-like, Self-attention, CNN | Inpainting, Irregular holes, Benchmark |
| 33 | Tango [359] | 2020 | GAN, pix2pix, Minimax game, U-Net, 30.38 FID | Anime character image-Real item, Cosplay |
| 34 | Sarmiento [360] | 2020 | Variational autoencoder, Latent code, User interface | Interactive synthesis, Attr. manipulation |
| 35 | K. Wang [220] | 2020 | Unpaired shape transformer, AdaIN, 61.19 SSIM | Model-Item, Clothing take-off |

| No | Article Reference | Year | Technical Keywords/Claimed Results | Application Notes |
|---|---|---|---|---|
| 36 | Li [361] | 2020 | Bi-colored edge Rep., Residual Conv., cGAN, 4.076 IS | (Sketch+Texture)-Item, Interactive |
| 37 | Dong [362] | 2020 | Adversarial parsing learning, FE-GAN, U-net, 0.938 SSIM | Fashion editing, Sketch, Inpainting |
| 38 | P. Zhang [164] | 2020 | PConvNet, Graphonomy, U-Net, Parsing | Inpainting of fashion model images |
| 39 | Ak [363] | 2020 | e-AttnGAN, LSTM, FiLM-ed ResBlock, 4.98 IS | Text-Model, Semantically consistent |
| 40 | Gu [364] | 2020 | Multi-modal, GAN, PatchGAN, 3.124 IS | (Pose+Text)-Model, Fashion translation |
| 41 | Z. Zhu [365] | 2020 | Semantically multi-modal, GroupDNet, VAE, SPADE | (Parsing map+Where to change)-Model |
| 42 | Zhan [366] | 2020 | Appearance-preserved, PNAPGAN, U-Net, Triplet loss | Street photo-Item, Street2shop generation |
| 43 | Wolff [329] | 2021 | 3D Scans, Pose, Design out of standard size garments | 3D Custom fit garment design |

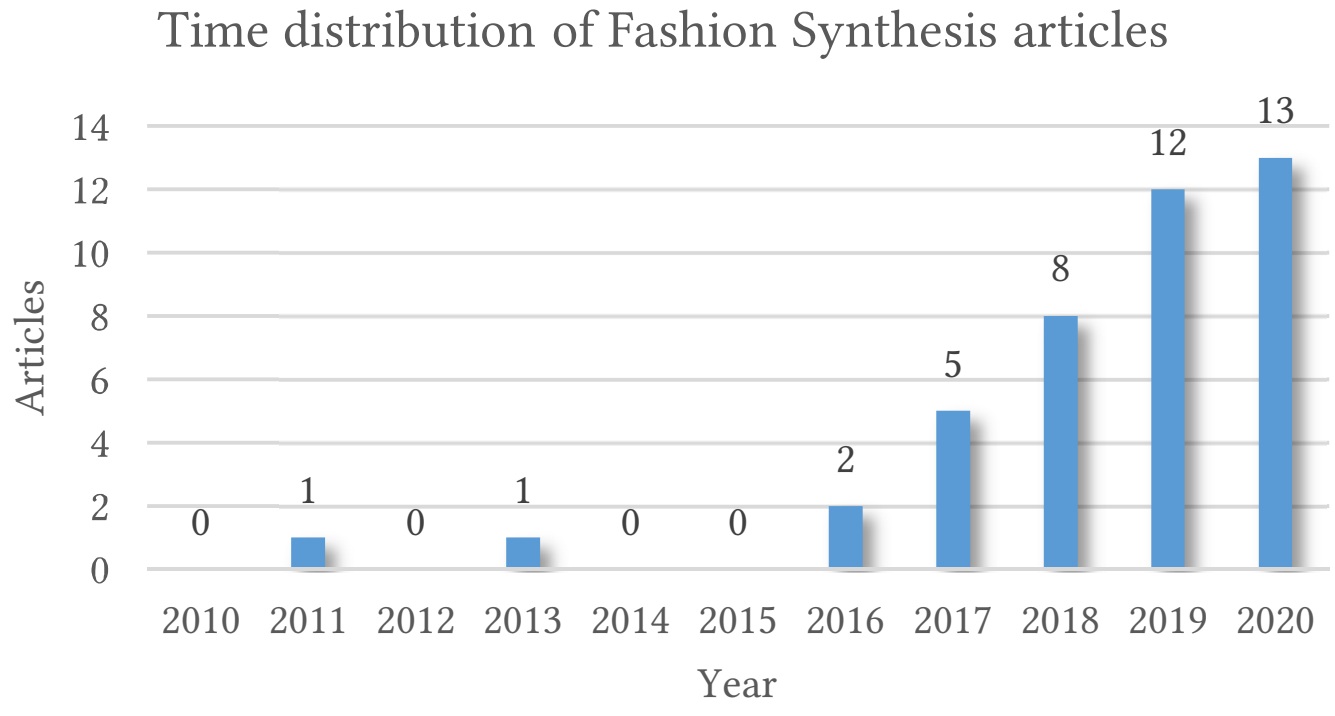

Figure 35: Time Analysis of Fashion Synthesis Articles

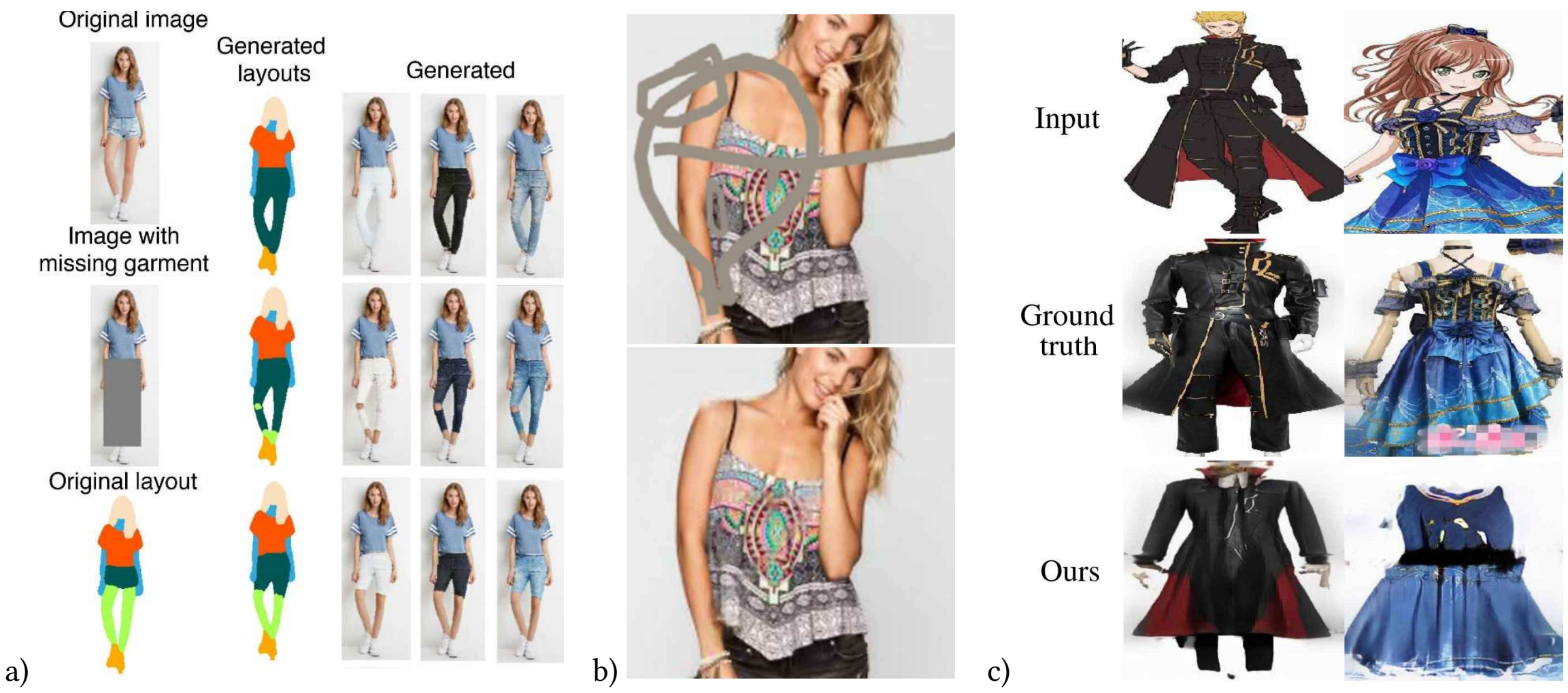

Figure 36: a) Controllable fashion inpainting [190] b) Irregular holes fashion image inpainting [358] c) Anime images to real-life clothing image synthesis [359]

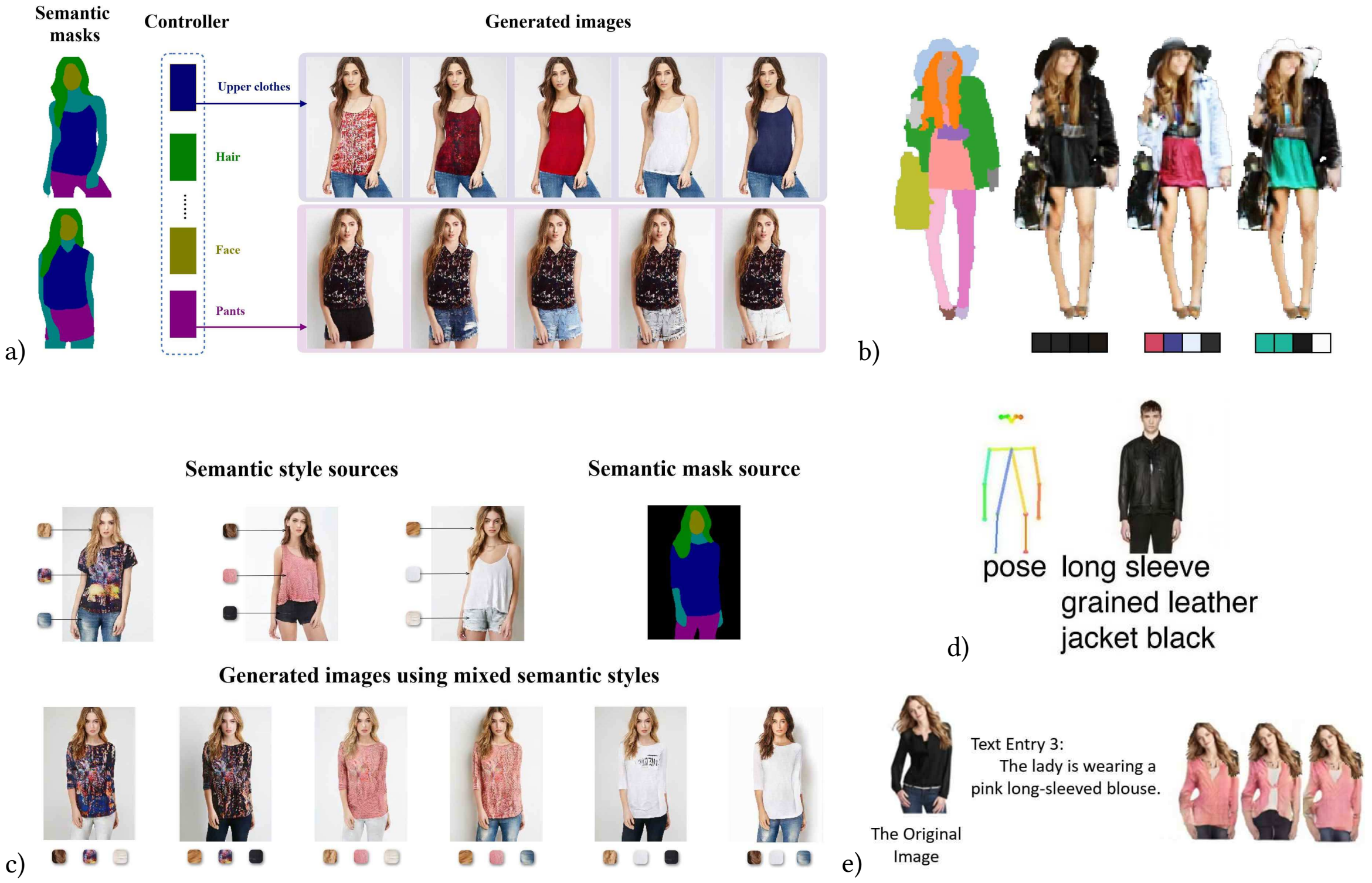

Figure 37: a) Controllable fashion image synthesis [365] b) Color-controlled clothed person image synthesis [342] c) Appearance mixture [365] d) Pose and text-guided Model image synthesis [364] e) Text-guided fashion synthesis [184]

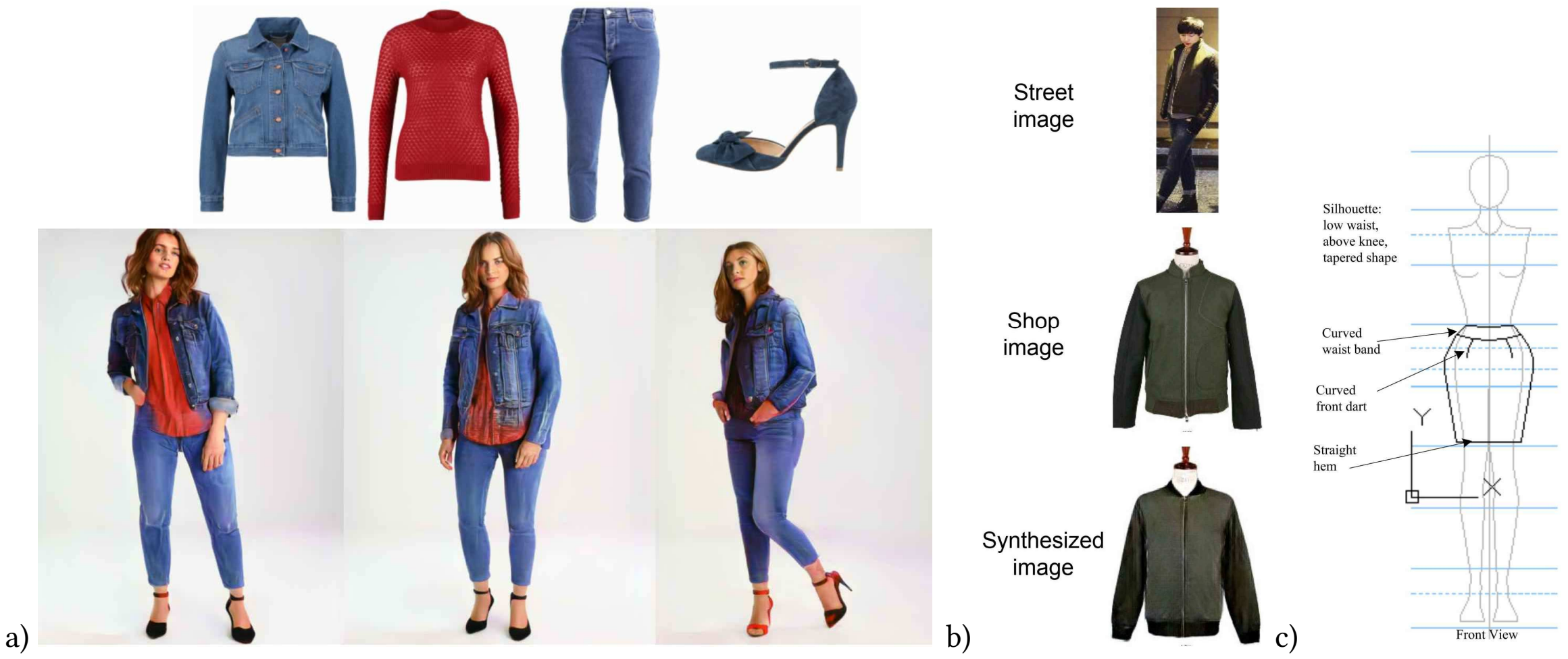

Figure 38: a) Outfit-to-Model high-resolution Model image synthesis [194] b) Street-to-Item synthesis (clothing take-off) [366] c) Design support system for fashion designs [336]

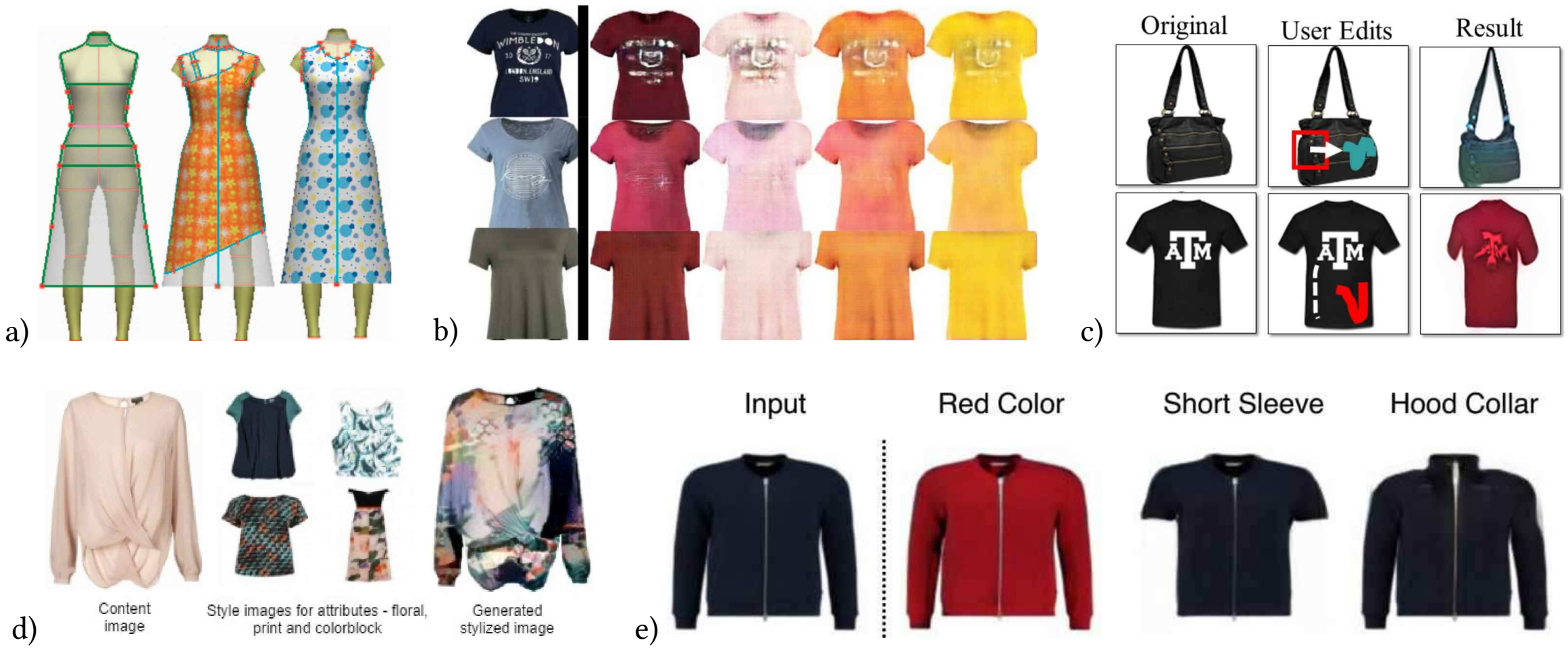

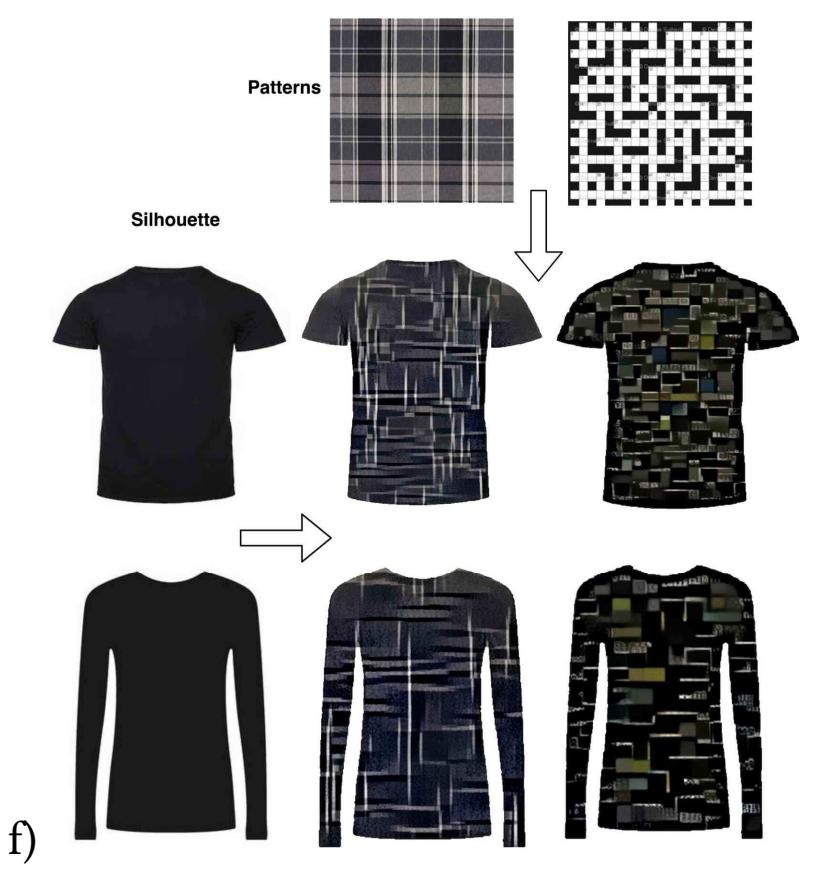

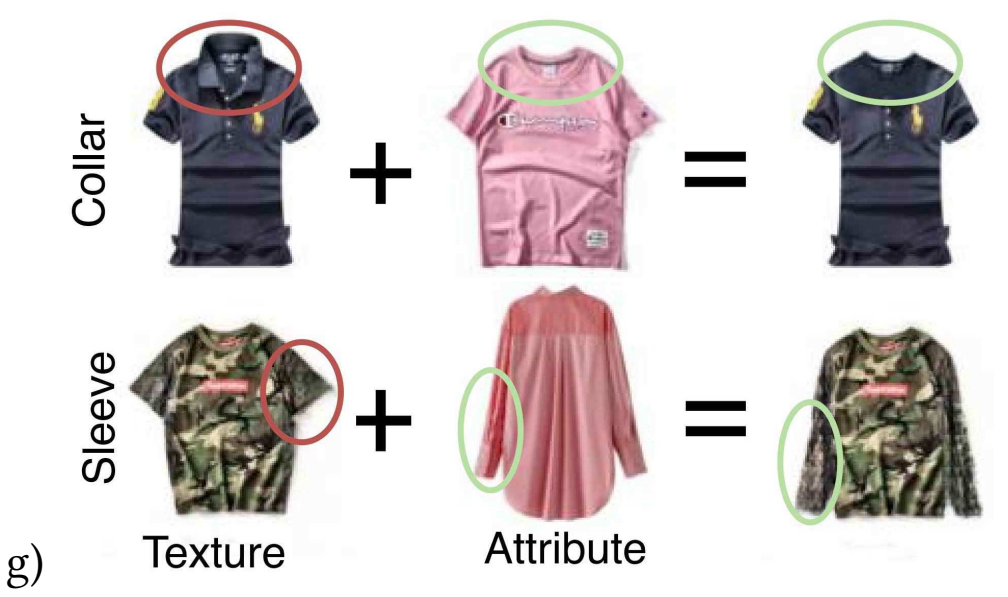

Figure 39: a) 3D garment design and mass personalization [335] b) Item color editing [353] c) User-controlled attribute manipulation [338] d) Style mixing [340] e) Attribute manipulation [356] f) Pattern transfer for fast fashion design [354] g) Controllable attribute editing [357]

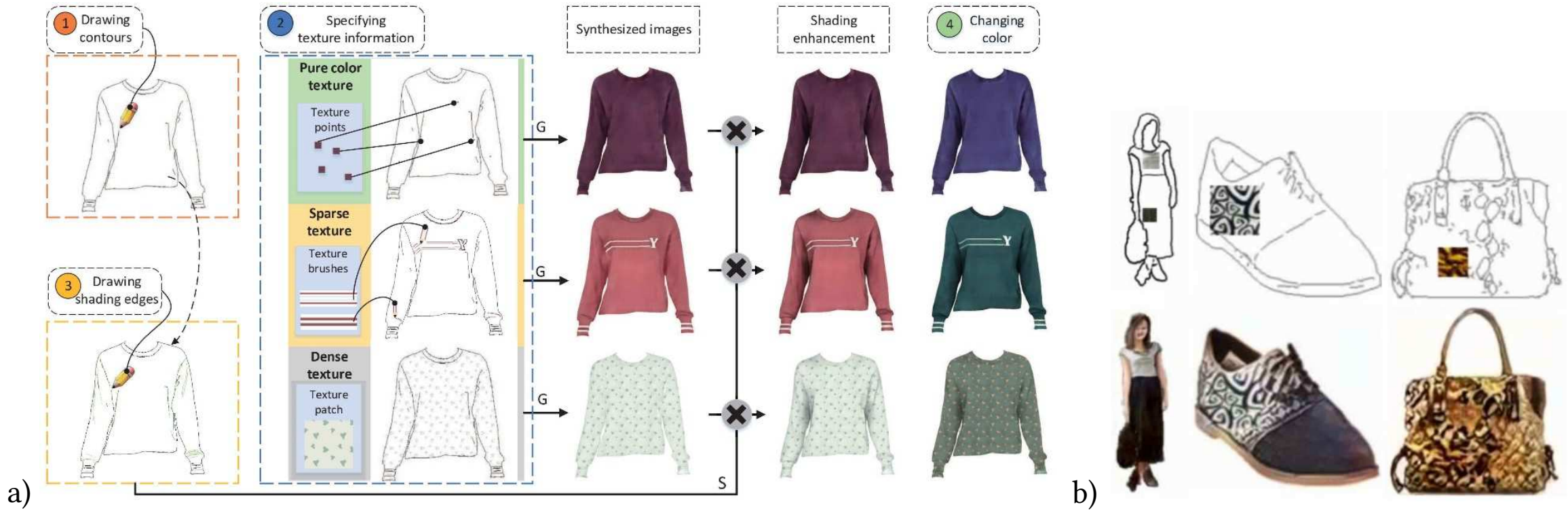

Figure 40: a) Interactive Item image synthesis using sketches [361] b) Texture-guided sketch-to-image synthesis [348]

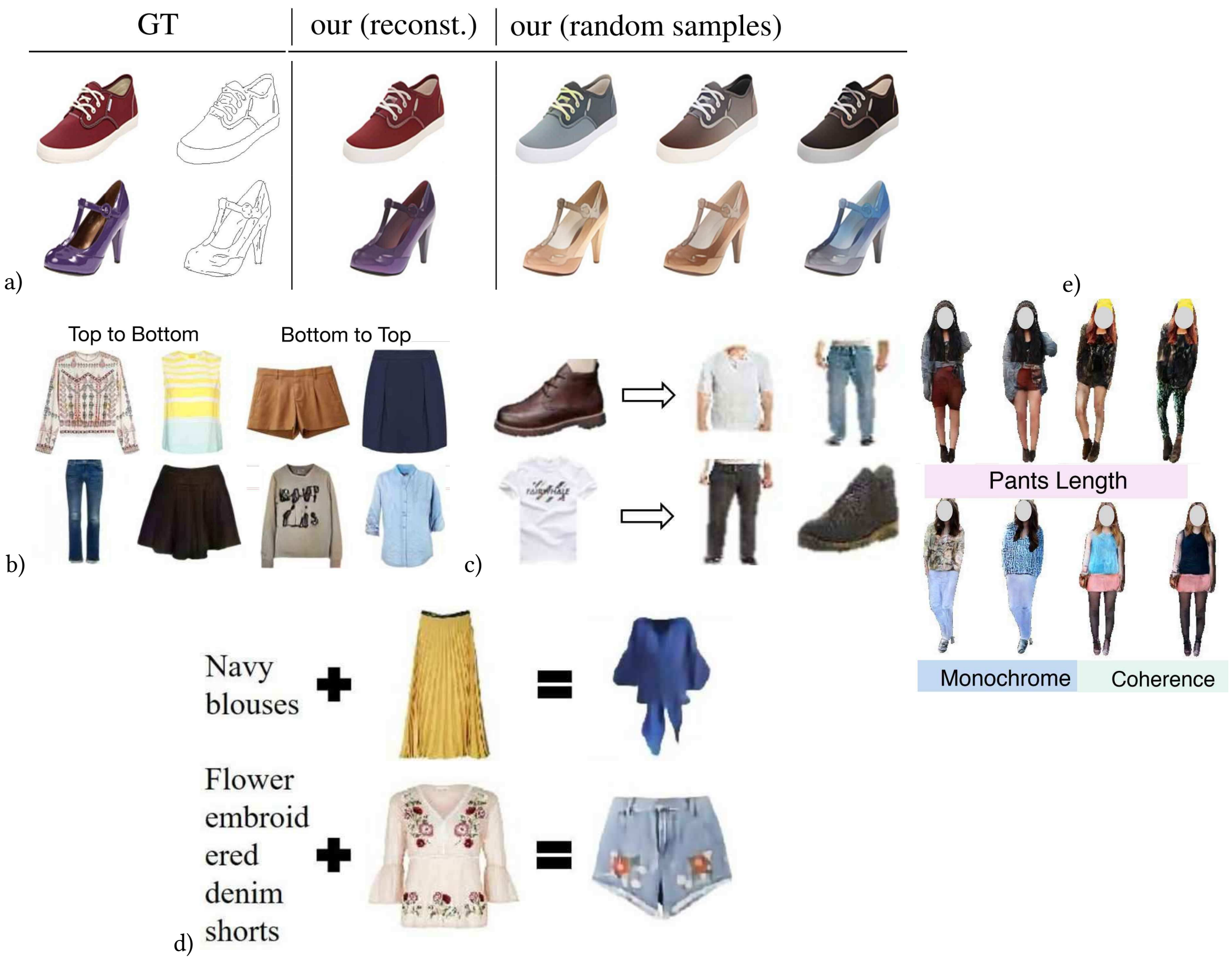

Figure 41: a) Conditional Sketch-to-Item synthesis [228] b) Top/Bottom fashion collocation synthesis [355] c) Top/Bottom/Shoe collocation synthesis [346] d) Text-guided Top/Bottom compatible Item synthesis [352] e) Minimal edits problem, making an outfit more fashionable by minimal adjustments [350]

### 2.6 Fashion Retrieval

This application is devoted to the search and retrieval of fashion items in a database of images. A keyword search cannot always describe the complexities of fashion and target the users' needs; thus, we use content-based retrieval instead to capture the visual features of each item. In this section, by retrieval, we mean 'exact match' retrieval. Note that 'similar item' retrieval also exists, but it has a heavy overlap with recommender systems, and we cover it in the next section. The ultimate goal of retrieval is to find an exact match in the item database for the fashion item query input. These systems fall into three sub-categories: 1) Domain-Specific, 2) Cross-Domain, and 3) Attribute Manipulation.

### *2.6.1 Domain-Specific Retrieval*

These systems are trained to retrieve the exact items on a specific image domain, which means the input and the outputs belong to the same domain. Examples are retrieving a clothing item image with a different angle or model pose in online shops (view-invariant clothing retrieval/in-shop retrieval) or retrieving a person with the same outfit from different CCTV camera images. We try to report each study's work domain in the "Application Notes" section of Table 12 using single keywords (Wild, Street, Shop, Item, Model).

Table 12: Articles Related to Domain-Specific Retrieval

| No | Article Reference | Year | Technical Keywords/Claimed Results | Application Notes |
|---|---|---|---|---|
| 1 | X. Wang [367] | 2011 | Color-based BoW, LS posterior classifier, ~45% Prec. | Wild, Dominant color, Attributes |
| 2 | S. Liu [368] | 2012 | Parts alignment, Classic features, ~67% P@10 | Shop/Street, Upper/Lower body |
| 3 | Fu [369] | 2012 | BoW, Detection, Hierarchical Vocab. tree, ~61.5% P@10 | Shop/Street, Upper/Lower body |
| 4 | Q. Chen [370] | 2013 | Bundled features matching, SIFT, MSER, LWF | Wild images |
| 5 | Yamaguchi [139], [142] | 2013 2014 | Parsing, Style descriptor, KNN, KD-tree | Street, Style retrieval for parsing |
| 6 | Lin [371] | 2015 | AlexNet, Binary code, Hierarchical search, ~59% P@10 | Model, Fast |
| 7 | Vittayakorn [23] | 2015 | Low-level features, Semantic parse, SVM, 73-76% AUC | Runway images |
| 8 | Z. Liu [9] | 2016 | VGG-16, FashionNet, Landmark, ~72% Acc@10 | Shop, Benchmark, DeepFashion |
| 9 | K. Liu [91] | 2016 | VGG-16, Decision fusion, Euclidean Dist., ~19.74% P@10 | Model, View-Invariant, MVC |
| 10 | Sha [92] | 2016 | Classic, Seg., Color matrix, ULBP, PHOG, Fourier, GIST | Model, Attribute-specific retrieval |
| 11 | Sun [94] | 2016 | Classic features, Pose Det., PCA, ~78% Avg. Prec. | Street, Part-based annotation & search |
| 12 | Z. Chen [372] | 2017 | Relevance feedback, Feature re-weighting, Bayesian | Shop/Street, Query-free, Interactive |
| 13 | Z. Wang [373] | 2017 | Visual attention, CNN, ImpDrop, ~88.7% Acc@10 | Street/Shop |
| 14 | He [374] | 2017 | BoW, CNN, DML, HOG, LAB, Triplet ranking, 92% P@10 | Runway images |
| 15 | Corbiere [33] | 2017 | ResNet50, Bag-of-words, ~71% Acc@10 | Model, Weakly Annotated Data |
| 16 | Yang [375] | 2017 | ResNet50, Binary hash, Gradient boosting, ~26% P@10 | Wild/Shop, Ebay, Speed, Memory, System |
| 17 | Verma [41] | 2018 | StyleNet, CNN, Attention, ST-LSTM, ~72% mAcc@10 | Street/Shop, Multiple items |
| 18 | X. Wang [376] | 2018 | CNN, Center loss, 99.89% Retrieval Acc | Fabric & Pattern retrieval |
| 19 | Meng [377] | 2018 | Classic, Voting-based, Color, Shape, Back projection | Material image retrieval, Fashion accessory |
| 20 | Kuang [45] | 2018 | Hierarchical, CNN, Divide-and-conquer, 73.80% mAcc | Street, Hierarchical, Path-based |
| 21 | Bhatnagar [48] | 2018 | Compact bilinear CNN, Triplet, 76.26% Acc@20 | Model, Weak annotations |
| 22 | Manandhar [127] | 2018 | Faster-RCNN, RPN, PMAC, VGGNet, 53.60% mAP | Shop, Brand-aware retrieval |
| 23 | Dinh [378] | 2018 | MobileNet SSD, Quantization indexing, 78.21% mAP | Shop, Low latency, Benchmark |
| 24 | Lodkaew [379] | 2018 | Parsing, VGG, DenseNet, Euclidean Dist., 69% P@20 | Street, Instagram, Fashion Finder |
| 25 | Zakizadeh [380] | 2018 | Bilinear CNN, VGG16, Multi-Attr., ~86% Acc@10 | Model, Fine-grained, Mobile-device |
| 26 | Ak [381] | 2018 | AlexNet, Global pooling, Global ranking, ~33% Acc@10 | Item, Weakly supervised localization |
| 27 | Manandhar [382] | 2018 | Attribute-guided triplet, Multi-task CNN, 71.25% mAP | Shop, Tiered similarity search |
| 28 | Kashilani [383] | 2018 | Summary of techniques, Table of 10 previous works | Review |
| 29 | R. Li [109] | 2019 | Multi-task, Multi-weight, Multi-label, CNN, Attr. | Shop, Benchmark 3 methods, Imbalance |
| 30 | Kinli [384] | 2019 | Capsule Net., Stacked Conv., RC block, 75.2% R@10 | Model, Triplet-based |
| 31 | Chopra [385] | 2019 | Inception-v1, Grid Search, Transformation, 95.9% R@10 | Wild/Shop, Robust |
| 32 | Park [62] | 2019 | Multiple methods, CNN, SEResNeXt50, ~92% Acc@10 | Wild/Shop, Benchmark |
| 33 | X. Liu [115] | 2020 | MaskRCNN, ResNet50, Landmark, ~60% Acc@10 | Model, MMFashion Toolbox |
| 34 | Ji [132] | 2020 | Detection, WBF, PCA, KNN, Re-ranking, 85.4% Acc@10 | Wild, 2nd in DeepFashion2 2020 challenge |
| 35 | Jo [386] | 2020 | Implicit profiling, CNN, cGAN, 80.9% P@10 | Shop, Also sketch-based retrieval |
| 36 | Sarmiento [360] | 2020 | VAE, Log-Likelihood, K-Means, 95.5% mAP@10 | Item, Retrieve from synthesized images |
| 37 | Fadhilla [387] | 2020 | DenseNet121, Cosine similarity, 86.23% Acc@10 | Model, Multi-view clothing search |
| 38 | Zhang [79] | 2020 | TS-FashionNet, Two-Stream, 79.04% R@30 | Model, Landmark-aware attention |
| 39 | Ma [388] | 2020 | ASEN, Attribute-aware attention, ACA, ASA, ResNet50 | Model, Attribute-specific retrieval |

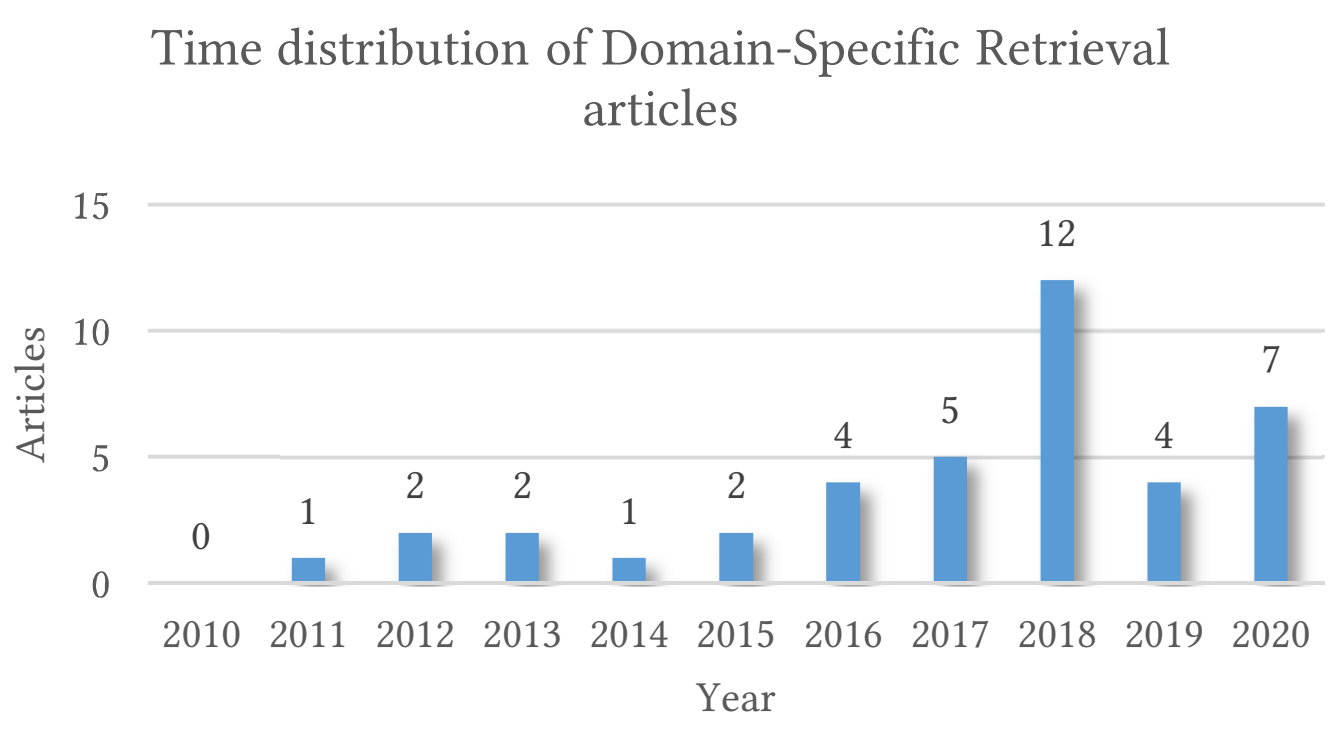

Figure 42: Time Analysis of Domain-Specific Retrieval Articles

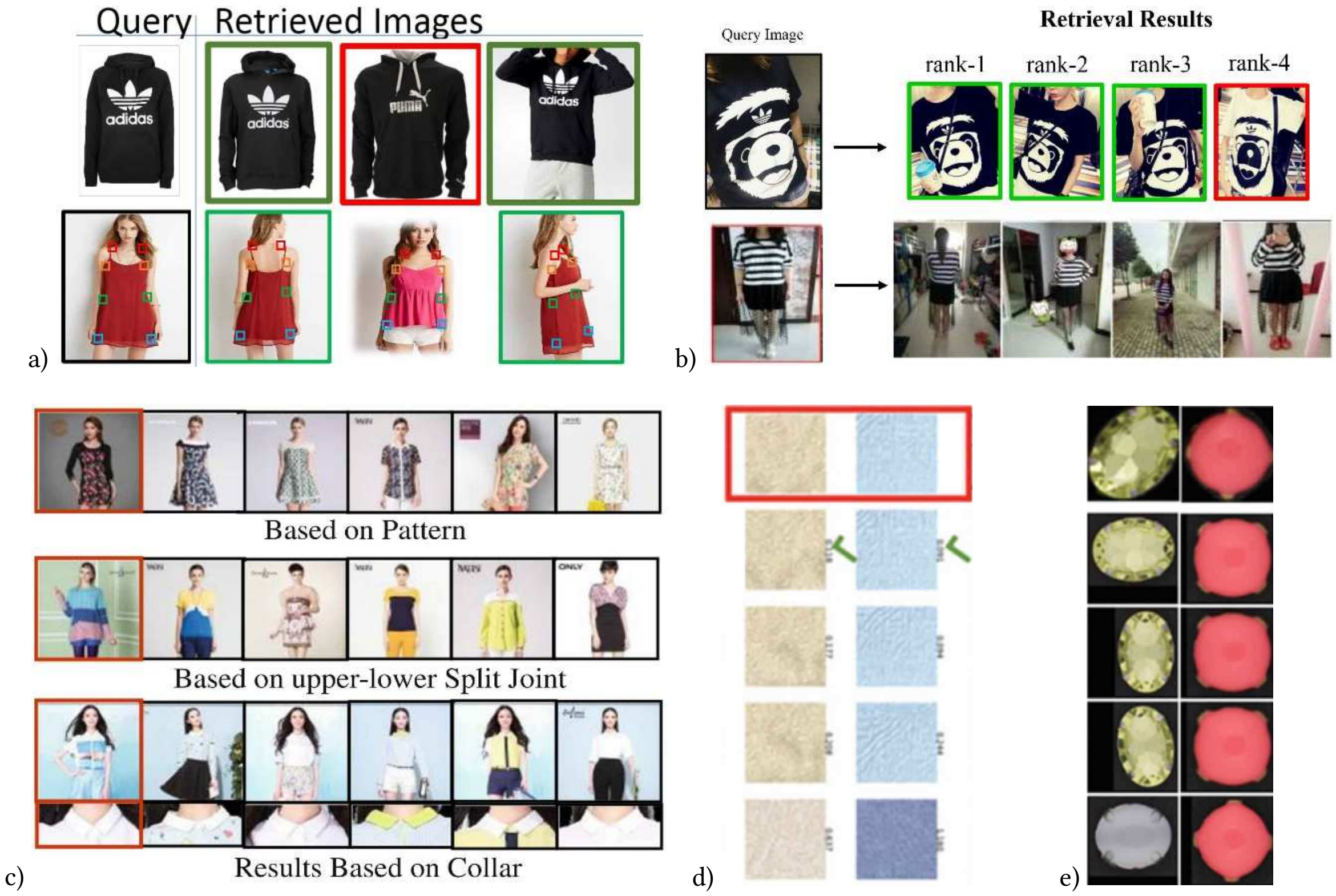

Figure 43: a) In-shop clothing retrieval [123], [127] b) Clothing retrieval in wild images [132], [385] c) Attribute-specific retrieval [92] d) Fabric and pattern retrieval [376] e) Fashion accessory material retrieval [377]

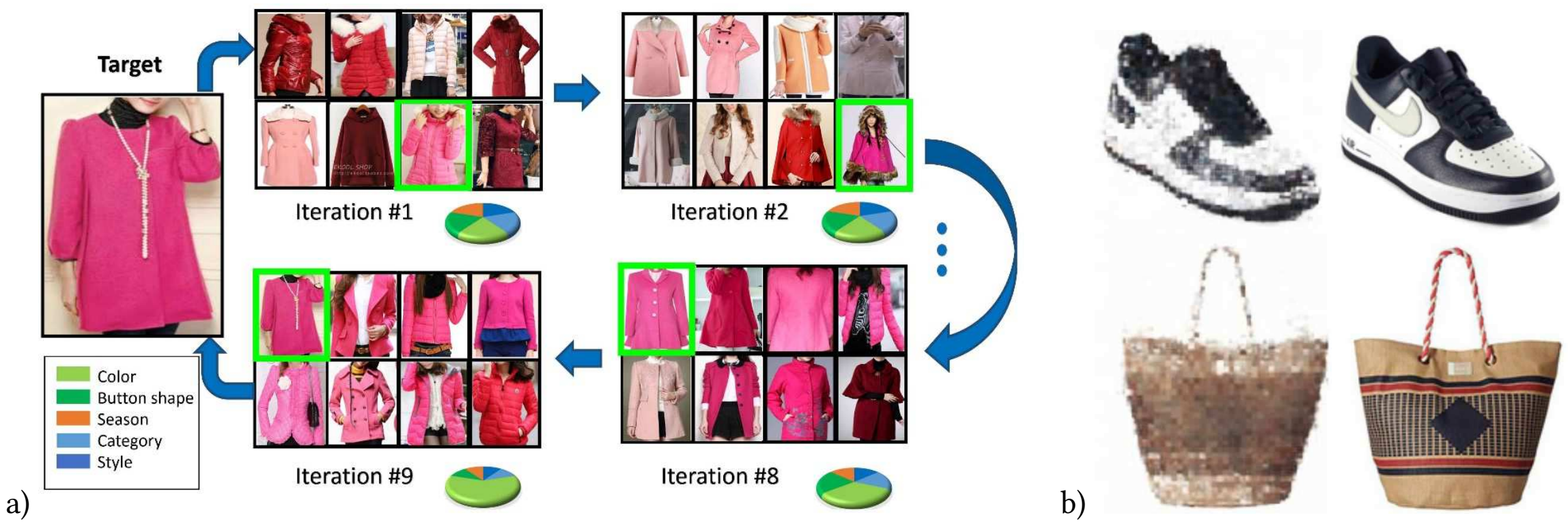

Figure 44: a) Query-free retrieval via implicit relevance feedback [372] b) Retrieval from user-synthesized Item images [360]

### *2.6.2 Cross-Domain Retrieval*

Unlike domain-specific retrieval, these systems bridge the gap between different domains. One example is sketch-to-image retrieval. Another important example is street-to-shop retrieval which uses a user photo to find the exact item in online shops and directly connects street photos to shop items. This task is usually more complex than its domain-specific counterpart and requires particular training data or methods. Multi-modal retrieval systems also do the same, mixing various input types to search in a second domain, e.g., text-to-image retrieval systems and search engines. The first dual keyword (Input domain-Search domain) in the “Application Notes” column of Table 13 shows each work’s input and output domains.

Table 13: Articles Related to Cross-Domain Retrieval

| No | Article Reference | Year | Technical Keywords/Claimed Results | Application Notes |
|---|---|---|---|---|
| 1 | S. Liu [368] | 2012 | Classic, Parts alignment, Auxiliary set, ~67.5% P@10 | Street-Shop, Upper/Lower body |
| 2 | Fu [369] | 2012 | BoW, Detection, Hierarchical Vocab. tree, ~53% P@10 | Street-Shop, Upper/Lower body |
| 3 | Huang [86] | 2015 | DARN, NIN, MLPConv, SVR, PCA, ~49% Acc@10 | Wild-Street/Shop, Attribute-aware |
| 4 | Chen [87] | 2015 | DDAN, R-CNN, SVR, NIN, Alignment cost layer | Street-Shop, Deep domain adaptation |
| 5 | Kiapour [389] | 2015 | CNN, Selective search, Pairs, 30.59% mAcc@20 | Street-Shop, Three methods |
| 6 | Vittayakorn [23] | 2015 | Low-level features, Semantic parse, SVM, 53-55% AUC | Runway-Street |
| 7 | Z. Liu [9] | 2016 | VGG-16, FashionNet, Landmark, ~15% Acc@10 | Wild-Street, Benchmark, DeepFashion |
| 8 | Jiang [390] | 2016 | Bi-directional cross-triplet, AlexNet, 20.34% mAcc@20 | Street-Shop, Shop-Street |
| 9 | Z. Liu [167] | 2016 | DFA, VGG-16, Pseudo-labels, Cascading network | Wild-Street, For landmark detection |
| 10 | X. Wang [391] | 2016 | Robust contrastive loss, Siamese, 37.24% mAcc@20 | Wild-Shop/Street |
| 11 | Yu [392] | 2016 | Triplet-ranking, Siamese, Sketch-a-Net, 87.83% Acc@10 | Sketch-Item, Shoes, Sketch-based retrieval |
| 12 | Z. Li [96] | 2016 | Seg., SVM, Domain-adaptive Dict., K-SVD, ~72% P@10 | Street-Shop, Upper/Lower body |
| 13 | Z. Wang [373] | 2017 | Visual attention, CNN, ImpDrop, ~35% Acc@10 | Wild-Street/Shop |
| 14 | Garcia [393] | 2017 | Feat. tracking, Binary features, KD-tree, FIFO, 87% Acc | Video-Shop |
| 15 | Jaradat [394] | 2017 | CNN, Segmentation, MNC, Localization, VGG-19, LDA | Street (Instagram)-Shop, Methodology only |
| 16 | Shankar [125] | 2017 | VisNet, VGG-16, Faster R-CNN, Triplet, 50.46% mAcc@20 | Wild-Shop, System speed, Memory |
| 17 | He [374] | 2017 | BoW, CNN, DML, HOG, LAB, Triplet ranking, 48% HS | Runway-Street |
| 18 | Cheng [395] | 2017 | VGG-16, LSTM, Spatial pyramid pooling, ~33% Acc@10 | Video-Shop |
| 19 | Verma [41] | 2018 | StyleNet, CNN, Attention, ST-LSTM, ~14% mAcc@10 | Wild-Street/Shop, Multiple items |

| No | Article Reference | Year | Technical Keywords/Claimed Results | Application Notes |
|---|---|---|---|---|
| 20 | Bhatnagar [48] | 2018 | Compact bilinear CNN, Triplet, 17.49% Acc@20 | Wild-Street/Shop |
| 21 | Gajic [396] | 2018 | Siamese, ResNet50, Triplet loss, SGD, ~38% Acc@10 | Wild-Street/Shop |
| 22 | Jiang [397] | 2018 | Robust contrastive loss, Siamese, Inception, ~19% Acc@10 | Wild-Street/Shop |
| 23 | Lasserre [398] | 2018 | fDNA, VGG-16, PCA, 71.2% Acc@10 | Street/Model-Item |
| 24 | Kucer [399] | 2019 | Mask R-CNN, RMAC, Triplet, Ensemble, 60.4% Acc@20 | Street-Shop, Detect then retrieve |
| 25 | Ge [129] | 2019 | Mask R-CNN, Match R-CNN, 52.2% Acc@10 | Wild-Street, Benchmark, DeepFashion2 |
| 26 | Sharma [400] | 2019 | RankNet, Siamese, Fractional Distance, 88.57% R@20 | Wild-Shop, Multi-scale |
| 27 | Tran [61] | 2019 | YOLO, ResNet-18, KNN, Background augmentation | Street-Shop, Detect then retrieve |
| 28 | Lasserre [157] | 2019 | CNN, Seg., U-net, fDNA1.1, 78.5% Acc@10 | Street/Model-Item, Background removal |
| 29 | Park [62] | 2019 | Multiple methods, CNN, SEResNeXt50, ~29% Acc@10 | Wild-Street/Shop, Benchmark |
| 30 | Luo [401] | 2019 | DMCH, LSTM, CNN, Sequential learning, ~7% P@10 | Street-Shop, Efficient |
| 31 | Wu [402] | 2019 | CNN, Multi-modal transformer, 38.62% mR@10 | Natural language-Shop, Fashion IQ |
| 32 | H. Zhang [131] | 2020 | CatGAN, Yolo v2, VGG-19, 53.97% P@10, 2.16% F1@10 | Model-Item, Retrieve from generated items |
| 33 | X. Liu [115] | 2020 | MaskRCNN, ResNet50, Landmark, ~10% Acc@10 | Wild-Street/Shop, MMFashion Toolbox |
| 34 | J. Jo [386] | 2020 | Implicit user profiling, CNN, cGAN, 49.3% P@10 | Sketch-Item |
| 35 | D. Gao [403] | 2020 | BERT, WordPieces, Adaptive loss, 55.74% Rank@10 | Text-Shop, FashionBERT |
| 36 | Y. Jo [404] | 2020 | CNN, LSTM, RNN, VGG-16, Pseudo-SQL, 86.60% F1 | (Gender+Cat.+Color)-Shop, Multi-modal |
| 37 | Su [118] | 2020 | Attentional bilinear Net., Landmark, 53.5%Acc@10 | Wild-Street/Shop |
| 38 | Miao [405] | 2020 | Feature fusion, Quadruplet loss, ResNet-50, ~29% P@10 | Wild-Street/Shop |
| 39 | Y. Zhang [79] | 2020 | TS-FashionNet, Two-Stream, 70.40% R@20 | Wild-Street/Shop, Landmark-aware |
| 40 | Ma [388] | 2020 | ASEN, Attribute-aware, ACA, ASA, ResNet, 61.02% mAP | Wild-Street/Shop, Attribute-specific Retr. |
| 41 | Y. Gao [406] | 2020 | Graph reasoning, Similarity pyramid, ~57% Acc@10 | Wild-Shop |
| 42 | Zhan [366] | 2020 | PNAPGAN, U-Net, Triplet loss, 78.89% mAP@10 | Wild-Item, Pose normalization |
| 43 | Y. Zhang [407] | 2021 | Detector, DLA-34, NLP, Tracklet, Audio to text, 16.8% mR | Video-Shop, Live, Multi-modal, Demo |

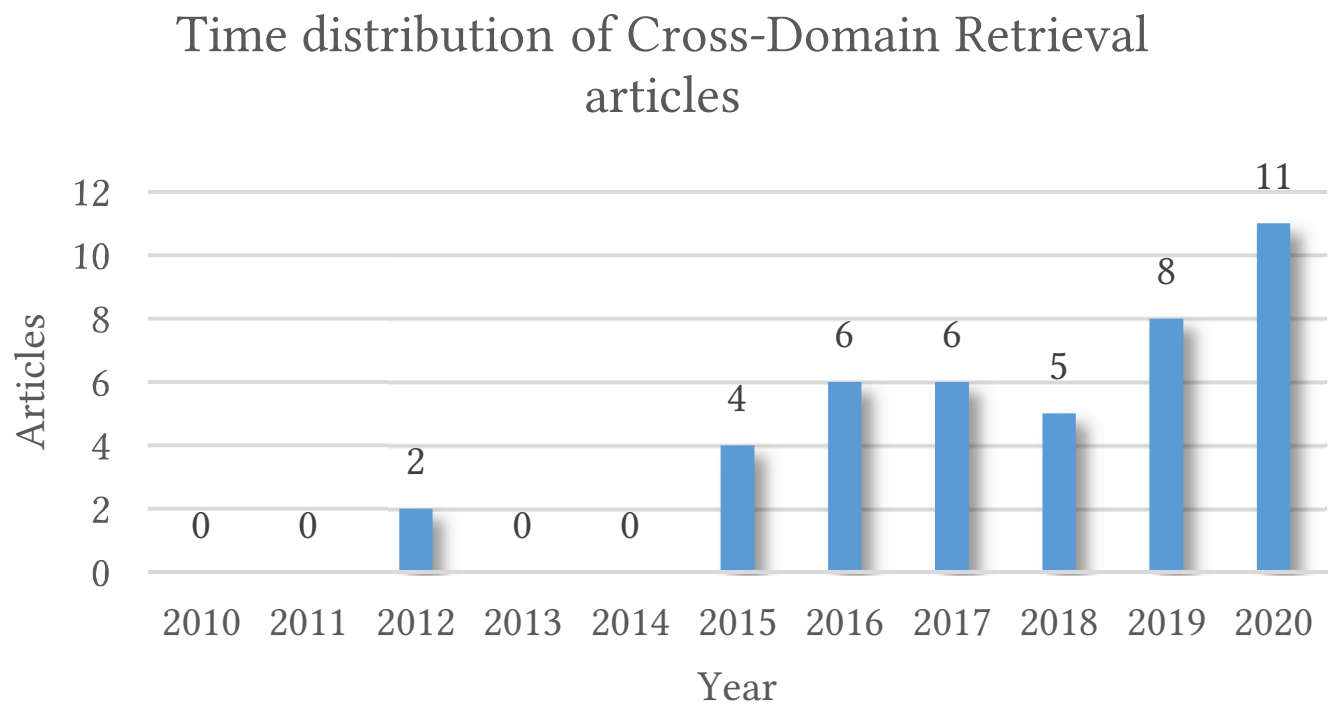

Figure 45: Time Analysis of Cross-Domain Retrieval Articles

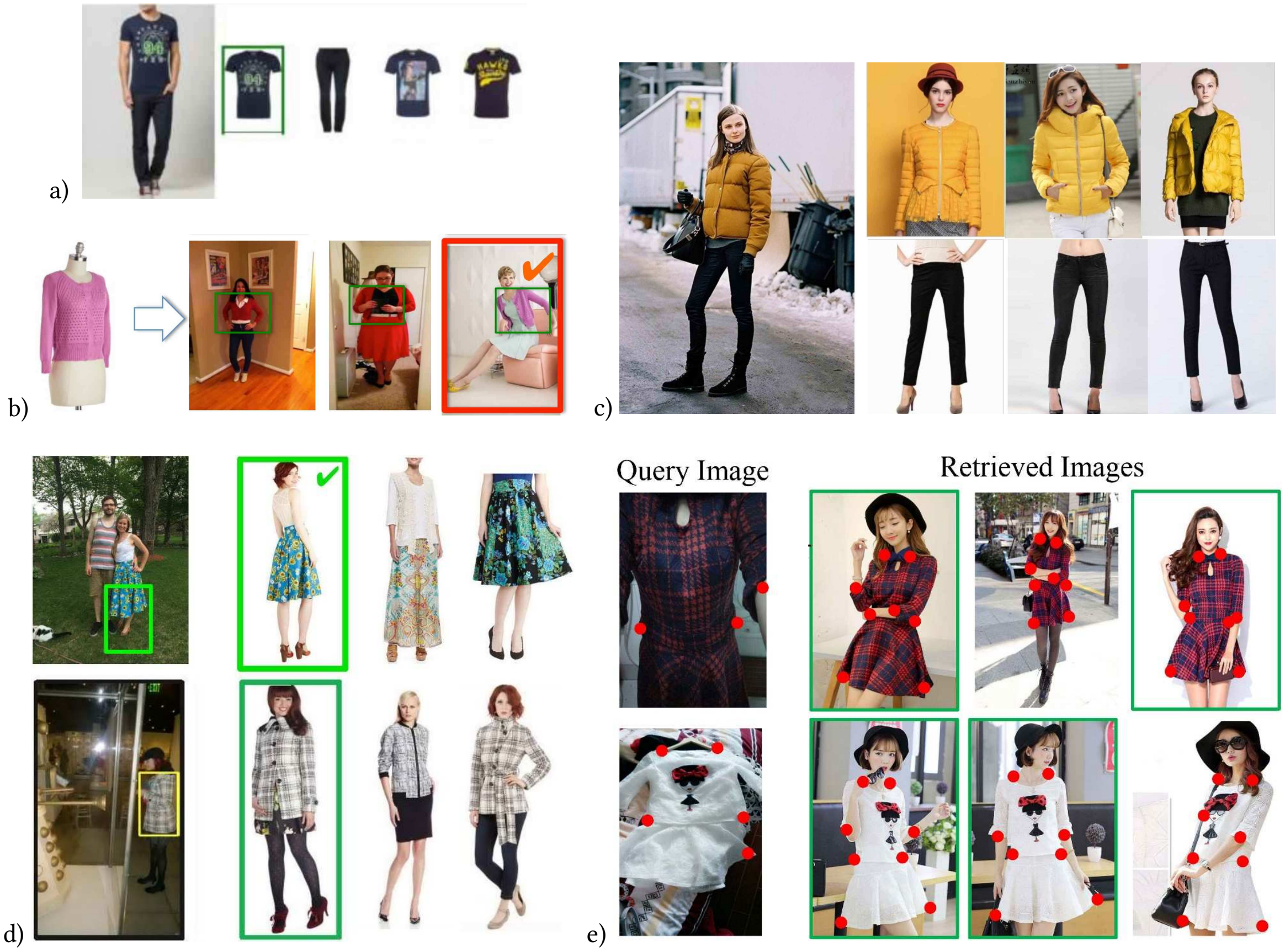

Figure 46: Different cross-domain retrieval schemes. a) Model-Item [398] b) Item-Wild [390] c) Street-Shop [96] d) Wild-Shop [389], [399] e) Wild-Street [167]

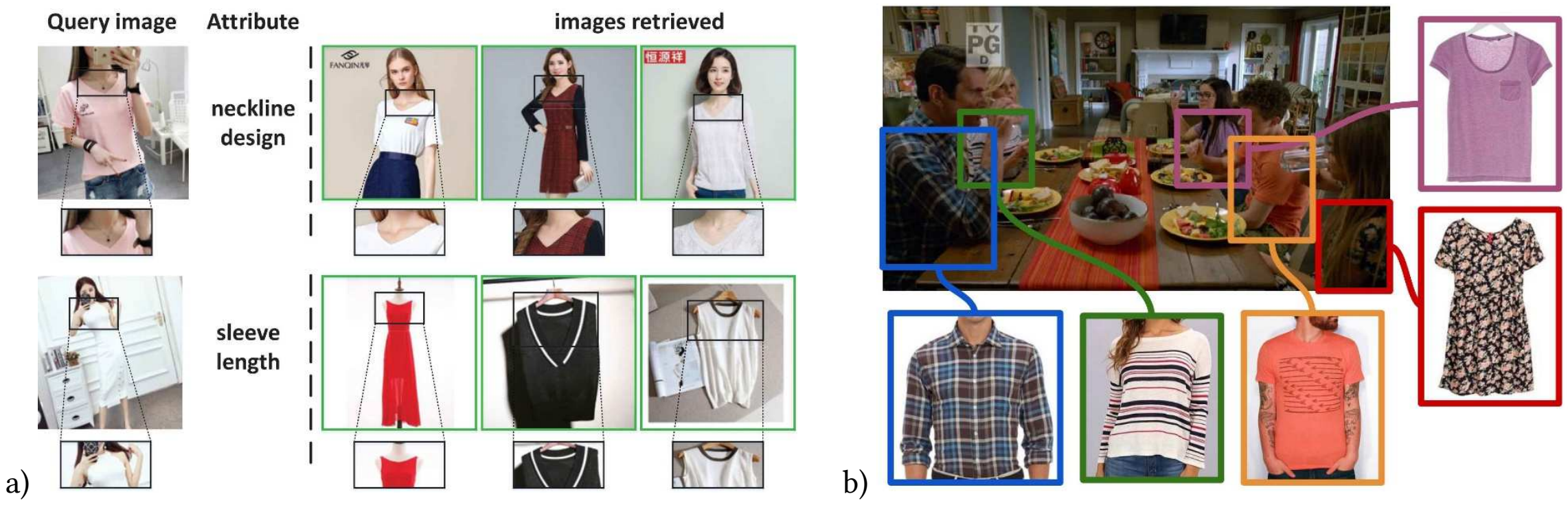

Figure 47: a) Attribute-specific Wild-Shop retrieval [388] b) Fashion retrieval from videos [393]

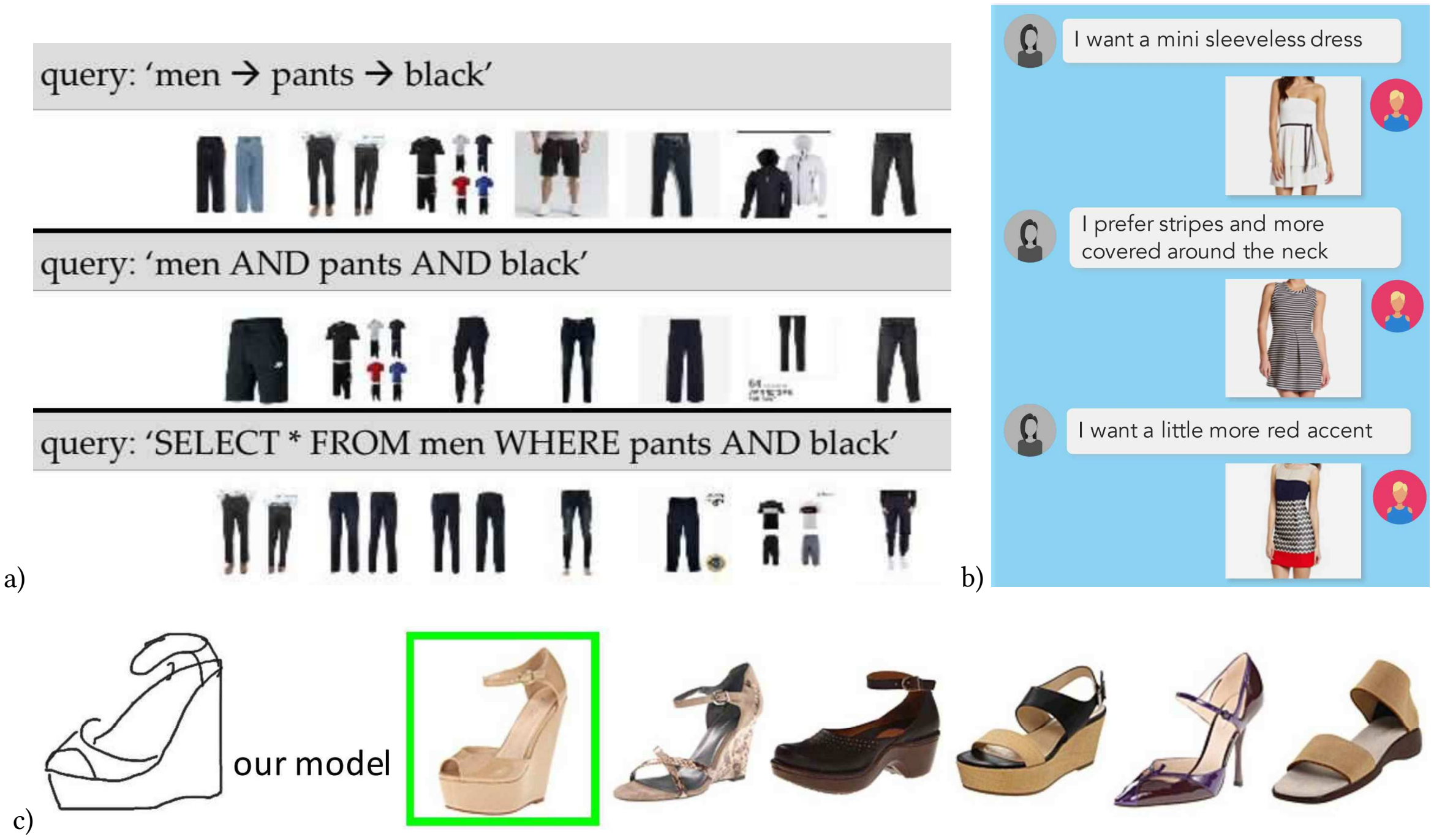

Figure 48: a) Multimodal (category/Boolean/SQL) search and retrieval [404] b) Dialog-based fashion search [402] c) Sketch-Item shoe retrival [392]

### *2.6.3 Retrieval with Attribute Manipulation*

Sometimes we need a match for our item, but with a bit of change, that is when attribute manipulation comes in. These systems change some attributes of the query item based on the user's specification, then retrieve item matches. For example, they can retrieve a long-sleeved version of a short-sleeve shirt or a red version of a blue dress. Meaning, all attributes of the item stay intact until we specifically change any of them. Another example is interactive search using relative attributes, asking the system for a "more comfortable" shoe or a "less formal" dress. We use dual keywords (Input item-Target attribute) to show the input types of each system in Table 14 if possible.

Table 14: Articles Related to Retrieval with Attribute Manipulation

| No | Article Reference | Year | Technical Keywords/Claimed Results | Application Notes |
|---|---|---|---|---|
| 1 | Liu [368] | 2012 | Classic, Part-based, Multi-task sparse representation | Mix features from different parts |
| 2 | Kovashka [408] | 2012 | Binary relevance feedback, Re-ranking, SVMRank | Relative Attr. feedback, Interactive |
| 3 | Yu [409] | 2014 | Classic, Local learning, Relative attributes, ITML | Visual comparison, Fine-grained, Shoes |
| 4 | Koike [410] | 2015 | Icons, User interface, Graph, Force-directed, Space-filling | Category, Color, Pattern, Interactive |
| 5 | Kovashka [411] | 2015 | Classic, WhittleSearch, Binary search tree, SVMRank | Relative Attr. feedback, Interactive |
| 6 | Zhou [412] | 2016 | Classic, Hybrid topic, HOG, LBP, BOW, 0.66 NDCG@20 | Graphical user interface |
| 7 | Lee [15] | 2017 | CNN, Representation learning, Style2Vec, VGG, Style set | Item-Item, Style manipulation |
| 8 | Han [413] | 2017 | GoogleNet, BOW, Word2vec, EAAM, ~20% mAcc@10 | Model-Text, Concept discovery |
| 9 | Zhao [414] | 2017 | AMNet, CNN, Memory-augmented, 0.39 NDCG@20 | Street/Shop-Text |

| No | Article Reference | Year | Technical Keywords/Claimed Results | Application Notes |
|---|---|---|---|---|
| 10 | Liao [101] | 2018 | EI tree, BLSTM, ResNet50, Ranking loss, ~65% R@10 | Street/Shop-Text, Interpretable retrieval |
| 11 | Laenen [415] | 2018 | Text Seg., Word embedding, BVLC CaffeNet CNN | Shop-Text, Multi-modal search |
| 12 | Ak [381], [416] | 2018 | AlexNet, ROI, Global ranking, Triplet, ~25% Acc@10 | Item images, Replace Attr., Localization |
| 13 | Tan [417] | 2019 | SCE-Net, CNN, Condition weight branch, Triplet | Similarity condition masks, Multi-modal |
| 14 | Tautkute [418] | 2019 | DeepStyle, ResNet-50, CBOW, KNN, Fusion | Item-Text, Multi-modal search |
| 15 | Wu [402] | 2019 | CNN, Multi-modal transformer, 38.62% mR@10 | Relative Attr. feedback, Interactive |

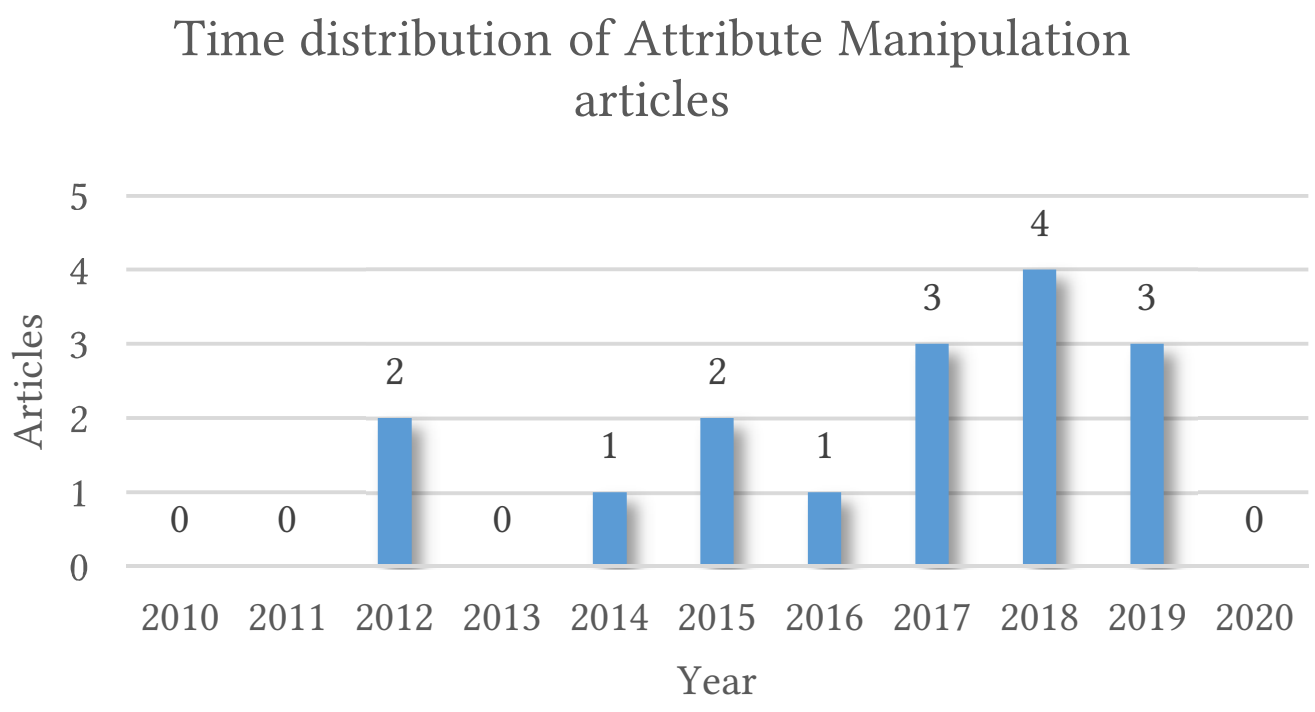

Figure 49: Time Analysis of Attribute Manipulation Articles

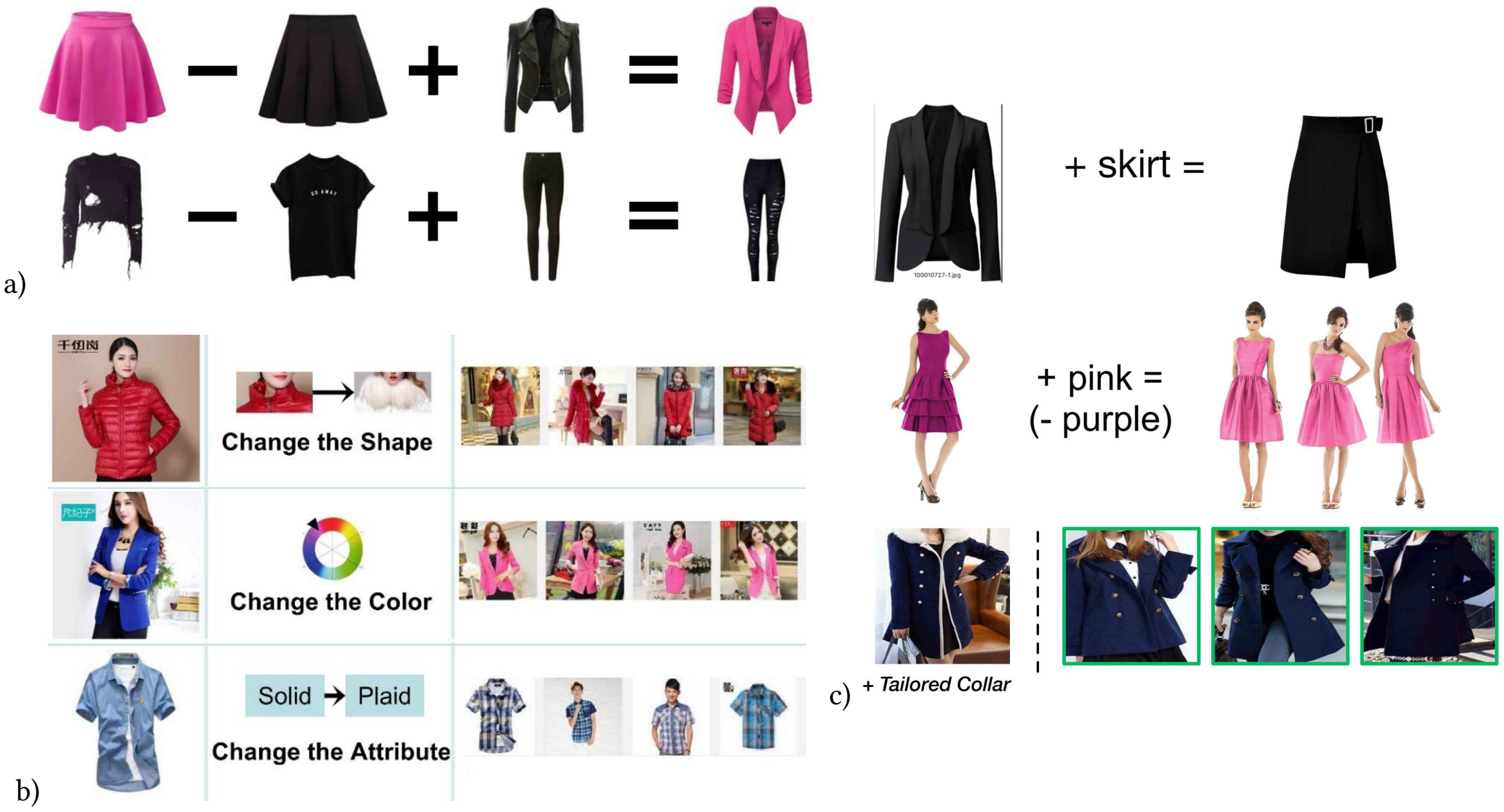

Figure 50: a) Manipulation of style factors [15] b) Interactive fashion search with manipulation [412] c) Attribute manipulation on Item [418], Model [413], and Street [414] images

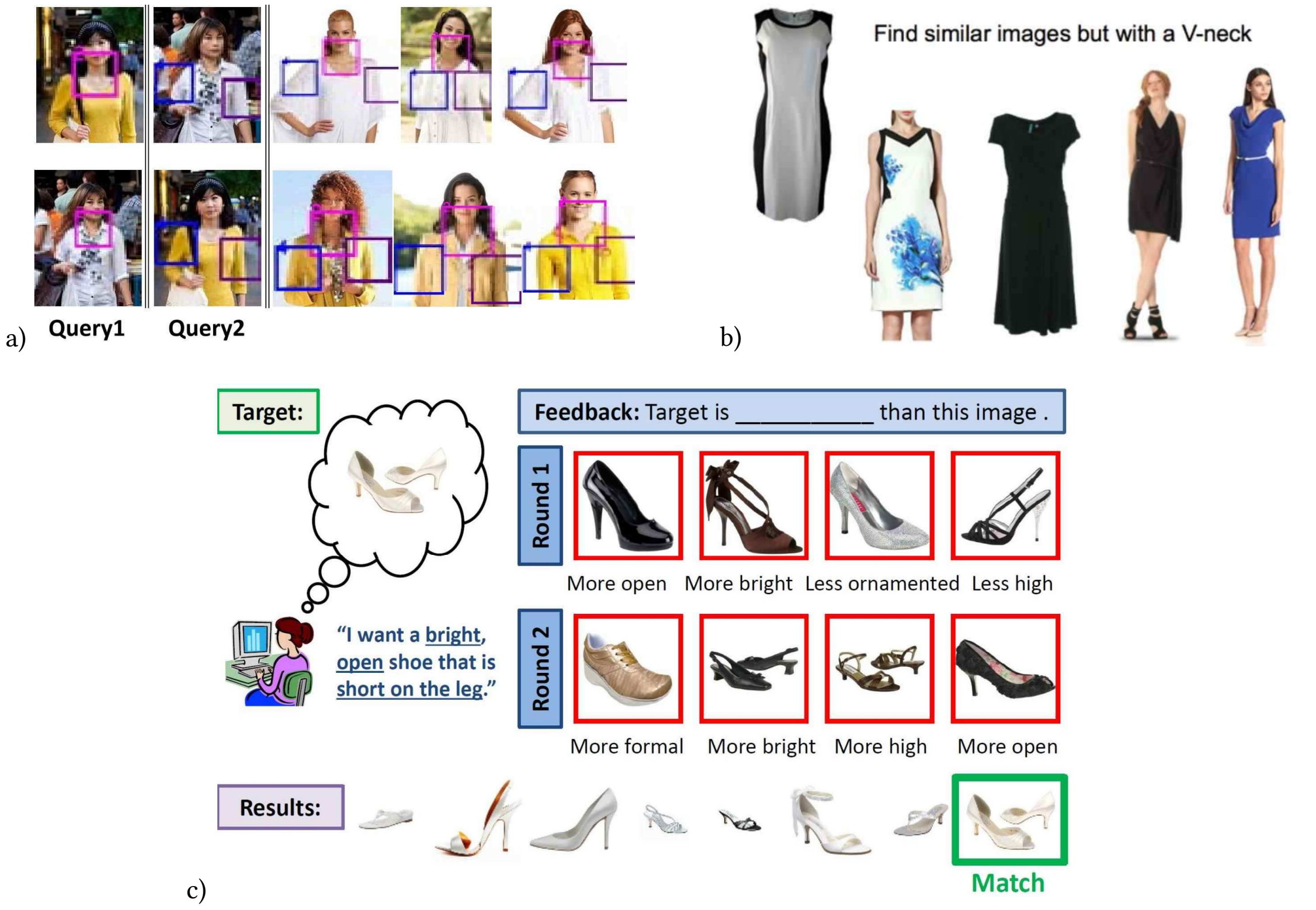

Figure 51: a) Attribute mixing [368] b) Multimodal text-guided retrieval with attribute manipulation [415] c) Interactive search with relative attribute feedback [411]

### 2.7 Recommender Systems

Recommender systems suggest fashion items based on similarity, style, color, user preference, and many more different schemes. Recommender systems study is a highly active research field, and recommenders are already used in many online shops, including Amazon, Google Shopping, and Shop It To Me. These systems are not only beneficial to online fashion retail shops, but they are also finding their way to physical stores [419]. We group these systems into four sub-categories: 1) Single-Item Recommender, 2) Style or Outfit Recommender, 3) Personalized Recommender, and 4) Fashion Compatibility. We can also use attribute-specific retrieval systems in Sec. 2.6 as attribute-guided recommender systems.

Various types of recommender systems exist, including collaborative filtering (CF), content-based (CB), knowledge-based (KB), and hybrid systems. Different systems use purchase history, images, reviews, user ratings, clicks, temporal information, and various other input data to generate recommendation lists. A list of survey studies on deep learning-based recommender systems is provided in Table 15.

It is essential to mention that pure content-based recommender systems have significant overlap with retrieval systems. Researchers should be aware that although these are two different applications with distinct purposes, the fundamentals of these two systems can be very similar. We do not want exact matches in recommender systems, however, obtaining the exact match is the goal of retrieval systems. The methods to tackle these two problems sometimes are the same, and it is only a matter of perspective on how to use them. These two applications both take advantage of the similarity between items; thus, a recommender system might find two items 100% similar and actually retrieve that item. On the other hand, we can use a retrieval system to list items, emit the exact match and use the rest as recommendations. So it is a good idea also to consult Sec. 2.6 to know more about different retrieval methods.

Table 15: Survey Articles on Recommender Systems

| No | Article Reference | Year | Information |
|---|---|---|---|
| 1 | Guan [269] | 2016 | Empirical review, Market (A list of online apparel recommendation platforms), Methods, Types |
| 2 | Liu [420] | 2018 | Not fashion-specific, Different datasets, Models, Application |
| 3 | Zhang [421] | 2019 | Not fashion-specific, Comprehensive, Techniques, Analysis, Applications, Future research |
| 4 | Sachdeva [422] | 2020 | Interactive systems, A table of 11 studies, each with an objective and the proposed solution |
| 5 | Chakraborty [423] | 2020 | Image-based style prediction & Recom., A table of 64 studies with key features of each study |
| 6 | Laenen [424] | 2020 | Comparative study of attention-based fusion methods, Fashion compatibility, Four benchmarks |
| 7 | Gong [425] | 2021 | Comprehensive, Different Recom. systems (Full explanation), Aesthetics, Personalization |

### *2.7.1 Single-Item Recommenders*

These systems only recommend one fashion item may it be a shirt, dress, shoe, etc. They can be content-based recommender systems that retrieve similar articles based on visual features of images or semantic attributes. The system input can be a query image or text, and the output is a sorted list of recommended fashion items. One thing to keep in mind is that these recommenders only work within clothing categories. If the input is an image of a shirt, the output list will be the same and not from another category. Table 16 lists these recommender systems. Some systems only target one fashion article in each full-body photo; on the other hand, other systems detect multiple pieces in each image and provide a separate list of recommendations for each item. We assign the "Multiple" keyword to these systems in the last column. We also report each work domain using a single keyword (Item, Model, Shop, etc.) or dual keywords (Input domain-Search domain) for cross-domain systems wherever needed.

Table 16: Articles Related to Single-Item Recommenders

| No | Article Reference | Year | Technical Keywords/Claimed Results | Application Notes |
|---|---|---|---|---|
| 1 | Chao [10] | 2009 | Classic, ROI, HOG, LBP, Web camera | Smart mirror, Style, Robust |
| 2 | Goh [426] | 2011 | Framework only, Color-based, RFID tag, Prototype | Smart wardrobe, Occasion, Mood |
| 3 | Sekozawa [427] | 2011 | Classic, AHP, Cluster analysis, Market basket analysis | Online system |
| 4 | Huang [428] | 2013 | Active learning, Support vector regression, Sparse coding | Shop |
| 5 | Kalantidis [19] | 2013 | Seg., Multi-Probe LSH, LBP, Randomized kd-tree, 54% HS | Street-Item, Multiple, Detection |
| 6 | Hu [429] | 2014 | Classic, Hybrid, Collaborative, Cluster, HOG, HSV | Features+Ratings |
| 7 | Bhardwaj [430] | 2014 | Crowd-sourcing, Deterministic/Stochastic Recom. | User understanding |
| 8 | He [431] | 2015 | VBPR, Matrix factorization, Deep CNN, 0.7364 mAUC | Shop, Visual BPR, Personalized |
| 9 | Lao [22] | 2015 | Classification, Attributes, R-CNN, AlexNet, KNN | Wild images |
| 10 | McAuley [8] | 2015 | CNN, Shifted sigmoid, Mahalanobis, KNN, 91.13% mAcc | Amazon, Substitutes, Complements |
| 11 | Viriato de M. [432] | 2015 | Eye fixation & saccade, Semantic parts, 0.58% AP@9 | Eye tracking, Human visual attention |
| 12 | He [433] | 2016 | One-Class CF, CNN, Temporal dynamics, TVBPR+ | Item, Temporal-aware |
| 13 | K. Liu [91] | 2016 | VGG-16, Decision fusion, Euclidean Dist., ~69% P@10 | Model, View-Invariant, MVC |
| 14 | He [434] | 2016 | Fashionista, One-Class CF, MF, Deep CNN, t-SNE | Item, Graphical interface |
| 15 | Sha [92] | 2016 | Classic, Seg., Color matrix, ULBP, PHOG, Fourier, GIST | Model/Street, Attribute-guided |
| 16 | Vaccaro [93] | 2016 | Polylingual topic model, Gibbs sampling, MALLET | (Street+Natural language)-Item |
| 17 | Bracher [13] | 2016 | fDNA, k-means, Logistic factorization, DNN, CNN | Item, Content+Sale data |

| No | Article Reference | Year | Technical Keywords/Claimed Results | Application Notes |
|---|---|---|---|---|
| 18 | Zhou [435] | 2017 | Statistical NLP, Fuzzy Math., Clustering, 97.96% HS | Item, Also mix & match |
| 19 | Sharma [436] | 2017 | Classic, MATLAB, Gabor filter, Circle Hough transform | Shop, Lines and patterns, GUI, Server |
| 20 | Qian [31] | 2017 | Seg., ASPP, CRF, Faster R-CNN, k-means, VGG-16 | Street-Shop, Multiple, Color, Pattern |
| 21 | Shankar [125] | 2017 | VisNet, VGG-16, Faster R-CNN, Triplet, 97% HS | Wild-Shop, Retrieval, Speed, Memory |
| 22 | Chen [32] | 2017 | CNN, Distributed computing, ~55.5% P@10 | Wild/Street, Four Architectures, Datasets |
| 23 | Kang [339] | 2017 | Siamese CNN, MF, GAN, BPR, 0.7547 mAUC | Shop-Shop/Synthesized images |
| 24 | Y. Liu [437] | 2017 | Advanced user-based CF, Cosine similarity, 22.87% Prec | Shop |
| 25 | Heinz [438] | 2017 | Fashion DNA, LSTM, Dynamical, 0.885 AUC | Shop, Purchase sequences |
| 26 | Zhang [36] | 2017 | mCNN-SVM, AlexNet, Cutting-plane, 43.30% mAP | Street, Multiple, Scenario-Oriented, Color |
| 27 | Veit [39] | 2017 | Conditional similarity Net., CNN, Triplet, Similarity mask | Item, Learning specific notions of similarity |
| 28 | Kottage [439] | 2018 | NLP, TF-IDF, Text mining, NER, Classic classifiers | User reviews, Sentiment, Hybrid |
| 29 | Hwangbo [440] | 2018 | Item-based CF, K-RecSys, ORACLE PL/SQL | Clicks, Sales, Preference |
| 30 | Wen [441] | 2018 | Knowledge graph, Collaborative filtering, Top-N | User's context: Weather, Occasion, ... |
| 31 | Packer [442] | 2018 | MF, BPR, I-VBPR, Temporal dynamics, 0.7497 AUC | Shop, Interpretable |
| 32 | Li [419] | 2018 | CF, CB, FAST, PCA, K-Means, 26.70% HS | Shop, Offline shopping |
| 33 | Verma [41] | 2018 | StyleNet, CNN, Attentional LSTM, Spatial transformer | Street, Multiple, Part-based |
| 34 | Yang [443] | 2018 | Classic, Knowledge base, Matching rules | Expert knowledge |
| 35 | Yu [444] | 2018 | CNN, Brain-inspired deep Net., DCFA, BPR, ~5% R@10 | Shop, Aesthetic-based |
| 36 | Andreeva [445] | 2018 | ResNet101, VisNet, Multi-label, Shallow, ~16% R@10 | Shop |
| 37 | Vasileva [446] | 2018 | CNN, Type-aware embedding, Euclidean Dist., Triplet | Item, Compatibility |
| 38 | Ramesh [128] | 2018 | Object detection, Faster RCNN, NN parse, 0.82 NDCG | Street, Scenario-oriented, Events |
| 39 | Deng [106] | 2018 | CNN, CF, LBP, Fine-grained Attr., 71.44% F@15 | Mobile application |
| 40 | K. Gorripati [49] | 2018 | CNN, VGG16, Cosine similarity, Classification-based | Shop |
| 41 | Cardoso [107] | 2018 | VGG-16, Multi-modal Fusion, RNN, CF/CB hybrid Recom. | Shop, ASOS fashion e-commerce |
| 42 | Hidayati [320] | 2018 | Graph, Auxiliary visual words, BoVW, AP clustering | Celebrities, Body shape, Style |
| 43 | Ok [447] | 2019 | User-based CF, Graph-based random walk, ~7% R@10 | Fashion trends, Seasons |
| 44 | Hou [448] | 2019 | SAERS, CNN, Siamese, Grad-AAM, ROI, ResNet-50, BPR | Shop+User history, Explainable |
| 45 | Tan [417] | 2019 | SCE-Net, CNN, Condition weight branch, Triplet | Similarity condition subspace learning |
| 46 | Asiroglu [12] | 2019 | CNN, Inception, Haar-cascade, DoG, 75% HS | Embedded Linux system |
| 47 | Yan [449] | 2019 | FiDC, Stacked autoencoder, 84.3% clustering Acc | Item, Unsupervised deep clustering |
| 48 | Tuinhof [60] | 2019 | CNN, AlexNet, Batch normalized Inception, KNN, ADAM | Shop, Classification |
| 49 | Vishvakarma [450] | 2019 | CNN, MILDNet, Multi-scale, Skip, 93.69% Triplet Acc | Shop, Compact |
| 50 | Lasserre [157] | 2019 | CNN, Seg., U-net, fDNA1.1, 71.8% Retrieval Acc@10 | Street/Model-Item, Background removal |
| 51 | Cheng [451] | 2019 | Multi-modal aspect-aware topic model, 40.4% P@10 | Shop, Reviews, Explainable, Not just fashion |
| 52 | Ramampiaro [452] | 2019 | KNN, BPR, ALS-WR, Implicit feedback, 3.6% mAP | Shop, Methods survey, SoBazaar |
| 53 | Sherman [453] | 2019 | Multifaceted offline Evaluation, MP, CF, CB | Benchmark, Assessing three methods |
| 54 | Ravi [133] | 2020 | Mask RCNN, CNN, Triplet, Active learning, 15.8% P@10 | Street-Shop, Multiple, Detect |
| 55 | Kavitha [454] | 2020 | CNN, VGG-16, BoW, Word2Vec, TDF-IDF | Shop, Also text-based |
| 56 | Kotouza [455] | 2020 | UI, NLP, Deep reinforcement learning, Clustering | Multiple clustering methods, User feedback |
| 57 | Hsiao [326] | 2020 | ViBE, HMD, SMPL, CNN, Binary classifier, 0.58 AUC | Body shape aware, Explainable |
| 58 | Fengzi [78] | 2020 | ResNet-50, Autoencoder, Cosine similarity | Shop, For classification |
| 59 | Mohammadi [122] | 2021 | EfficientNet, Cosine similarity, Background Aug., OHS | Shop, Street-Shop, Introduces OHS metric |

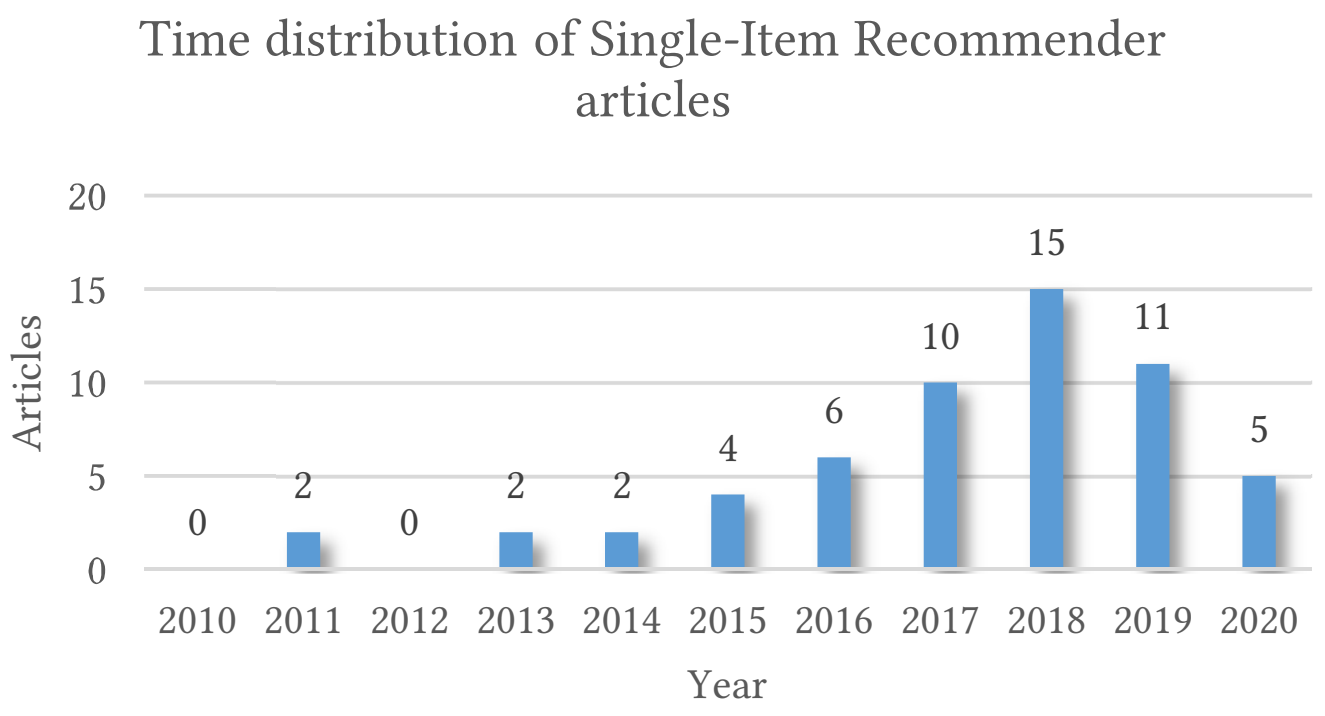

Figure 52: Time Analysis of Single-Item Recommender Articles

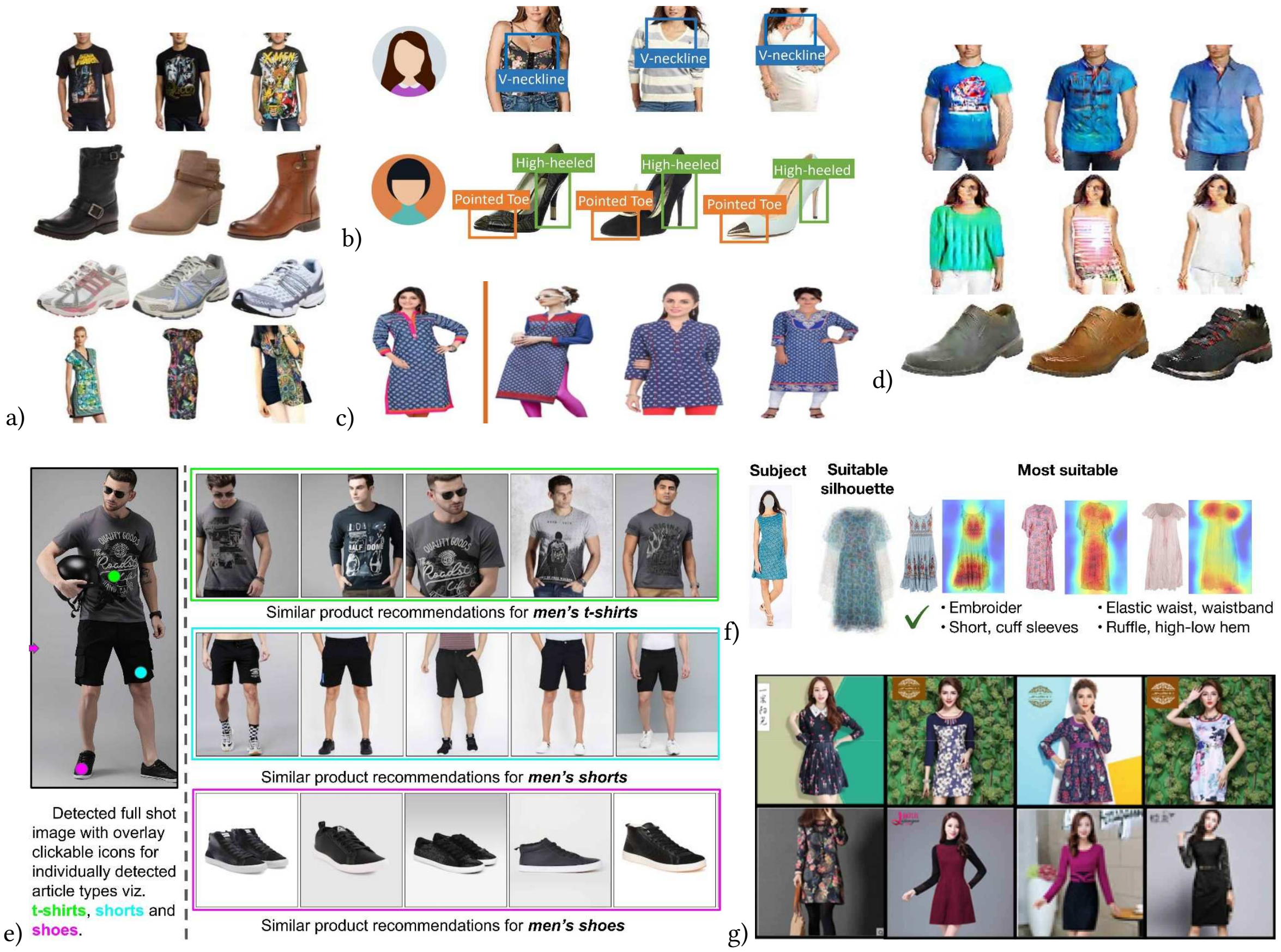

Figure 53: Different kinds of single-item recommendation systems. a) In-shop Recom. [8] b) Explainable semantic region guided Recom. [448] c) Model images [125] d) Personalized Recom. image synthesis [339] e) Selective Item Recom. from full shot images [133] f) Explainable body shape-aware Recom. [326] g) Similar Street images [92]

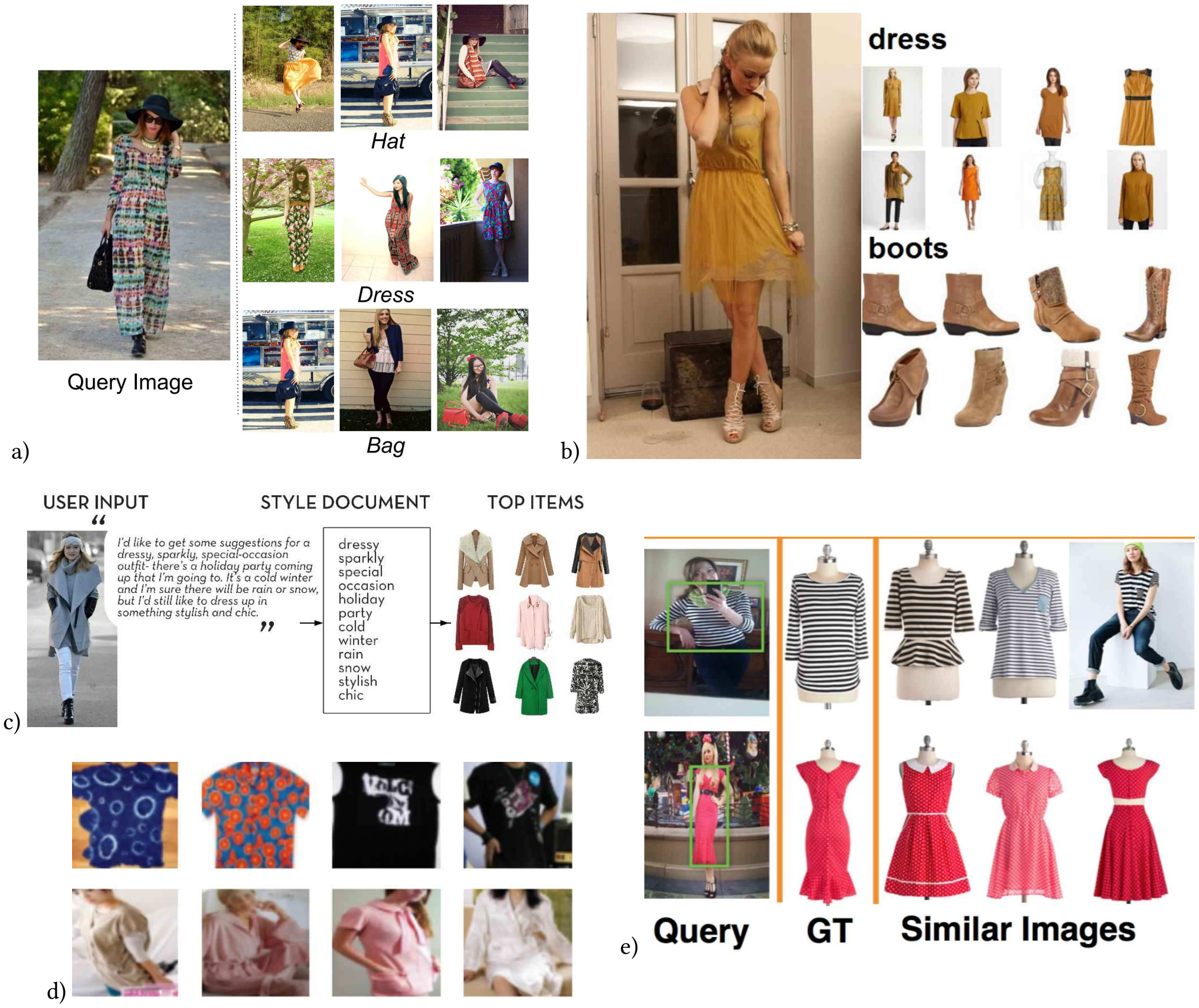

Figure 54: a) Street images Recom. with selective article [41] b) Street-Shop multiple items Recom. [19] c) Natural Language guided Street-to-Item Recom. [93] d) Wild images Recom. [22] e) Wild-Shop Recom. [22]

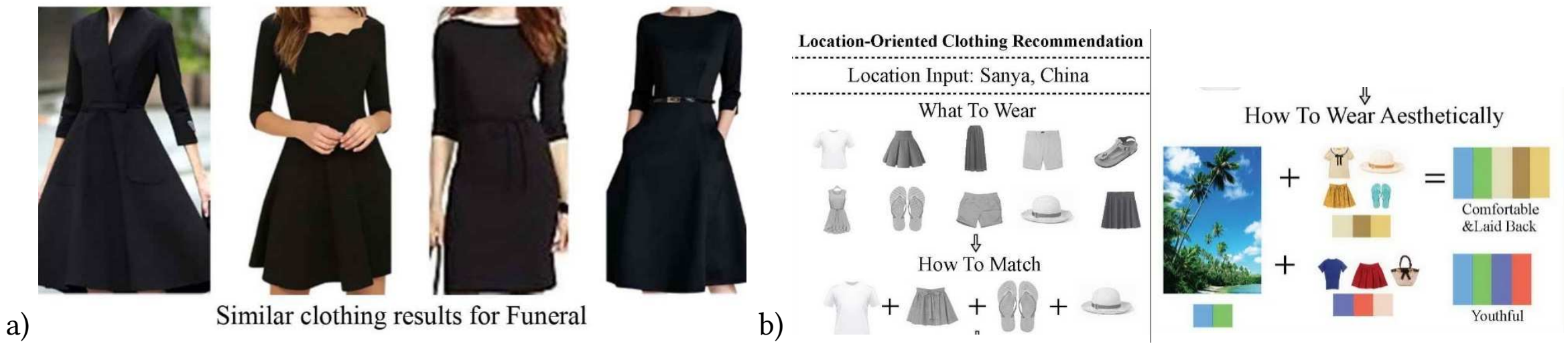

Figure 55: a) Scenario-oriented Recom. [128] b) Location-oriented Recom. [36]

### *2.7.2 Outfit Style Recommenders*

Style recommenders do not focus on one item but on all clothing items in an image as a style and fashion instead. This task should not be confused with a single-item style recommender because, unlike single-item versions, these systems output a single image of an entire outfit. Other versions might output multiple images of items that make a whole outfit; these systems are discussed in Sec. 2.7.3.

Table 17: Articles Related to Outfit Style Recommenders

| No | Article Reference | Year | Technical Keywords/Claimed Results | Application Notes |
|---|---|---|---|---|
| 1 | Yu-Chu [456] | 2012 | Classic, Modified Bayesian Net., User feedback | Color, Season, Occasion, Usage history |
| 2 | Liu [83] | 2012 | Latent SVM, Non-convex cutting plane, 0.75 NDCG@10 | Magic closet, Scenario/Occasion-oriented |
| 3 | Simo-Serra [457] | 2015 | Conditional random field, BoW, DNN, 17.36% IOU | Fashionability, Neuroaesthetics |
| 4 | Hsiao [99] | 2017 | Polylingual LDA, Topic model, 28.48% mAP | Style-coherent, Mix styles, Street |
| 5 | Ding [458] | 2018 | Bilinear supervised hashing, SURF | Fashion shows images |
| 6 | Verma [68] | 2020 | Faster RCNN, MobileNet MTL, Feature-weighted clusters | Occasion-oriented style Recom. |
| 7 | Kavitha [454] | 2020 | CNN, VGG-16, BoW, Word2Vec, TDF-IDF | Text to Model outfit image |
| 8 | Zheng [459] | 2020 | Multi-modal, VGG-19, Seg., Triplet, MLP, 24.20% R@10 | Street images, Hashtags, Social media |

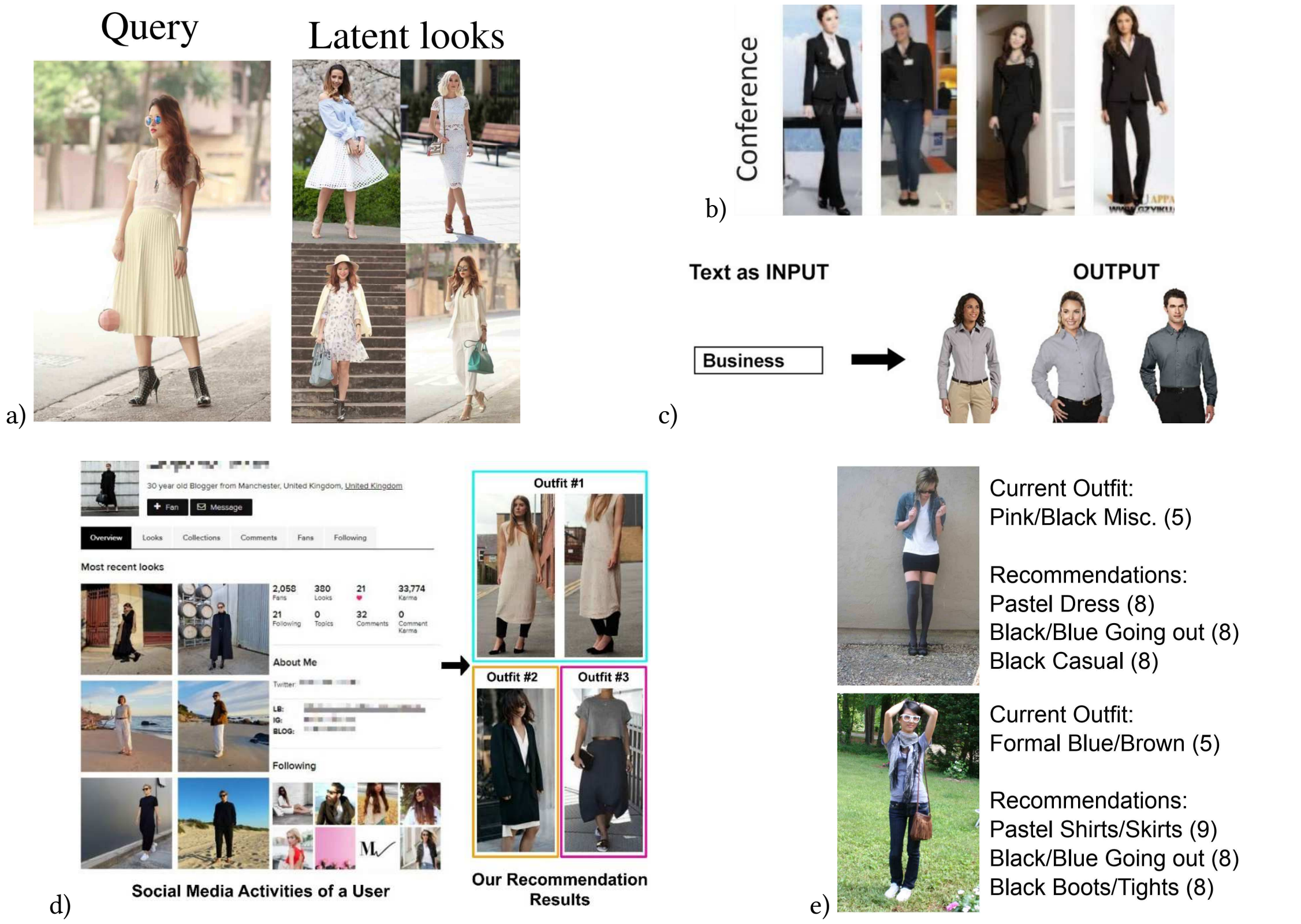

Figure 56: Outfit style recommenders. a) Style-coherent Street images [99] b) Scenario-oriented [83] c) Text-guided scenario-oriented [454] d) Personalized Recom. from social media activities [459] e) Improve outfit fashionability (scores in parenthesis) [457]

### 2.7.3 Fashion Compatibility

These systems predict whether different fashion items go together or not. This application is also known as Fashion Collocation, Outfit Matching, Mix & Match, and Fill In The Blank problem. It can also be seen as a cross-category item recommender, which recommends a list of shoes compatible with a query image of jeans, for example. These recommender systems bridge different fashion item categories; instead of recommending substitutes, they suggest complementary items. Various systems exist in this category; some only need one input item to recommend multiple missing articles and form a compatible set of clothing; on the other hand, others take several items as input and recommend one missing item to make an outfit whole known as "Fill in the Blank (FITB)" task. The number of recommended output articles also varies in different structures; we report this, using keywords in the "Application Notes" column of Table 18 if needed. For example, "Top/Bottom" shows a dual-item system, "Outfit" shows three or more predefined outputs, and "Multiple" refers to other systems with various input/output lengths.

Table 18: Articles Related to Fashion Compatibility

| No | Article Reference | Year | Technical Keywords/Claimed Results | Application Notes |
|---|---|---|---|---|
| 1 | Iwata [124] | 2011 | Probabilistic topic model, SIFT, LDA, ~55% mAcc | Top/Bottom, Full-body shots, Detection |
| 2 | S. Liu [83] | 2012 | SVM, Non-convex cutting plane, ~0.70 mNDCG@10 | Top/Bottom, Scenario/Occasion-oriented |
| 3 | Jagadeesh [460] | 2014 | Gaussian mixture, KNN Consensus, Markov chain LDA | Top/Bottom |
| 4 | Veit [461] | 2015 | Siamese CNN, Heterogeneous dyads, 0.826 AUC | Outfit, Shirt/Jeans/Shoes |
| 5 | McAuley [8] | 2015 | CNN, Shifted sigmoid, Mahalanobis, 91.02% mAcc | Outfit, Amazon, Complements |
| 6 | Hu [462] | 2015 | Tensor Fact., Functional gradient descent, 0.251 NDCG | Top/Bottom/Shoes, FITB, Multi-modal |
| 7 | Huang [463] | 2016 | ResNet-50, Binary classifier, MLP, 84% Acc | Outfit scorer, Good/Bad |
| 8 | W. Zhou [435] | 2017 | NLP, NLTK, NN+Fuzzy Math., Score matrix, 96.94% HS | Feature/Category level mix & match |
| 9 | Qian [31] | 2017 | Seg., ASPP, CRF, R-FCN, k-means, VGG-16 | Top/Bottom, Complimentary color/pattern |
| 10 | Han [464] | 2017 | Inception-V3, Bi-LSTM, Multi-modal, 68.6% FITB | Multiple, FITB, Text, Image, Outfit scorer |
| 11 | Y. Li [465] | 2017 | CNN, Word2vec, AlexNet, GloVe, MLP, RNN, 36.4% AP | Outfit, Scorer, FITB |
| 12 | X. Zhang [36] | 2017 | Detection, mCNN-SVM, AlexNet, Label correlations | Street, Co-occurrence, Color, Location |
| 13 | Song [466] | 2017 | BPR Dual autoencoder, CNN, BoW, 0.7616 AUC | Top/Bottom |
| 14 | Yuan [467] | 2018 | Category2vec, Siamese, Metric learning, 3.72% mR@10 | Top/Bottom, Street/Shop |
| 15 | Hwangbo [440] | 2018 | Item-based CF, K-RecSys, Time discounting | Clicks, Sales, Preference |
| 16 | Y. Liu [126] | 2018 | Faster R-CNN, Dual Siamese, AlexNet, ~2.25 NDCG@10 | Top/Bottom, Street |
| 17 | Tangseng [468] | 2018 | ResNet-50, MLP, Beam search, 84.26% Acc | Multiple, Scorer, Item, Outfit generation |
| 18 | Z. Zhou [469] | 2018 | Hierarchical topic model, BoW, VGG-16, 45.4% mAP | Top/Bottom, Trends, Street |
| 19 | Strakovskaia [470] | 2018 | Inception, Random forest, Transfer Learn., 96% P@10 | Top/Bottom/Shoes, Lack of data problem |
| 20 | Valle [471] | 2018 | Semantic compositional Net., SkipGram, 72.4% R@10 | Semantics, Style, Occasions, Season |
| 21 | Vasileva [446] | 2018 | CNN, ResNet-18, Triplet, 65.0% FITB, 0.93 Comp. AUC | FITB, Item, Type-aware embedding |
| 22 | Sun [472] | 2018 | Siamese CNN, Probabilistic matrix factorization | Top/Bottom, Social circle, Style consistency |
| 23 | Z. Yang [346] | 2018 | Siamese, BPR, DCGAN, LSGAN, SE-Net, Inception-V3 | Outfit, Collocation item image synthesize |
| 24 | He [473] | 2018 | FashionNet, CNN, VGG, Rank loss, MLP, 81.82% R@10 | Top/Bottom/Shoes, Personalized, Item |
| 25 | Huynh [474] | 2018 | Adversarial feature transformer, Unsupervised, ~70% HS | Top/Bottom, Street |
| 26 | Hsiao [475] | 2018 | Subset selection, Correlated topic models, ResNet-50 | Capsule wardrobes, Item, Wild |
| 27 | Feng [476] | 2018 | Partitioned embedding, VAE, GAN, Composition graph | Outfit, Item, Interpretable, Trend |
| 28 | Song [477] | 2018 | Attentive knowledge distillation, CNN, BPR, Word2vec | Top/Bottom, Item |
| 29 | Nakamura [478] | 2018 | CNN, BiLSTM, VSE, Autoencoder, 73.2% FITB Acc | Multiple, FITB, Style-guided outfit |
| 30 | Dalmia [479] | 2018 | Social media mining, Encoder-decoder RNN, LSTM | Multiple, Item |
| 31 | L. Chen [480] | 2018 | Deep mixed-category metric learning, Triplet, 45% R@20 | Outfit, Mixed-category, Street |
| 32 | W. Chen [481] | 2019 | Multi-modal, TextCNN, CF, Transformer, 68.71% FITB | Outfit, Personalized, Alibaba |

| No | Article Reference | Year | Technical Keywords/Claimed Results | Application Notes |
|---|---|---|---|---|
| 33 | Lei [482] | 2019 | MF, Variant time SVD++, Hierarchical clustering | Item matching, User preference |
| 34 | Yin [483] | 2019 | CNN, AlexNet, Triplet, VBPR, 0.7077 mAUC | Top/Bottom |
| 35 | Gao [484] | 2019 | BPR-TAE, Siamese, AlexNet, Triple AutoEncoder, BoW | Top/Bottom |
| 36 | X. Yang [108] | 2019 | Tree-based model, GBDT, CNN, MLP, 50.66% Hit@10 | Item pair, Attribute-based, Interpretable |
| 37 | J. Liu [485] | 2019 | BPR-MAE, Multiple autoencoder, BoW, 0.8377 AUC | Top/Bottom, Bottom/Shoe, Multi-modal |
| 38 | Kang [486] | 2019 | ResNet-50, Local/Global compatibility, Triplet, 75.3% Acc | Scene-based, Street query-Shop item |
| 39 | Tan [417] | 2019 | SCE-Net, CNN, Condition weight branch, Triplet | Similarity conditions learning, Item |
| 40 | Lu [487] | 2019 | FHN, Binary code, HashNet, CNN, 64.61% FITB | Top/Bottom/Shoes, Personalized, Item |
| 41 | Yus. Lin [488] | 2019 | CNN, DenseNet, Xavier, 69.5% F1 | Outfit, Personalized, Scorer |
| 42 | Griebel [158] | 2019 | MaskRCNN, Bidirectional LSTM, VGG-16, Triplet | Social media, Detection, Style, Matching |
| 43 | Han [190] | 2019 | FiNet, Human parser, Encoder-decoder, VGG-19 | Fashion image inpainting, Compatibility |
| 44 | Stan [63] | 2019 | CNN, AlexNet, Two-stage, Category & Attribute | GUI, User-Item & Item-Item scores |
| 45 | Shin [489] | 2019 | Style features, Siamese, GoogLeNet, 0.8779 AUC | Top/Bottom/Shoes, Shop |
| 46 | K. Li [490] | 2019 | Multi-modal, ResNet18, BERT, FCNN, 62.8% FITB | Outfit, Natural language, Controllable |
| 47 | Cucurull [491] | 2019 | Graph auto-encoder, GCN, Metric learning, 62.2% FITB | Context-aware, FITB, Compatibility |
| 48 | Cui [492] | 2019 | Node-wise graph Neural Net., Multi-modal, 78.13% FITB | Outfit, Item, FITB, Compatibility |
| 49 | Bettaney [493] | 2019 | GORDN, Multi-modal, GloVe, LSTM, VGG, 0.75 mAUC | Top/Bottom/Shoes, Model images |
| 50 | Kumar [351] | 2019 | $c^{+}$GAN, ResNet-50, DCT, Faster R-CNN | Shirt/Pants, Synthesis, Street |
| 51 | Yuj. Lin [352] | 2019 | Variational transformer, DCNN, BoW, 74.5% AUC | Top/Bottom, Image+Text, Synthesis |
| 52 | Polania [494] | 2019 | Siamese, VGG-16, Color Hist., 4.42X P@12 than random | Substitute, Complementary, Item |
| 53 | Wu [495] | 2019 | Sampling, ZSF-c, STAMP, CNN, FDNA, 29.41% R@5 | Session-based, Personalized, Shop, Zalando |
| 54 | Kuhn [496] | 2019 | Neural Net. Word2vec, Attention mechanism, 36.6% AP | Pair/Outfit generation, Shop |
| 55 | Wang [497] | 2019 | Multi-Layered comparison Net., CNN, MLP, 64.35% FITB | Outfit, Comp. prediction/diagnosis/revision |
| 56 | Song [498] | 2019 | GP-BPR, CNN, TextCNN, BPR, MLP, 0.8388 AUC | Top/Bottom, Item, Personalized |
| 57 | Dong [324] | 2019 | PCW-DC, BPR, Bi-LSTM, MLP, Body shape modeling | Capsule wardrobe, Personalized, Shop |
| 58 | Yu [355] | 2019 | VGG-16, LSGAN, Encoder-decoder, Siamese | Top/Bottom, Synthesis, Personalized |
| 59 | X. Yang [499] | 2019 | TransNFCM, TextCNN, AlexNet, Triplet, 38.1% Hit@10 | Translation-Based, Category Comp., Item |
| 60 | X. Liu [115] | 2020 | ResNet50, Metric learning, 55.6% FITB, 0.85 Comp. AUC | FITB, Outfit scorer, Shop, MMFashion |
| 61 | Yuj. Lin [500] | 2020 | NOR, Mutual attention, MLP, GRU, RNN, 12.51% mAP | Top/Bottom, Comment generator, Item |
| 62 | E. Li [501] | 2020 | Unified embedding, SE-ResNext101, Triplet, 68.8% R@10 | Outfit, Complete The Look, Pinterest, Item |
| 63 | Denk [502] | 2020 | Contextual BERT, Global state, 29.40% R@5 | Outfit, FITB, Item |
| 64 | Jo [386] | 2020 | Implicit profiling, CNN, cGAN, Ranking loss, 80.9% P@10 | Top/Bottom, Shop, Also sketch retrieval |
| 65 | Sarmiento [360] | 2020 | VAE, Log-Likelihood, K-Means, Fixed-epsilon sampling | Using synthesized item images |
| 66 | Y. Lin [503] | 2020 | CNN, Category-based subspace Attn. Net., 63.73% FITB | Multiple, FITB, Compatibility, Item |
| 67 | Moosaei [504] | 2020 | Relation Net., FashionRN-VSE, DenseNet, 0.88 AUC | Multiple, FITB, Scorer, Item |
| 68 | De Divitiis [505] | 2020 | Memory augmented Net., MF, Best-of-K, 45% Acc@10 | Top/Bottom, Item |
| 69 | X. Li [506] | 2020 | Hierarchical graph Net., Self-attention, BPR, 87.97% FITB | Outfit, Personalized, Item, FITB |
| 70 | S. Liu [507] | 2020 | Adversarial inverse RL, MVAE, BERT, MDP, ~43% mAP | Top/Bottom/Shoes, Street/Shop, Text |
| 71 | H. Zhang [508] | 2020 | Graph, Color palette, K-means, Pseudo label, 59.9% FITB | Multiple, Color compatibility, Item |
| 72 | Liu [509] | 2020 | Neural graph filtering, CNN, Aggregation, 58.8% FITB | Multiple, Item, Diverse |
| 73 | Sun [510] | 2020 | VSFM, LSTM, CNN, ResNet, Fusion, Triplet, 0.968 AUC | Top/Bottom, Item, Multi-modal (Text) |
| 74 | X. Yang [511] | 2020 | Deep relational embedding Propa., Graph, 73.1% mR@5 | Outfit, Personalized |
| 75 | X. Yang [512] | 2020 | Mixed category attention, Tuple triplet, 84.13% mFITB | Multiple, Controllable, Alternative, Text |
| 76 | Sagar [513] | 2020 | PAI-BPR, Attr.-aware, Nwjc2vec, AlexNet, 0.8502 AUC | Top/Bottom, Personalized, Item, Attr. |
| 77 | Zou [514] | 2020 | CNN, ResNet-18, Grad-CAM-like, Manual decision tree | Top/Bottom, Comment generator, Item |
| 78 | Kim [515] | 2020 | Self-supervised, Shapeless local patch, 55.8% FITB | Unsupervised, Color, Texture |
| 79 | Lai [516] | 2020 | CNN, Theme Attention, Res. block, Triplet, 76.87% FITB | Theme-aware, Occasion, Fit, Style, Gender |
| 80 | Tangseng [517] | 2020 | Dominant color, Canny edge, K-mean, CNN, 76.36% Acc | Outfit, Flaw detection, Scorer, Explainable |

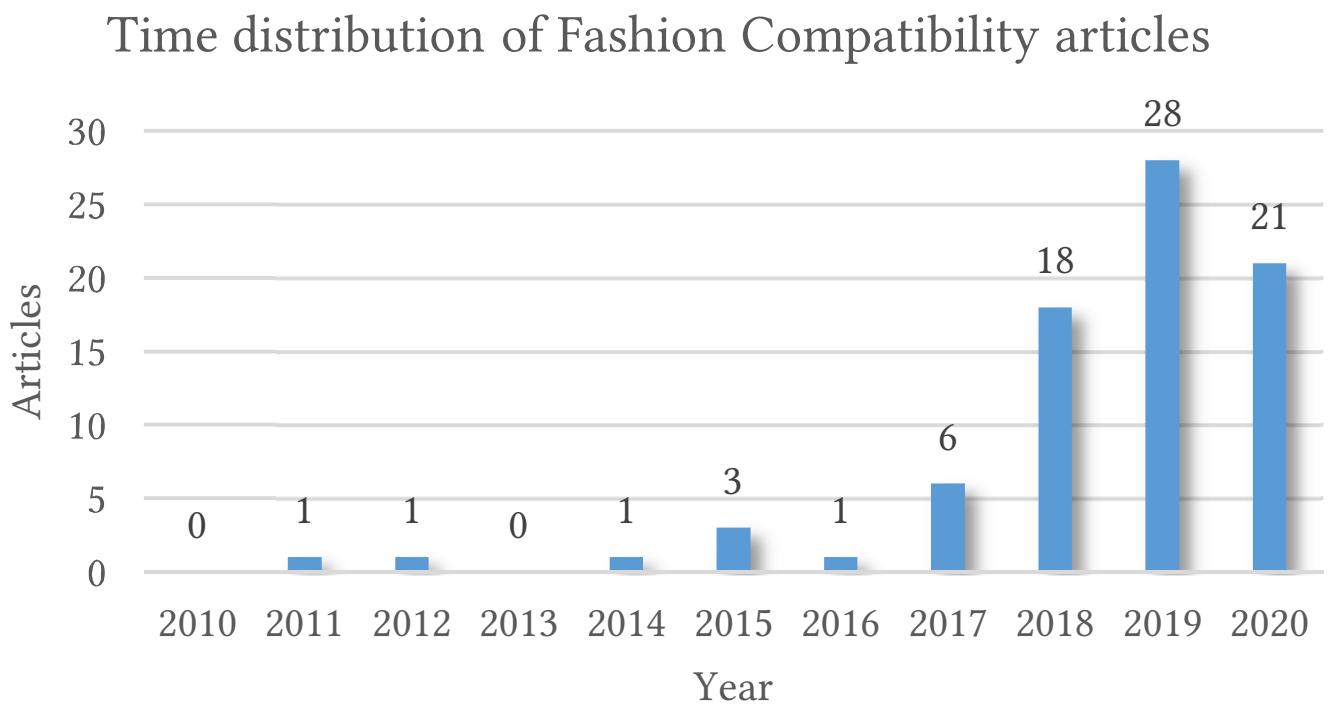

Figure 57: Time Analysis of Fashion Compatibility Articles

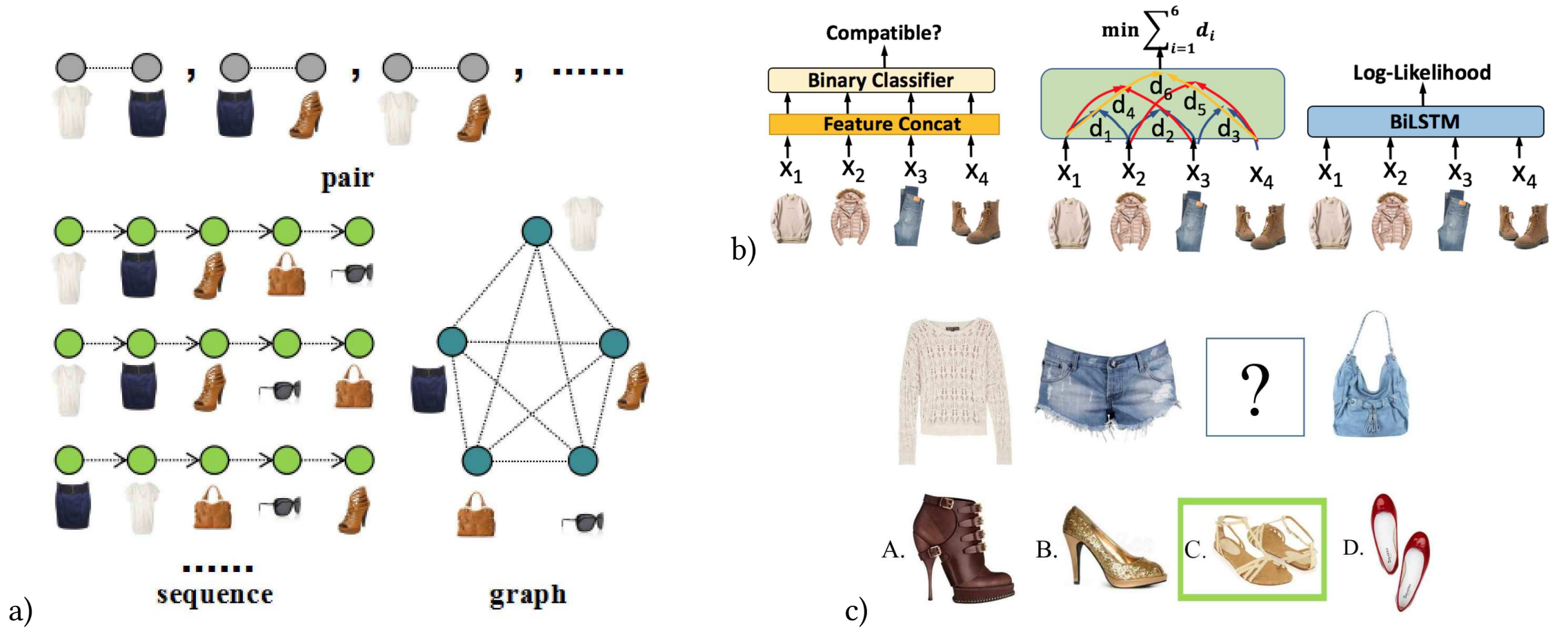

Figure 58: a) Different outfit representations [492] b) Different models for compatibility learning [512] c) Fill In The Blank [497]

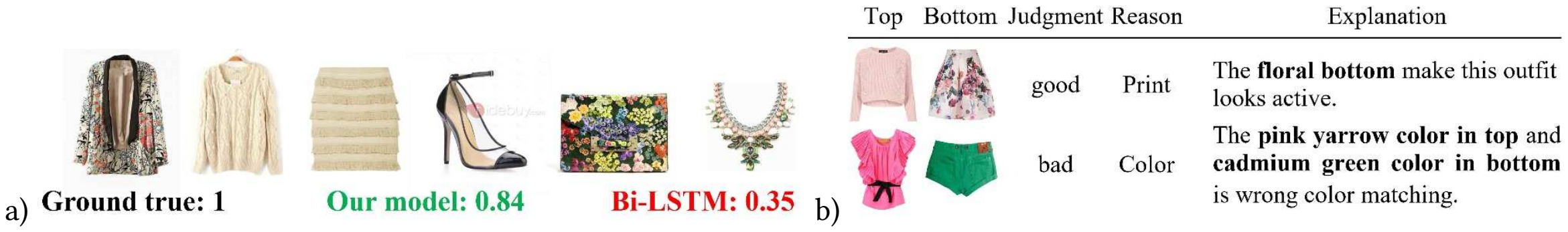

Figure 59: a) Outfit compatibility scoring [492] b) Explainable outfit compatibility evaluation [514]

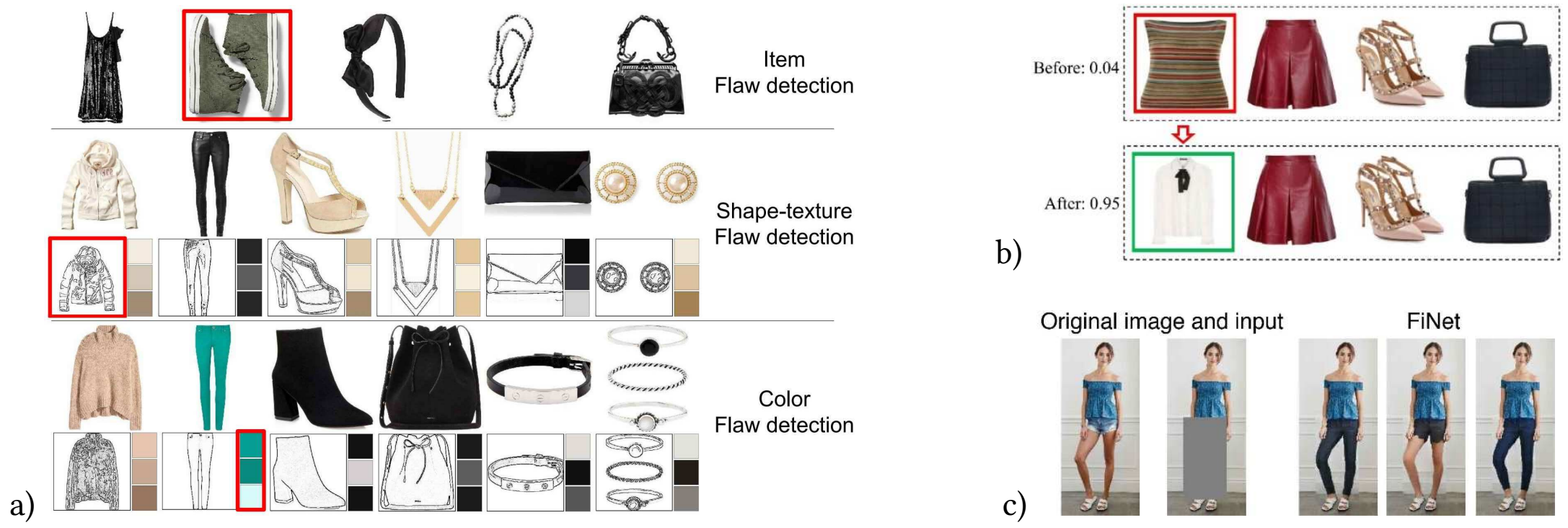

Figure 60: a) Outfit flaw detection [517] b) Outfit revision to improve compatibility [497] c) Compatible fashion inpainting [190]

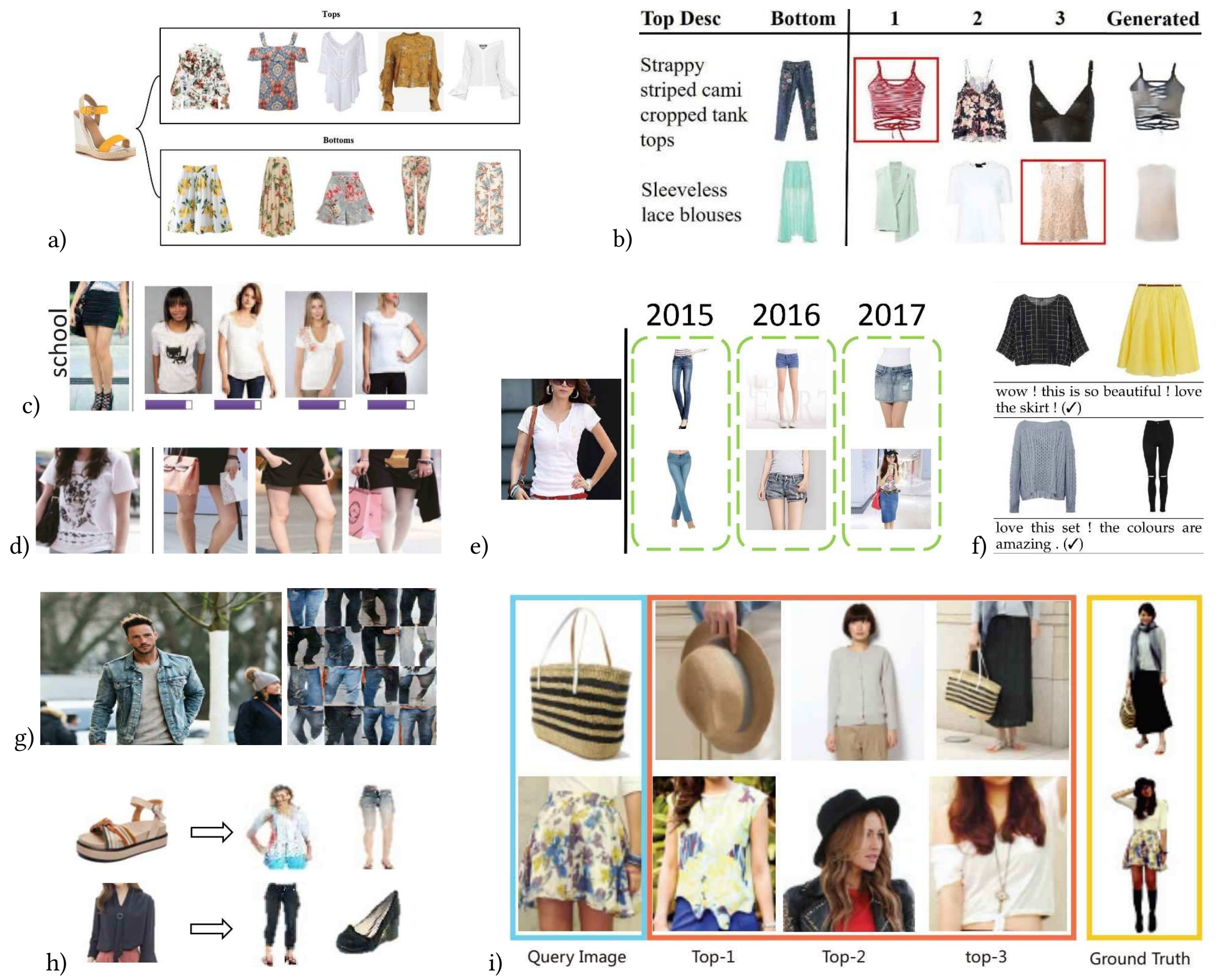

Figure 61: Single-product-based compatible item recommenders. a) For Item images [417] b) Text-guided, with synthesis [352] c) Scenario-oriented [83] d) For Street images [126] e) Trend-aware [469] f) Explainable via comment generation [500] g) Top/Bottom synthesis for Street images [351] h) Top/Bottom/Shoe synthesis for Shop images [346] i) Mixed-category set [480]

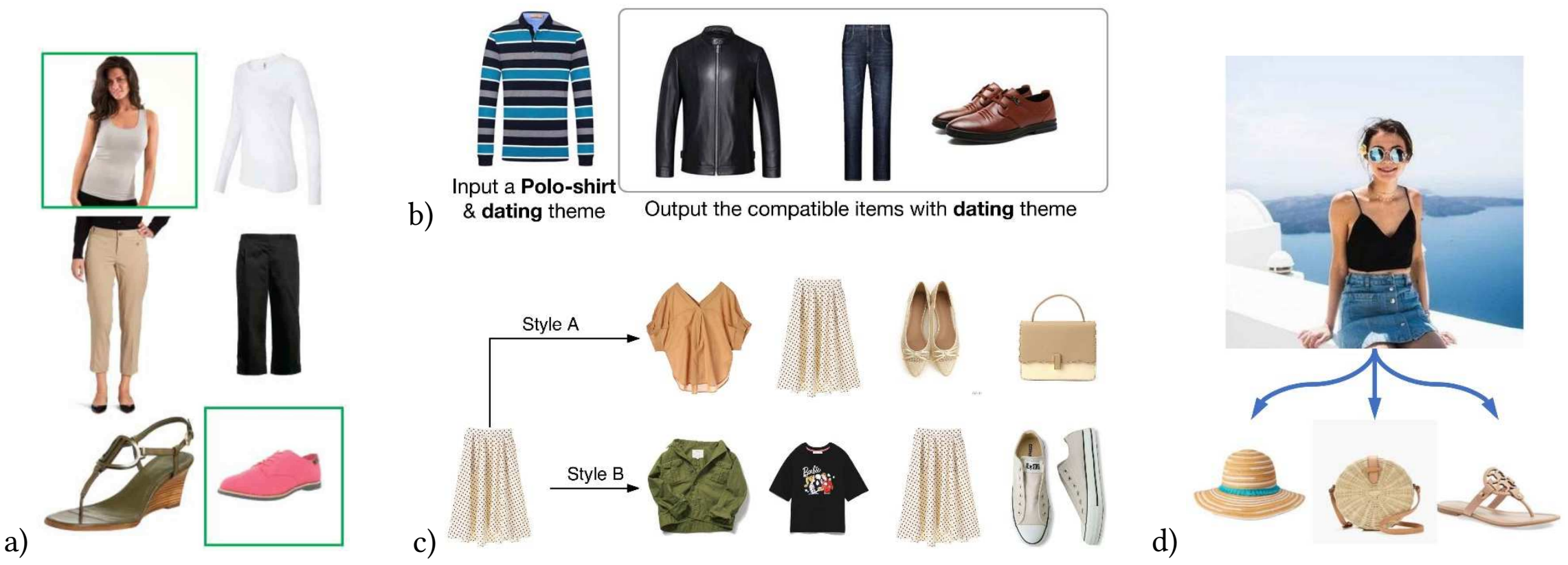

Figure 62: Single-product-based compatible item recommenders. a) Top/Bottom/Shoe [461] b) Theme-aware [516] c) Style-aware [478] d) Scene-aware [486]

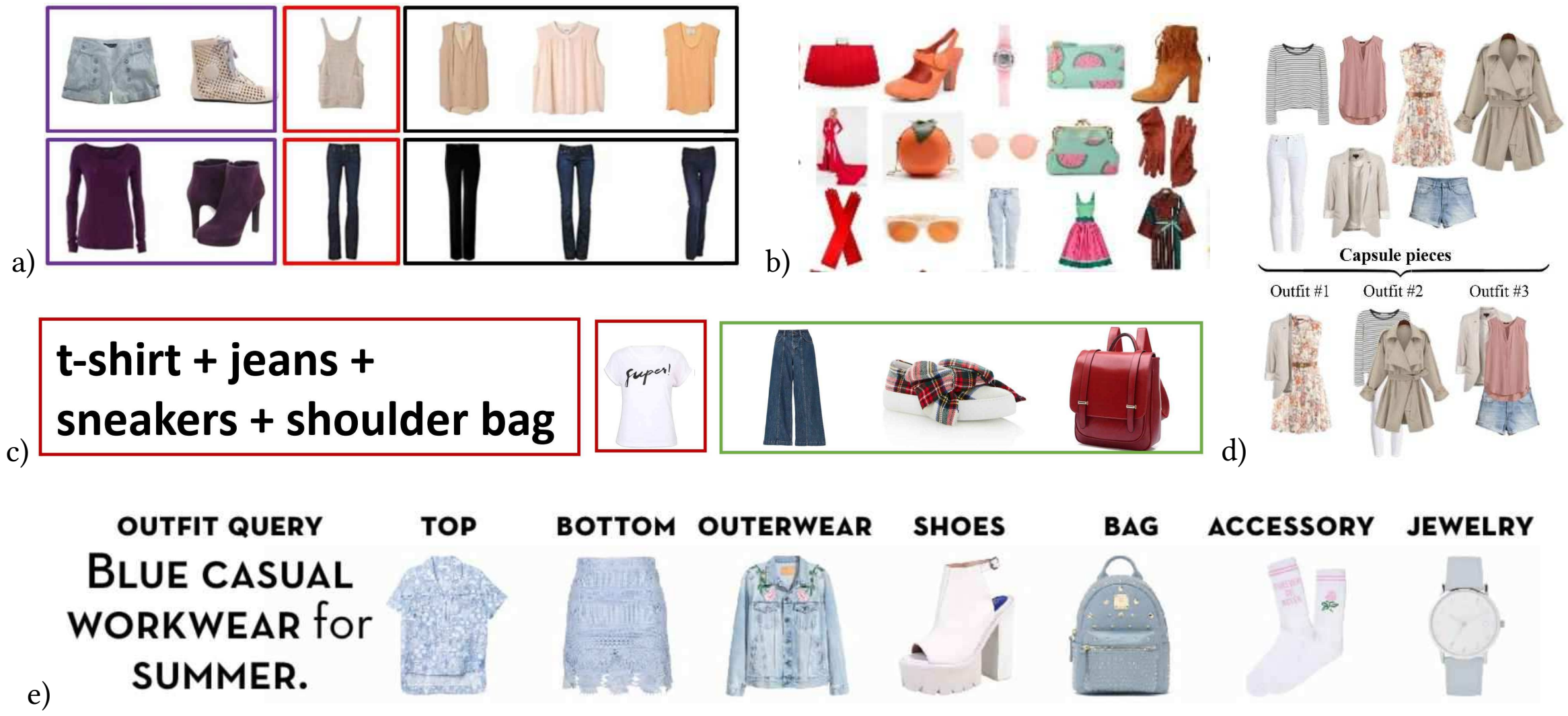

Figure 63: a) Multi-product-based compatible item Recom. [462] b) Color compatibility [508] c) Conditional compatible item Recom. [512] d) Capsule wardrobe [475] e) Natural language sentence-guided controllable outfit Recom. [490]

### *2.7.4 Personalized Recommenders*

These systems primarily focus on the users' preferences to build their recommendation list. It is noteworthy that all recommender systems implicitly use some data to personalize their recommendations, but this section is devoted to the strategies that target the user preference or the users' history to tailor unique results for each user.

Table 19: Articles Related to Personalized Recommenders

| No | Article Reference | Year | Technical Keywords/Claimed Results | Application Notes |
|---|---|---|---|---|
| 1 | Yu-Chu [456] | 2012 | Classic, Modified Bayesian Net., User feedback | Outfit Recom., Usage history, Feedback |
| 2 | R. He [431] | 2015 | VBPR, Matrix factorization, Deep CNN, 0.7364 mAUC | Visual appearance, Shop |
| 3 | McAuley [8] | 2015 | CNN, Shifted sigmoid, Mahalanobis, 91.15% mAcc | Co-purchases, Amazon, Shop |
| 4 | Woiceshyn [518] | 2017 | Learning-based personalization, Android, MLR, SGD | Social robot, User storage, GUI, Activity |
| 5 | Liu [519] | 2017 | DeepStyle, CNN, BPR, Style features, 0.7961 AUC | Style, Item |
| 6 | Kang [339] | 2017 | DVBPR, Siamese CNN, MF, GAN, BPR, 0.7547 mAUC | User ratings, Shop |
| 7 | Packer [442] | 2018 | MF, BPR, I-TVBPR, Temporal dynamics, 0.7215 AUC | Users' visual preferences, Time |
| 8 | P. Li [419] | 2018 | User-based CF, CB, FAST, PCA, K-Means, 26.70% HS | Offline shopping |
| 9 | Agarwal [520] | 2018 | CF, ALS-MF, BPR, 4.37% mAP@15 | Browsing behavior |
| 10 | T. Yang [443] | 2018 | Classic, Knowledge base, Matching rules | Expert knowledge, Age, Body, Skin, Color |
| 11 | Sun [472] | 2018 | Siamese, GoogleNet, Probabilistic matrix factorization | User social circle, Style consistency |
| 12 | X. Chen [521] | 2018 | Attentive NN, CF, VGG-19 , GRU, 1.21% mF1@5 | User history, Textual review, Explainable |
| 13 | Z. Yang [346] | 2018 | Siamese, BPR, DCGAN, LSGAN, SE-Net, 0.769 AUC | User rating, Collocation image synthesize |
| 14 | T. He [473] | 2018 | FashionNet, CNN, VGG, Rank loss, MLP, 81.82% R@10 | User-specific preferences, Outfit |
| 15 | Hou [448] | 2019 | SAERS, CNN, Siamese, Grad-AAM, ROI, ResNet-50, BPR | Shop image+User history, Explainable |
| 16 | W. Chen [481] | 2019 | POG, FOM, TextCNN, CF, Transformer, ~22.5% CTR | User clicks, Outfit, Alibaba |
| 17 | Lei [482] | 2019 | MF, Variant time SVD++, Hierarchical clustering | User preference, Item matching |
| 18 | Lu [487] | 2019 | FHN, Binary code, BPR, HashNet, CNN, 0.9156 mAUC | User-outfit, Top/Bottom/Shoes |
| 19 | Lin [488] | 2019 | CNN, DenseNet, Xavier, 69.5% F1 | Personal outfit scorer |
| 20 | X. Chen [522] | 2019 | VECF, VGG-19, LSTM, GRU, 3.65% mF1@10 | User history, Textual review, Explainable |
| 21 | J. Wu [495] | 2019 | ZSF-c, STAMP, CNN, FDNA, +6.23% CTR improvement | Session-based, Shop, Zalando |
| 22 | Song [498] | 2019 | GP-BPR, CNN, TextCNN, BPR, MLP, 0.8388 AUC | User history, Outfit, Top/Bottom |
| 23 | Dong [324] | 2019 | PCW-DC, BPR, Bi-LSTM, MLP, 80.56% Success rate | Personalized capsule wardrobe, Body shape |
| 24 | Yu [355] | 2019 | VGG-16, LSGAN, Siamese, 4.262 IS | Compatible item, User preference, Synthesis |
| 25 | X. Li [506] | 2020 | Hierarchical graph, Self-attention, BPR, 28.33% Hit@10 | User-Item-Outfit relation, FITB |
| 26 | Sagar [513] | 2020 | PAI-BPR, Attr.-aware, Nwjc2vec, AlexNet, MLP | User-item interaction, Compatibility |
| 27 | Zheng [459] | 2020 | Multi-modal, VGG-19, Seg., Triplet, MLP, 27.98% R@10 | Users' social media, Street images, Hashtags |
| 28 | Q. Wu [523] | 2020 | VTJEI, Bidir. two-layer adaptive attention, 22.29% Hit@10 | User history, Textual review, Explainable |
| 29 | Su [524] | 2020 | Multiclass SVM, Hybrid RCNN, LGBPHS, 85.4% AP | Users' facial expressions, Emotions |
| 30 | Mohammadi [122] | 2021 | ResNet-50, DenseNet, Clustering, ~25% R@10 | User history, Textual reviews, Ratings |

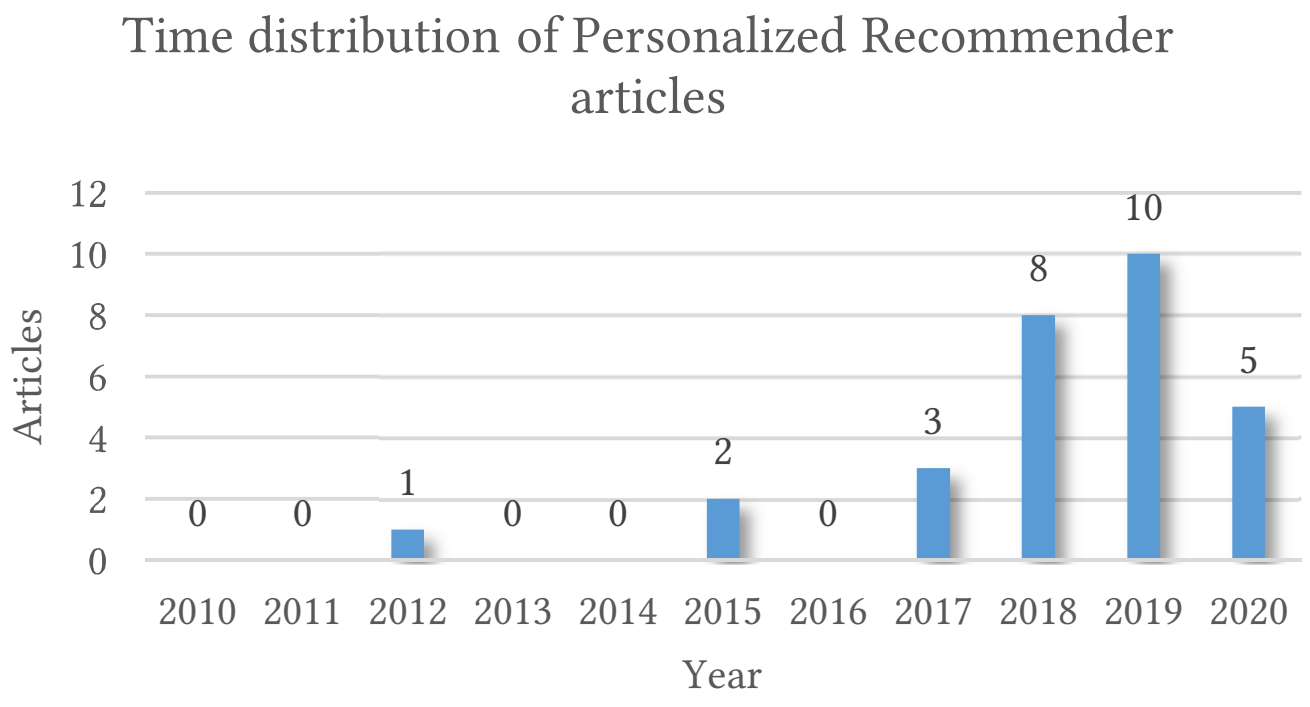

Figure 64: Time Analysis of Personalized Recommender Articles

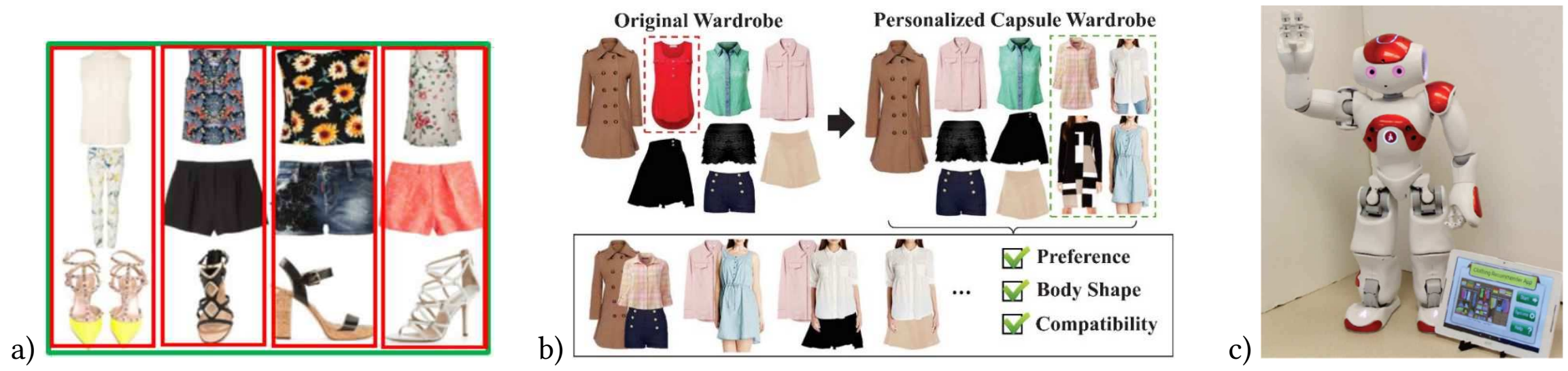

Figure 65: a) Personalized outfit Recom. [473] b) Personalize capsule wardrobe [324] c) Recom. for social robot [518]

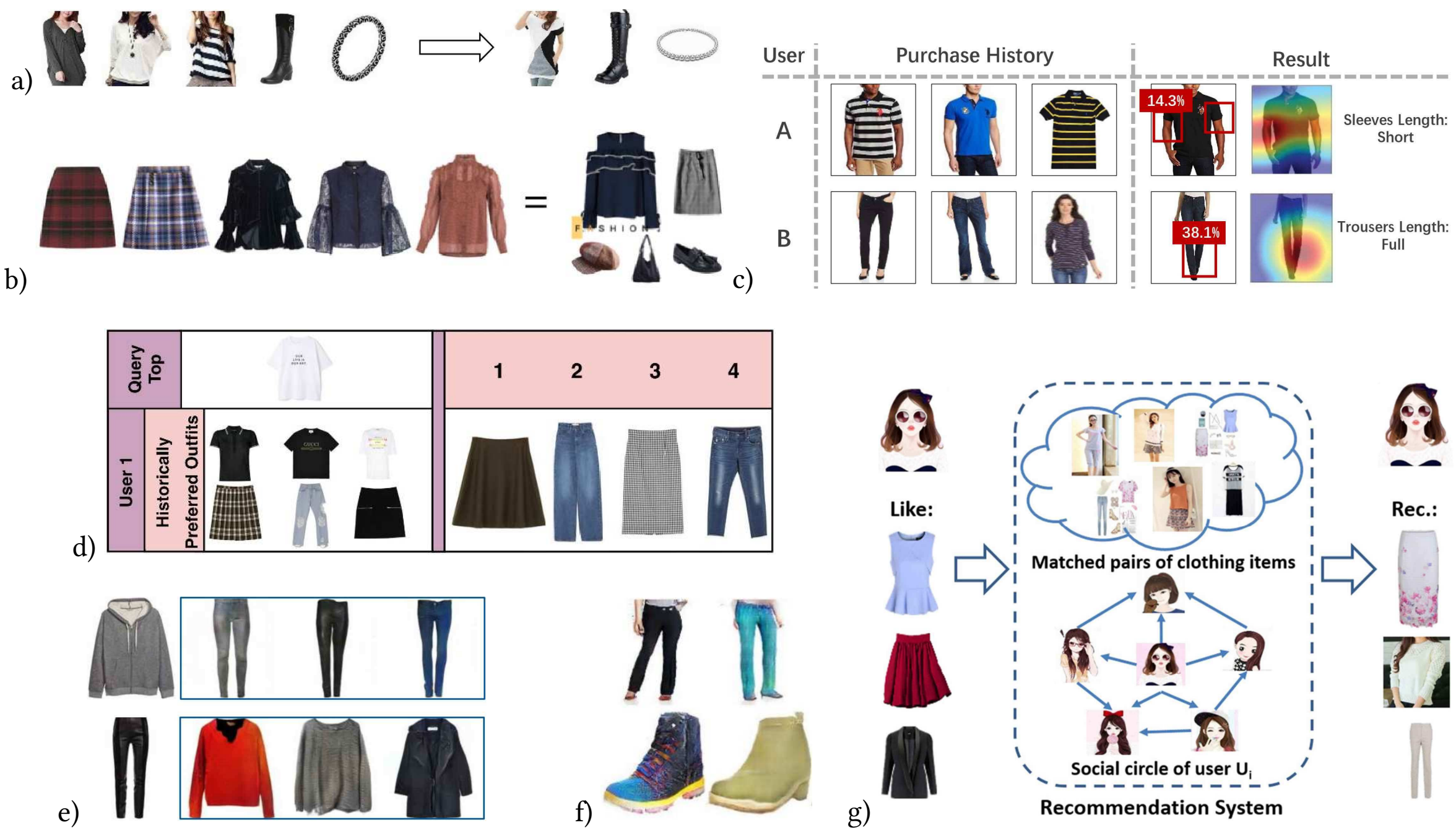

Figure 66: Personalized recommendations based on user history. a) Similar item Recom. [346] b) Alibaba iFashion outfit Recom. [481] c) Visually explainable item Recom. [448] d) Top/Bottom compatibility Recom. [513] e) Top/Bottom compatibility synthesis [355] f) Recom. synthesis [339] g) Compatibility Recom. combined with the user's social circle [472]

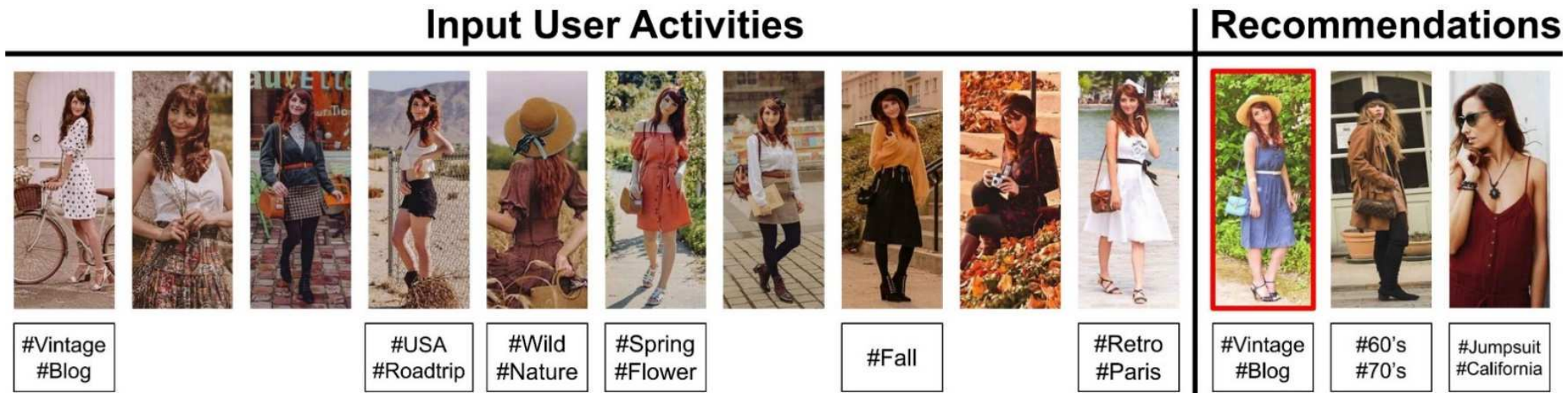

Figure 67: Personalized Recom. using user's social media pictures and hashtags [459]

## 2.8 Fashion Analysis & Trends

Some studies focus on fashion analysis, delving deep into fashionability, aesthetics, popularity, geographic analysis of the perception of fashion and beauty, the effects of fashion shows on real-life street fashion, and other related subjects. AI systems can also significantly help with fashion trends forecasting (such as color trends, seasonal trends, popularity, and regional trends), sales/demand prediction, and all kinds of fashion data analysis.

Table 20: Articles Related to Fashion Analysis & Trends

| No | Article Reference | Year | Technical Keywords/Claimed Results | Application Notes |
|---|---|---|---|---|
| 1 | Ni [525] | 2011 | Two-stage dynamic model, Autoregressive decision tree | Sales forecasting |
| 2 | Yu [526] | 2012 | Systematic comparison of ARIMA, ANN, GM, GRA-ELM | Color trend forecasting |
| 3 | Choi [527] | 2012 | Comparison of: ANN, GM, Markov regime-switching | Color trend forecasting, Very few data |
| 4 | Q. Chen [528] | 2013 | Active clustering, Window search, Latent structural SVM | Which makes dresses fashionable? |
| 5 | Nenni [529] | 2013 | Short review, Analysis of the products & approaches | Demand forecasting |
| 6 | Kiapour [20] | 2014 | Pose estimation, Style indicators, Linear kernel SVM | Discovering the elements of styles in outfits |
| 7 | Yamaguchi [530] | 2014 | TF-IDF, Style descriptor, 81.65% mAcc@75% | Visual popularity in social networks |
| 8 | Hidayati [531] | 2014 | Classic, Face Det., HSV, HOG, KNN, SVM, 80.63% Acc | New York Fashion Trends, Season, Catwalk |
| 9 | Choi [532] | 2014 | Book: Review, Methods, Applications | Intelligent fashion forecasting systems |
| 10 | Simo-Serra [457] | 2015 | Conditional random field, BoW, DNN, 17.36% IOU | Fashionability analysis, Country, Income |
| 11 | Vittayakorn [23] | 2015 | Classic features, KNN retrieval, Semantic parse, SVM | Influence of runways on street fashion |
| 12 | K. Chen [90] | 2015 | Classic, SIFT, Attr. learning, Pose Est., SVM, CRF | Influence of runways on street fashion |
| 13 | Wang [533] | 2015 | Classic, Feature selection, Color harmonic templates, SVR | Shopping photos aesthetic quality predictor |
| 14 | He [433] | 2016 | One-Class CF, CNN, Temporal dynamics, TVBPR+ | Visual evolution of fashion trends |
| 15 | Y. Liu [331] | 2016 | Bimodal deep autoencoder guided by correlative labels | Aesthetic rules, Top/Bottom influence |
| 16 | Jia [534] | 2016 | Stacked DAE, SVM, Correlative labels, 0.2366 MAE | Mapping visual features to aesthetic words |
| 17 | Zou [535] | 2016 | SIFT, RCC, BoW, IFV, Clustering, CN, LBP | Effects of style/color/texture on fashion |
| 18 | Park [536] | 2016 | ML, Vader, Decision tree, Random Forest, AdaBoost | Predict fashion model success, Instagram |
| 19 | Al-Halah [537] | 2017 | CNN, AlexNet-like, NMF, Exponential smoothing model | Forecast visual style popularity in fashion |
| 20 | Matzen [538] | 2017 | CNN, GoogLeNet, Isotonic regression, Clustering, PCA | Exploring worldwide clothing styles |
| 21 | K. Chen [98] | 2017 | Pose Est., VGG-16, SIFT, SVM, CRF, 62.6% Acc | Attribute popularity seasonal trends |
| 22 | Aghaei [539] | 2017 | Social signal processing, Brunswik lens model | Influence of clothing on people's impression |
| 23 | Ma [540] | 2017 | Bimodal correlative deep autoencoder, Decision tree | Style analysis, Trend, Co-occurrence |
| 24 | Takagi [37] | 2017 | CNN, VGG, Xception, Inception, ResNet50, 72% mAcc | Style analysis, What makes a style |
| 25 | Ha [541] | 2017 | CNN, ResNet50, Multi-label classification | Fashion conversation data on Instagram |
| 26 | Gu [38] | 2017 | QuadNet, Neighbor-constrained, SVM, t-SNE | Fashion trends analysis, Street |
| 27 | Abe [542] | 2017 | Fashion trend descriptor, StyleNet, BoW | Fashion trends analysis, Cities, Street |
| 28 | Chang [543] | 2017 | DNN, Prize-collecting Steiner tree, VGG19, ILP | Fashion world map, World trends, Colors |
| 29 | R. Liu [544] | 2017 | Systematic coding scheme, Image content+element | Style bloggers Analysis, Instagram |
| 30 | Vittayakorn [545] | 2017 | AlexNet, VGG, SVM, SVR, 11.54 MAE years | Temporal analysis, Production date |
| 31 | Packer [442] | 2018 | MF, BPR, I-TVBPR, Temporal dynamics, 0.7215 AUC | Fashion trends analysis and tracking |
| 32 | Tang [546] | 2018 | Group decision-making, Ordinal consensus, HFLPRs | Fashion sales forecasting |
| 33 | Jiang [547] | 2018 | Probabilistic linguistic linear least absolute regression | Fashion trend forecasting |
| 34 | Mall [548] | 2019 | Trust region reflective, TF-IDF, CNN, GoogLeNet | World temporal trends, Events |
| 35 | Kataoka [549] | 2019 | Fashion style distribution, K-means, StyleNet + SVM | World-wide fashion culture analysis, FCDB |
| 36 | Ma [64] | 2019 | CNN, Bi-LSTM, ResNet-18, Weak label modeling | Fashion knowledge analysis, Social media |
| 37 | Lo [550] | 2019 | Deep temporal sequence, LSTM, InceptionV3, Word2Vec | Style popularity analysis |
| 38 | Mall [551] | 2020 | Multi-task CNN, K-means, Analogy-inspired encoding | Underground neighborhood maps of cities |
| 39 | Al-Halah [552], [553] | 2020 | GoogLeNet, ResNet-18, NMF, Granger causality test, MLP | Fashion style influences around the world |
| 40 | Getman [554] | 2020 | ML classifier, Pattern recognition, 92.18% Acc | Fashion item trend tracking, Baseball cap |
| 41 | Shi [121] | 2020 | Faster R-CNN, Segmentation, 75% Acc | Trend analysis, Fashion show videos |

| No | Article Reference | Year | Technical Keywords/Claimed Results | Application Notes |
|---|---|---|---|---|
| 42 | Hsiao [555] | 2021 | Mask-RCNN, ResNet-18, LDA, Granger-causality test | World events timeline from fashion images |

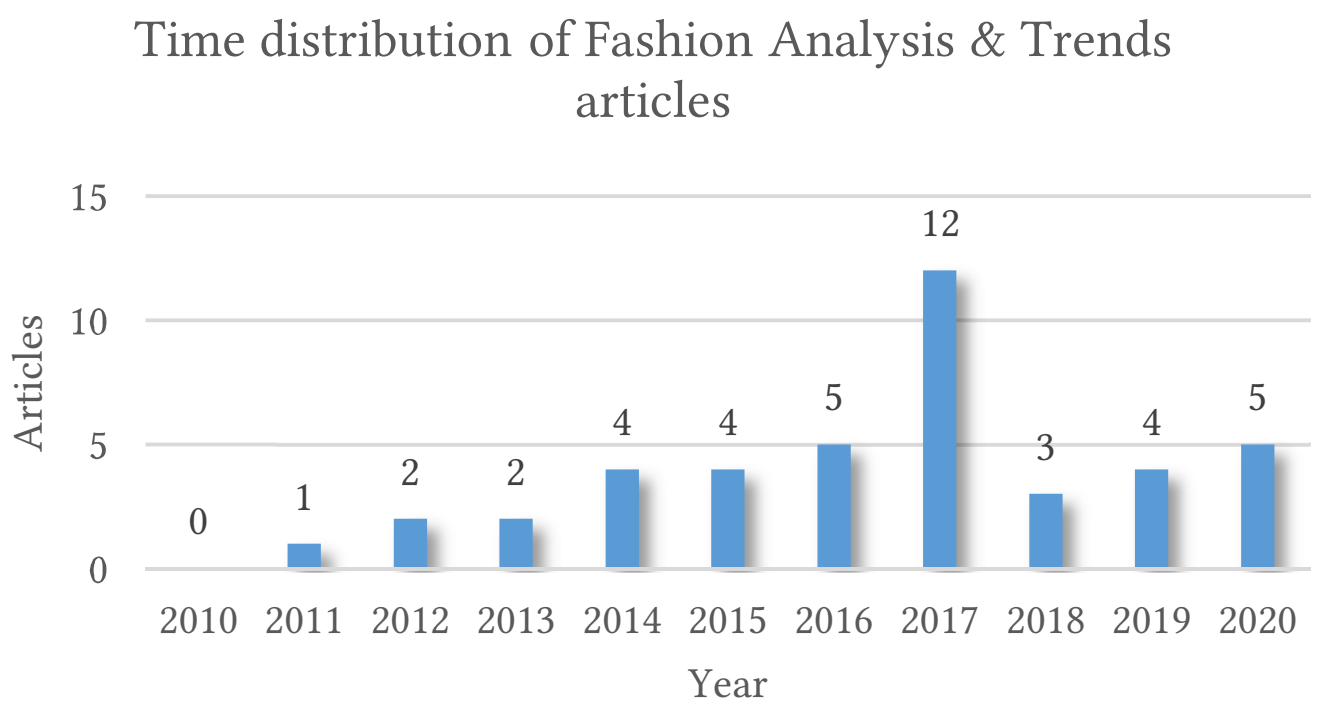

Figure 68: Time Analysis of Fashion Analysis & Trends articles

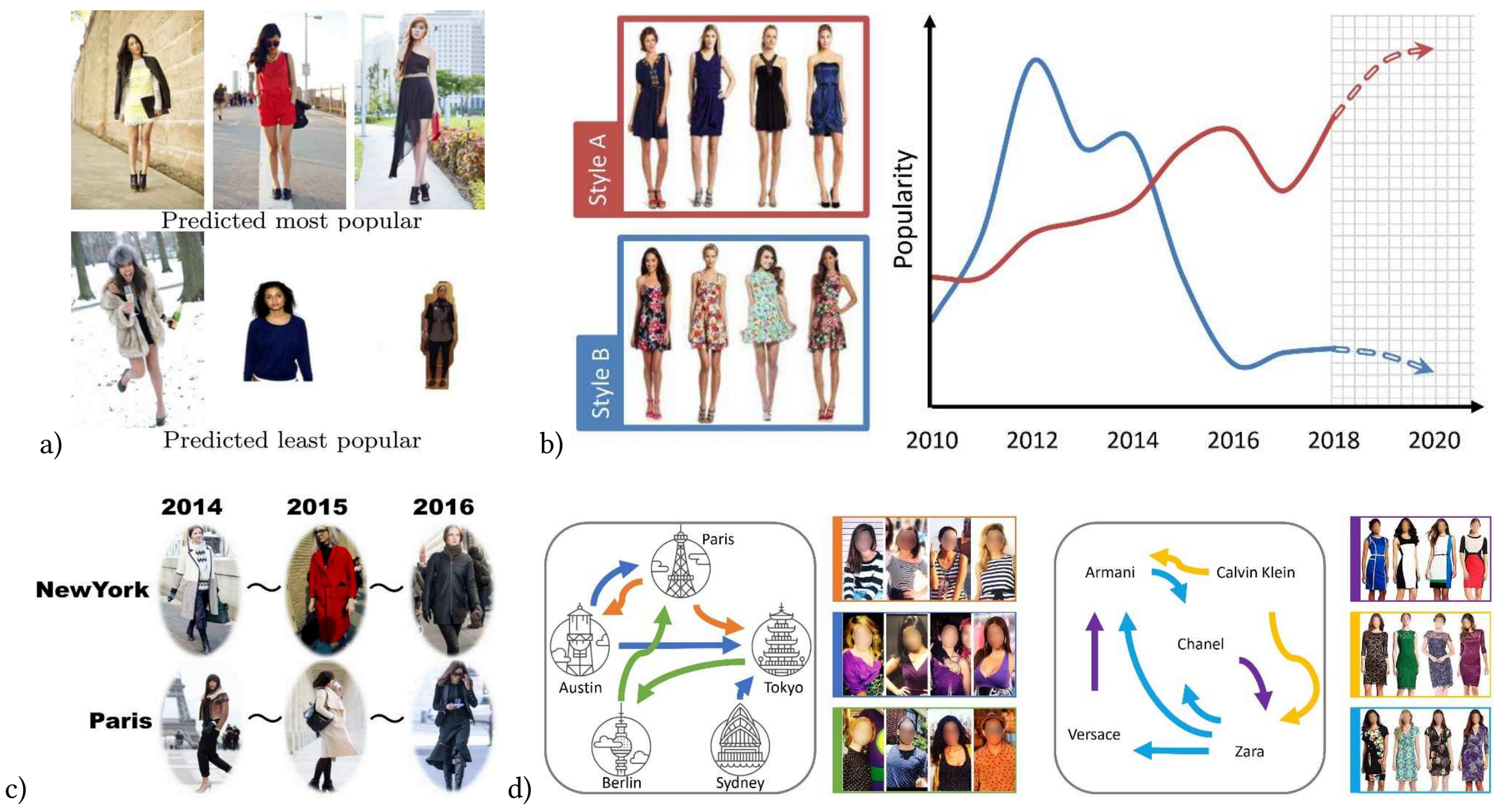

Figure 69: a) Visual popularity analysis [530] b) Style popularity trend forecasting [537] c) Fashion trends in different cities [542] d) Fashion influence of different cities and different brands [553]

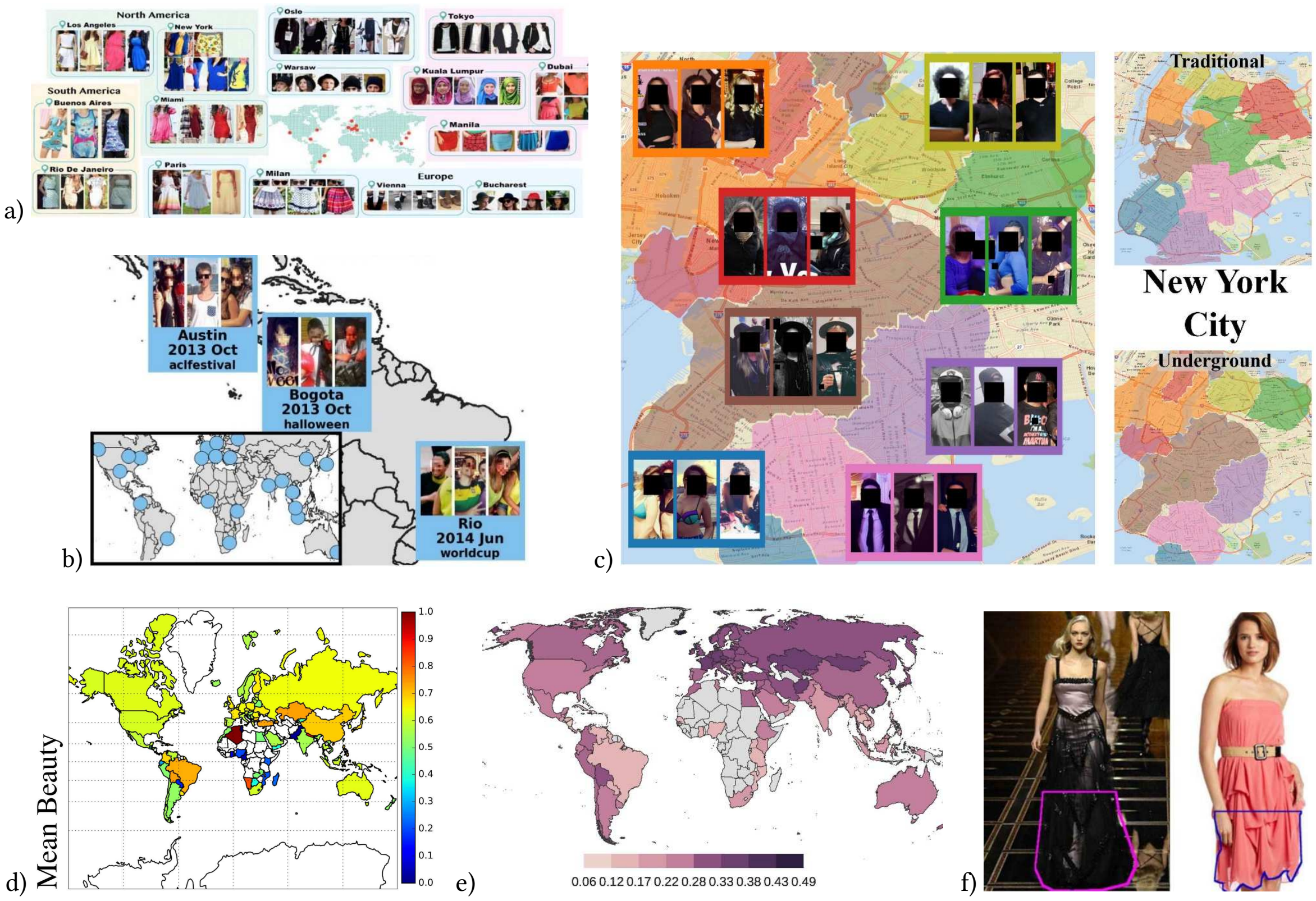

Figure 70: a) Fashion world map [543] b) Fashion events around the world [548] c) New York city fashion map [551] d) Worldwide fashionability and beauty map [457] e) Item (jacket) worldwide trend and frequency analysis [538] f) Detect visual fashionability factors [528]

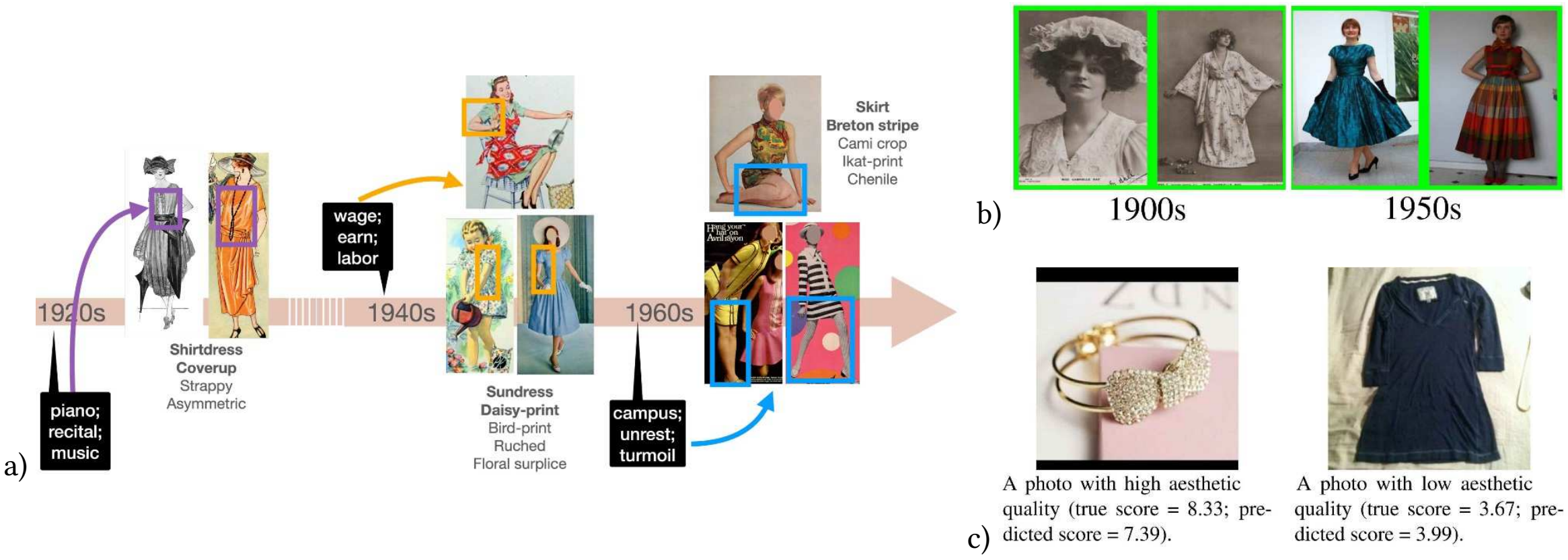

Figure 71: a) Part of a fashion history timeline [555] b) Temporal estimation for fashion trend analysis [545] c) Aesthetic quality assessment of online fashion shopping photos [533]

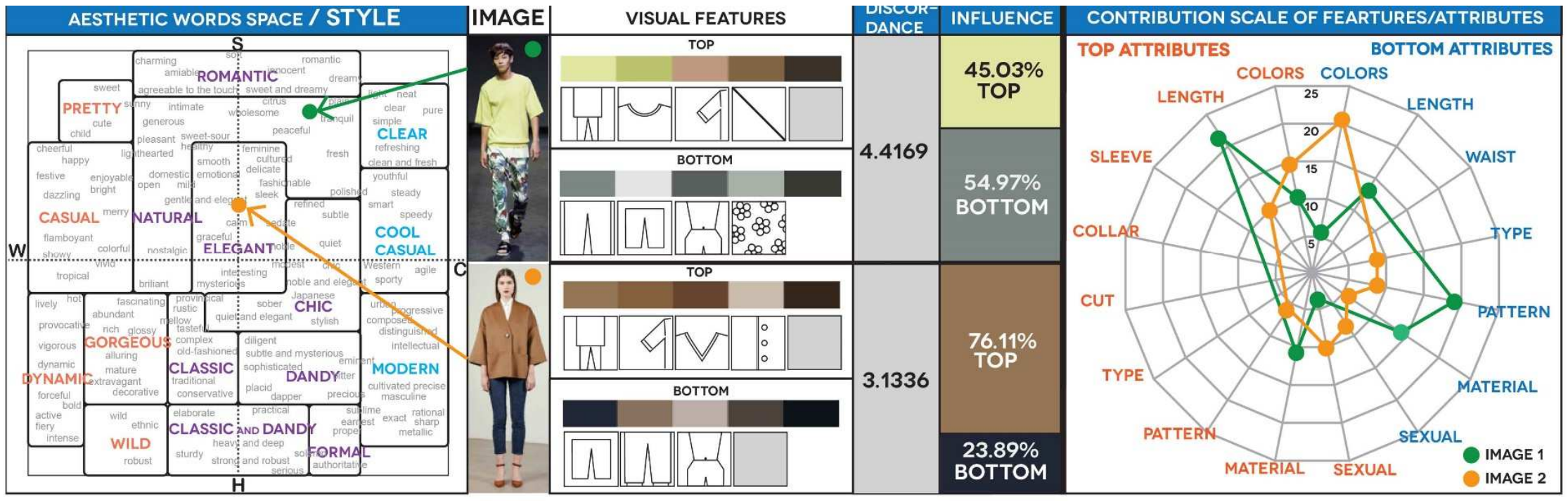

Figure 72: An example of fashion analysis including style, color, compatibility, attributes, etc. [331]

## 2.9 Production, Quality & Inspection

Computer, machine learning, and AI systems can shape apparel and textile production, introducing new, complex, more optimized, and environment-friendly fashion items. These systems are also used in factories' apparel production lines to check the quality and inspect the materials. Although these applications are out of our focus and this paper does not fully cover them due to the vast domain of such industrial applications of AI, this section presents some examples in Table 21. One can refer to [556], a review study dedicated to this matter published in 2011, for more information. The research mentioned earlier studies 95 articles focusing on AI applications in various domains of the apparel industry, including design, manufacturing, retailing, and supply chain management.

Table 21: Articles Related to Production, Quality & Inspection

| No | Article Reference | Year | Technical Keywords/Claimed Results | Application Notes |
|---|---|---|---|---|
| 1 | Satam [557] | 2011 | Intelligent design systems, CAD, CAM, CAPP | 2D/3D Garment mass customization |
| 2 | Gale [558] | 2018 | Influence of AI, big data, and new textile technologies | Complex textile, Waste management |
| 3 | Guo [559] | 2018 | Hybrid intelligent optimization framework | Optimized production/delivery operations |
| 4 | Wei [560] | 2018 | Faster RCNN, VGG16, Region proposal Net., 95.8% Acc | Fabric defect detection |
| 5 | Lv [561] | 2018 | Cartoon-texture decomposition, DCNN | Fabric defect detection |
| 6 | X. Wang [376] | 2018 | CNN, Inception-ResNet-v1, SqueezeNet, 99.89% Acc | Fabric identification |
| 7 | Tong [562] | 2018 | Optimal Gabor filtering, Adaptive threshold, CoDE | Striped fabric defect detection |
| 8 | Meng [377] | 2018 | Classic, Robust feature extraction, Color, Edge | Material image retrieval |
| 9 | Zhou [563] | 2018 | Rough possibilistic clustering, Shadowed Set, RCM | Fabric image segmentation |
| 10 | Gao [564] | 2018 | CNN, Binary classification, 96.52% Acc | Woven fabric defect detection |
| 11 | F. Wang [53] | 2019 | CNN, Region Proposal Strategy, 91.7% Acc | Cashmere/Wool classification |
| 12 | McQuillan [565] | 2020 | Digital 2D/3D design, 3D software | Zero-waste fashion design |

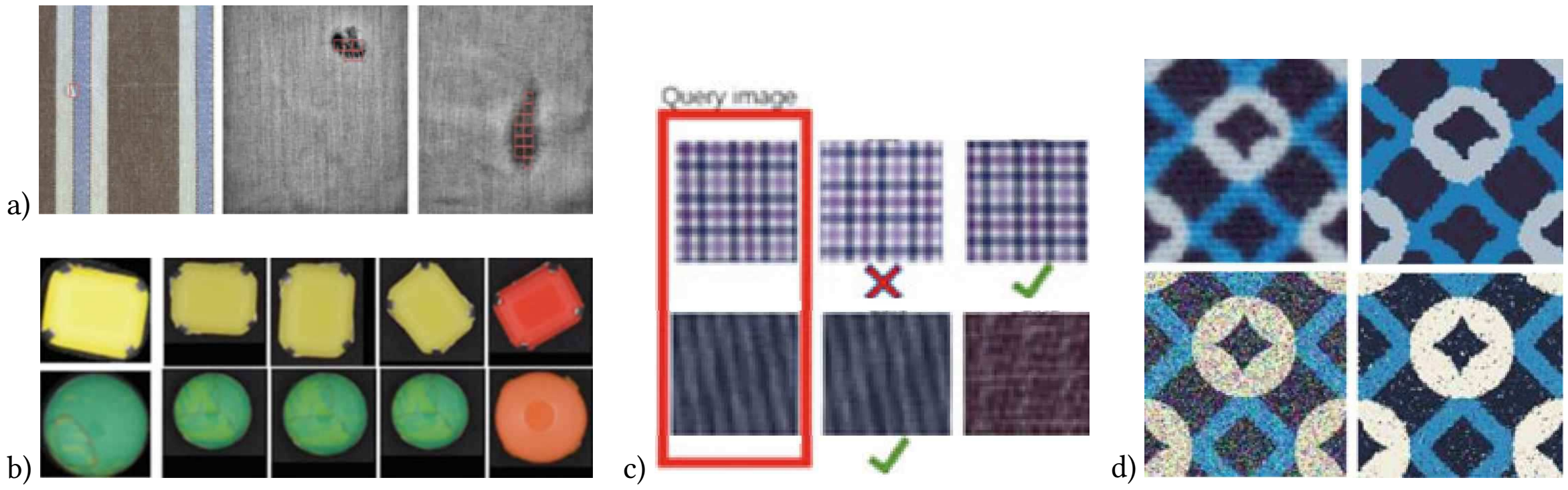

Figure 73: a) Fabric defect detection [561] b) Material image retrieval [377] c) Fabric identification [376] d) Fabric image segmentation on original (top row) and noisy image (bottom row) [563]

### 2.10 Miscellaneous

Here we list some inspiring fashion-related applications of ML and AI, including fashion captioning (natural language description, comment, or feedback generation on fashion images), apparel sorting (using humanoids or robots to sort, fold/unfold, and handle clothing articles automatically), and other miscellaneous applications that are out of the domain for our other categories. These articles are listed in Table 22.

Table 22: Miscellaneous Articles

| No | Article Reference | Year | Technical Keywords/Claimed Results | Application Notes |
|---|---|---|---|---|
| 1 | Kita [566] | 2010 | Classic, Visual recognition in cooperation with actions | Clothes sorting |
| 2 | Kita [567] | 2011 | Classic, Recognition by strategic observation | Clothes sorting |
| 3 | Bourdev [81] | 2011 | Classic, Classification, Poselet, HOG, Linear SVM | Fashion image captioning, Gender, Wild |
| 4 | Song [568] | 2011 | Det., HOG, LBP, Lasso-based sparse coding, 52.01% Acc | Occupation recognition, Clothing, Context |
| 5 | Shao [569] | 2013 | Classic, HOG, SVM, NMS-like greedy search, 41.1% mAP | Occupation recognition, Clothing, Context |
| 6 | Doumanoglou [570] | 2014 | POMDP, Random decision Forest, Hough Forest | Clothes sorting, Unfolding, Grasp selection |
| 7 | Zhang [571] | 2014 | Part detections, Latent structured SVM, HOG | Human pose detection using clothing Attr. |
| 8 | Sadeh [572] | 2019 | CNN, ResNet-18, SSD, MMI, RNN LM, LSTM, 0.56 BLEU4 | Natural language fashion image feedback |
| 9 | Lin [500] | 2020 | NOR, Mutual attention, MLP, GRU, RNN, 37.21 BLEU | Outfit matching, Comment generation |
| 10 | Qian [573] | 2020 | Region segmentation, U-Net, 70% Grasp success | Clothes sorting, Unfolding, Grasp selection |
| 11 | Nguyen [574] | 2020 | Encoder-Decoder, CNN-RNN, LSTM, Attention | Fashion image captioning, Shop |
| 12 | Banerjee [575] | 2020 | ResNet101, Attention-based LSTM, 32% Acc | Fashion image captioning, Shop |
| 13 | Yang [576] | 2020 | RL, ResNet, LSTM, Attribute/Sentence semantic reward | Fashion image captioning, Shop |

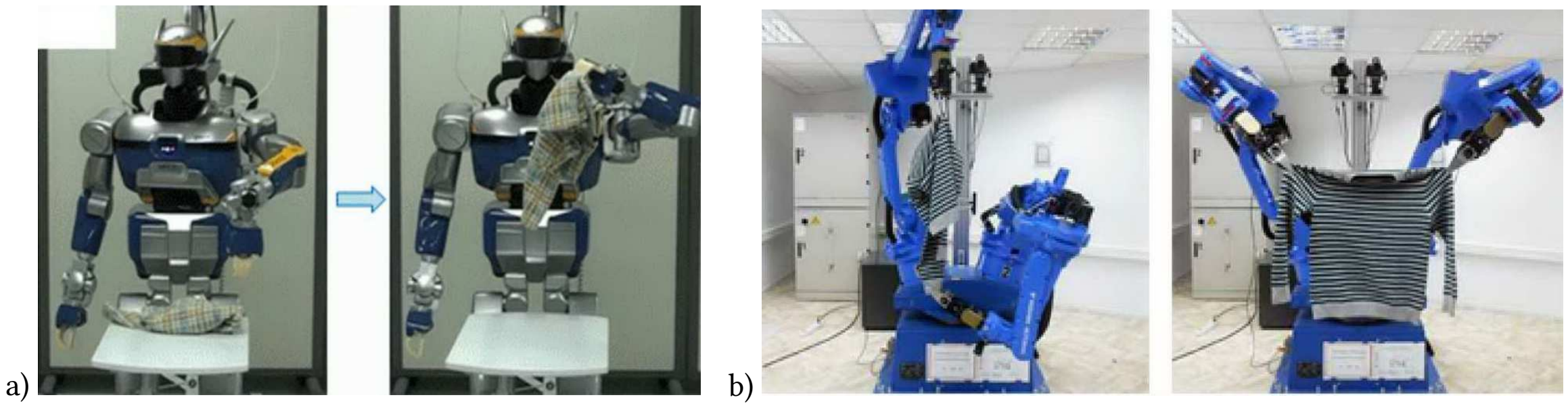

Figure 74: a) Clothes handling by a humanoid [567] b) Clothes unfolding by a robot [570]

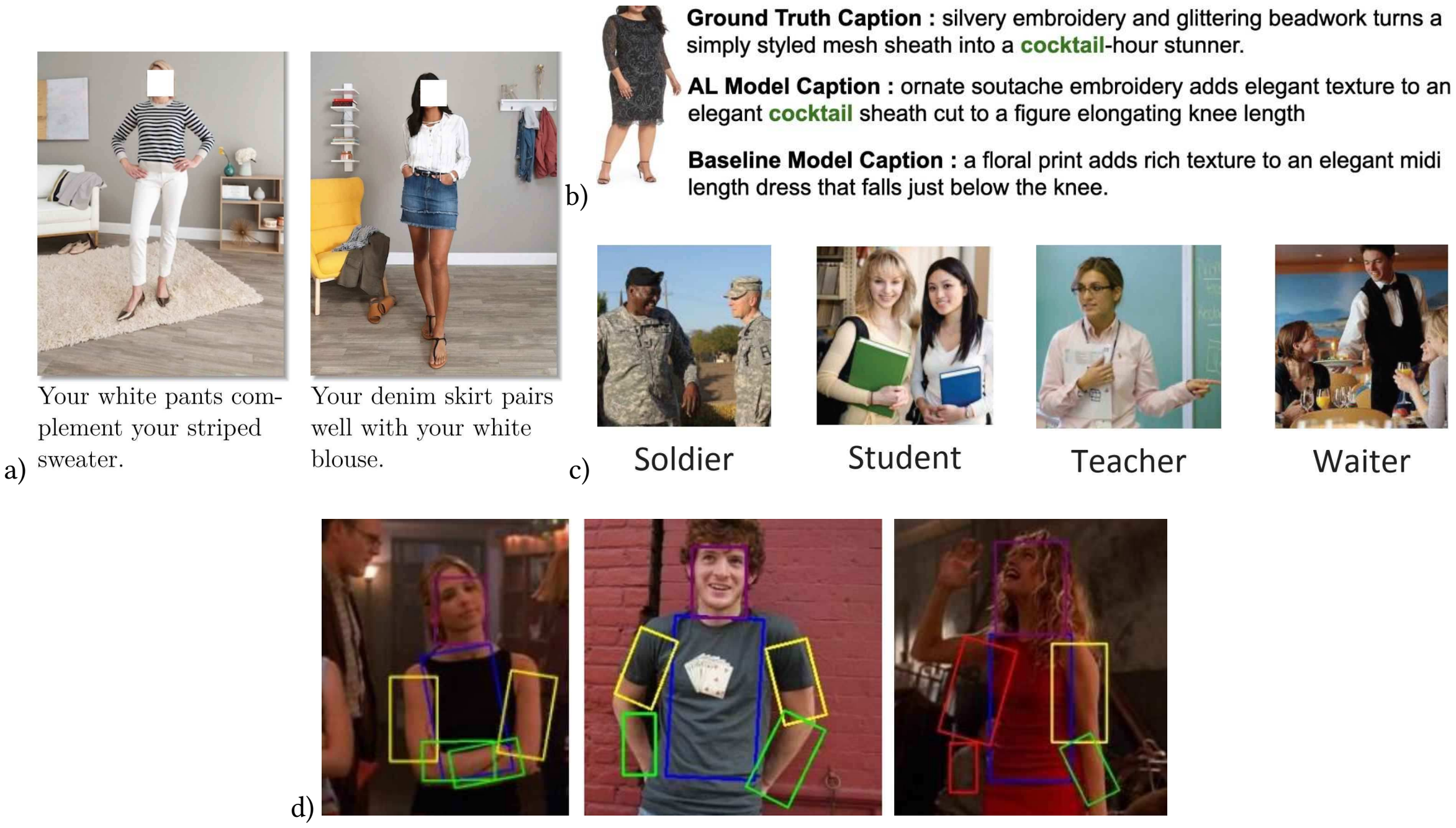

Figure 75: a) Natural language fashion image feedback [572] b) Fashion image captioning [575] c) Occupation recognition [569] d) Human pose estimation via clothing attributes [571]

## 3 DATASETS

As most fashion datasets are multi-task and can be used in various fashion applications based on their structure, we thought it would be misleading to report them in each section separately. Thus, we dedicate this section to the available fashion datasets. We report "suggested applications" for each dataset, meaning that the applications are not limited to these mentioned in Table 23; the primary application for each dataset comes first, then the rest follow. Although all studies use some datasets, many of them will not publish the data. Even amongst those who promise to do so, numerous datasets never make it to the internet due to copyright or other issues. Thus, unlike former survey studies, we only report easily accessible and publicly released datasets because they significantly contribute to the field and help the researchers.

Table 23: List of Fashion Datasets. Name of applications including Categorization (C), Attribute recognition (A), Item detection (I), Parsing (P), Landmark detection (L), image-based Try-on (T), 2D Modeling (2D), 3D Modeling (3D), Size & Fit (SF), Magic Mirror (M), fashion Synthesis (S), Domain-specific Retrieval (DR), Cross-domain Retrieval (CR), Attribute manipulation Retrieval (AR), Single-item Recommender (SR), Outfit style Recommender (OR), Fashion Compatibility (FC), Personalized Recommender (PR), and fashion Analysis & Trend (AT) are abbreviated.

| No | Dataset | Year | #Images | #Category | #Attributes | Type | Suggested Applications |
|---|---|---|---|---|---|---|---|
| 1 | Clothing Attributes [82] | 2012 | 1,856 | 7 | 26 | Street | C, A, DR |
| | | | Annotated with 23 binary-class attributes and 3 multi-class attributes. | | | | |
| 2 | Fashionista [137] | 2012 | 158,235 | 56 | - | Street | P, Pose, C |
| | | | 685 fully parsed images, Pose, tags, comments, links, and Person-tag, Chictopia. | | | | |
| 3 | Apparel Classification with Style [85] | 2013 | 80,000 | 15 | 78 | Street | C, A, DR |
| | | | Upper body, 8 Classes: Color, Pattern, Material, Structure, Look, Person, Sleeve | | | | |
| 4 | Colorful Fashion Parsing Data (CFPD) [141] | 2013 | 2,682 | 23 | - | Street | P, C |
| | | | Pixel-level 13 colors/23 classes labels | | | | |
| 5 | PaperDoll [139] | 2013 | 339,797 | 56 | Tags | Street | P, C, A |
| | | | Over 1 million pictures from chictopia, Color, Style, Occasion, Type, Brand | | | | |
| 6 | Fashion-Focused Creative Commons Social Dataset [577] | 2013 | 4,810 | 154 | 11,691 | Mixed | C, A |
| | | | Also general images, Tags (17.9 per image), Comments, Favorites, Contexts, Notes | | | | |
| 7 | Human3.6M [266] | 2013 | 3.6M | 17* | - | 3D poses/Image | 3D, 2D, Pose |
| | | | *Scenarios, 32 Joints, 11 actors, Pixel-level 24 body parts, Person bounding box | | | | |
| 8 | Fashion 10000 [578] | 2014 | 32,398 | 262 | 56,275* | Wild | DR |
| | | | *Tags, Geotag, Comment, Note, Favorite, Context | | | | |
| 9 | Clothing Co-parsing (CCP) [140] | 2014 | 2,098 | 57* | - | Street | P, I, C |
| | | | *1000 images with super-pixel tags, others with image-level tags, High-resolution | | | | |
| 10 | HipsterWars (Style) [20] | 2014 | 1,893 | 5* | - | Street | AT, Style classification |
| | | | *Styles: Bohemian, Goth, Hipster, Pinup, or Preppy. Style ratings | | | | |
| 11 | Chictopia [530] | 2014 | 328,604 | Tags | Tags | Street | A, C, DR, OR, AT |
| | | | 34,327 users, Popularity indicators: Votes, Comments, Bookmarks | | | | |
| 12 | UT-Zap50K [409] | 2014 | 50,025 | 4* | 4 | Item | Comparison tasks, A, DR |
| | | | *Only shoes in 4 categories, Metadata, 4k image pairs with +350 relative attributes | | | | |
| 13 | Aesthetics Based on Fashion Images [318] | 2014 | 1,064 | 11 | 4* | Model | SF, Body shape, A |
| | | | *Body shapes, 120 configurations (body shape with specific top/bottom categories) | | | | |
| 14 | Dual Attribute-aware Ranking Network (DARN) [86] | 2015 | 453,983* | 9 | 179 | Wild/Shop | CR, A |
| | | | *Now 214,619 excluding dead links, 91,390 image pairs, ~800 × 500 | | | | |
| 15 | Exact Street2Shop (WTBI) [389] | 2015 | 425,040* | 11 | Tags | Street/Shop | CR, SR, C, I |
| | | | *20,357 street+404,683 shop, 39,479 exact street2shop matches, Bboxes | | | | |

| No | Dataset | Year | #Images | #Category | #Attributes | Type | Suggested Applications |
|---|---|---|---|---|---|---|---|
| 16 | Amazon 2014 [8] | 2015 | 773,465 | Many | - | Shop | C, A, DR, SR, OR, PR, AT |
| | | | Clothing, Shoes, Jewelry, User-Item relations, Rating, Reviews, Geotags, K-cores | | | | |
| 17 | Fashion144K [457] | 2015 | 144,169 | Tags | Tags | Street | C, Style, A, AT |
| | | | Worldwide user posts containing diverse images, textual, and metadata + Geo-tags | | | | |
| 18 | Runway To Realway [23] | 2015 | 348,598 | Tags | Tags | Runway | C, Brand, A, AT |
| | | | Season, Category, 852 Brands, Date, Description | | | | |
| 19 | HumanParsing (ATR)/Chictopia10k [145] [146] | 2015 | 10,000 | 18* | - | Wild | P, C |
| | | | *12 clothing+background+5 features parsing labels, Frontal standing view | | | | |
| 20 | Deepfashion (DF) [9] | 2016 | 800,000 | 50 | 1,000 | Wild/Shop | A, C, P, L, T, 2D, S, DR, CR, SR |
| | | | 300K cross-pose/cross-domain pairs, 78,979 for Try-on, 4~8 landmarks | | | | |
| 21 | Multi-View Clothing (MVC) [91] | 2016 | 161,638 | * | 264* | Model | DR, T, C, A, S, SR |
| | | | *Hierarchical (Gender, Category, Attr.), Multiview (4+), 37,499 items, High-Res. | | | | |
| 22 | DeepFashion Alignment [167] | 2016 | 123,016 | 8* | - | Wild | L |
| | | | *Landmarks, Annotated with Clothing type, Pose, Visibility, Bbox, and Joints | | | | |
| 23 | Fashion144K (StyleNet) [14] | 2016 | 89,502 | Tags | Tags | Street | C, AT |
| | | | Built on Fashion 144k, Images centered, Bad images removed, Text, Tags, Votes | | | | |
| 24 | Sketch Me That Shoe [392] | 2016 | 419* | 2 | 21 | Item | S, CR, A |
| | | | *Sketch-photo pairs, Shoes with fine-grained triplet ranking annotations | | | | |
| 25 | LookBook [337] | 2016 | 84,748 | - | - | Street/Item | S, CR, T, Domain transfer |
| | | | Upper body, Item image+Models wearing that item, 9,732 Items-75,016 Models | | | | |
| 26 | WIDER Attribute [579] | 2016 | 13,789 | 30* | 14† | Street | A, C (Event), I |
| | | | *Event/Scene class, †Human attributes, 57524 Human Bbox, Not fashion specific | | | | |
| 27 | StreetStyle-27K [538] | 2017 | 27,000* | - | 12 | Wild | AT, A |
| | | | *All labeled+14.5 million unlabeled, From around the world, Geotag | | | | |
| 28 | Fashion550K [34] | 2017 | 1,061,468* | Tags | 3,627† | Street | C, Style, A |
| | | | *Noisy, †Tags, 550,661 posts, Extension on Fashion144K and StyleNet, 5,300 cleaned | | | | |
| 29 | FashionGAN [184] | 2017 | 78,979 | 50 | 1,000 | Model | T, S, P |
| | | | A subset of DeepFashion attribute enriched with sentence captions and Seg. Maps | | | | |
| 30 | Maryland Polyvore [464] | 2017 | 164,379* | 381 | - | Item | FC, OR, AT, C |
| | | | *Items forming 21,889 full outfits (max 8 items in each), Name, Price, Likes | | | | |
| 31 | Fashion-MNIST [580] | 2017 | 70,000 | 10 | - | Item, Grayscale | C |
| | | | 28x28 grayscale images from Zalando | | | | |
| 32 | UT-Zappos50K Synthetic Shoes [341] | 2017 | 4,000 | 1 | 10* | Item | S, A |
| | | | Only shoes, *Relational attribute pairs, ~2,000 pair labels per attribute | | | | |
| 33 | Bodies Under Flowing Fashion (BUFF) [272] | 2017 | 6,000* | 10~20† | - | 4D | 3D |
| | | | *Models, †Garments, 5 subjects, 2 clothing styles, 3 motions, Real, RGB, 0.4cm Res. | | | | |
| 34 | Fashion Semantic Space (FSS) [540] | 2017 | 32,133 | 2* | Many | Street | AT, Style, A, OR |
| | | | *Top/Bottom, Full-body fashion show images annotated with visual + style features | | | | |
| 35 | FashionStyle14 [37] | 2017 | 13,126 | 14* | - | Street | C (Style), OR |
| | | | *Japanese fashion style classes, No additional Info. | | | | |
| 36 | PASCAL-Person-Part [149] | 2017 | 3,533 | 14* | - | Wild | P, Pose |
| | | | *Human joints, Multiple humans per image, Unconstrained poses, Occlusions | | | | |
| 37 | Multiple Human Parsing (MHP) V1 [150] | 2017 | 4,980 | 18* | - | Wild | P (Multi-person) |
| | | | *7 Body parts+11 Clothing, Multiple persons (at least two, three on average) | | | | |
| 38 | Extended Chictopia [342] | 2017 | 14,411 | 14* | - | Street, 3D | P, Pose, 3D |
| | | | *Human joints, Chictopia10k+face annotations, pose and shape by 3D SMPL model | | | | |
| 39 | Fashion200K [413] | 2017 | 209,544* | 5 | 4,404 | Shop | A, C |
| | | | *Cleaned (All 300K are also available), Includes product descriptions, Lyst.com | | | | |

| No | Dataset | Year | #Images | #Category | #Attributes | Type | Suggested Applications |
|---|---|---|---|---|---|---|---|
| 40 | Learning the Latent "Look" [99] | 2017 | 18,878 | - | 195 | Street | AT, Style, A |
| | | | 70 to 600 positive images per attribute. Also, 2000 negative examples | | | | |
| 41 | Fashion Conversation Data On Instagram [541] | 2017 | 24,752* | - | - | Street | AT (Social), PR |
| | | | *Instagram posts by 13,350 people, Text, Hashtags, Account data | | | | |
| 42 | Street Fashion Style (SFS) [38] | 2017 | 293,105 | Tags | Tags | Street | AT, C, A |
| | | | Chictopia user posts, Season, Occasion, Style, Category, Color, Brand, Geotag, Year | | | | |
| 43 | FashionGen (FG) [345] | 2018 | 293,008 | 169* | - | Shop | S, C, A, DR, Captioning |
| | | | *48 Main categories+121 fine-grained, Multi-View, Expert description, 1360x1360 | | | | |
| 44 | ModaNet [104] | 2018 | 55,176 | 13 | - | Street | P, I, C, A |
| | | | Extension of Paperdoll, Includes bounding box and polygons, 119K masks | | | | |
| 45 | Polyvore Outfits [446] | 2018 | 365,054* | 19 | - | Item | FC, C, A |
| | | | *Items forming 68,306 full outfits, Title, Description, Category, Type | | | | |
| 46 | VITON (Zalando) [185] | 2018 | 32,506* | - | - | Item/Model | T, S, CR |
| | | | 16,253 frontal-view woman and top clothing item-model image pairs | | | | |
| 47 | Style4BodyShape [320] | 2018 | 347,948 | 5 | - | Female celebrities | SF, AT, OR, C |
| | | | 270 most stylish names, 3,150 body measurements, ~260 Pictures per Celebrity | | | | |
| 48 | Shared Shape Space For Multimodal Garment Design [275] | 2018 | 2,000 | 3* | - | 3D/2D Sketch | 3D, SF |
| | | | Synthesized by simulation, *Also various sizes, RGB, 1cm Res. | | | | |
| 49 | People-Snapshot [277] | 2018 | 264* | - | - | Video Sequence | 3D |
| | | | *24 sequences of 11 subjects varying in height and weight | | | | |
| 50 | 3DPW [581] | 2018 | 60* | - | - | Video/Pose/3D | 3D, 2D, Pose |
| | | | *Video sequences, 18 3D models with different clothing, 2.5cm Res., Pose | | | | |
| 51 | Crowd Instance-level Human Parsing (CIHP) [152] | 2018 | 38,280 | 20 | - | Wild | P (Multi-person) |
| | | | About 3.4 persons per image | | | | |
| 52 | Video Instance-level Parsing (VIP) [153] | 2018 | 404* | 19 | - | Video | P (Video, Multi-person) |
| | | | *Multiple-person videos, each 10-120 seconds, 20k frames, ~2.93 Person per frame | | | | |
| 53 | Look into Person (LIP) [154] | 2018 | 50,462 | 20 | - | Wild | P (Single-person), Pose |
| | | | Images of humans, 16 body joints, Complex poses, Occlusions, Back-view, etc. | | | | |
| 54 | Multiple Human Parsing (MHP) V2 [155] | 2018 | 25,403 | 58* | - | Wild | P (Multi-person) |
| | | | *11 Body parts+47 Clothing, Multiple persons (at least two, three in average) | | | | |
| 55 | Personalized Outfit Generation (POG) [481] | 2019 | 583,464* | 80 | - | Item | FC, PR, C |
| | | | *Items forming 1,013,136 full outfits, Context, 3,569,112 users with behavior Info. | | | | |
| 56 | Atlas [57] | 2019 | 183,996 | 52 | - | Model/Zoomed | C |
| | | | Image title, Price, 3-level category taxonomy. Normal+Zoomed photos | | | | |
| 57 | Deepfashion2 (DF2) [129] | 2019 | 491,000 | 13 | - | Wild/Street | I, P, C, L, Pose, CR, SR |
| | | | 801K Items/Bboxes/Landmarks/Masks, 873K Cross pairs, 13 Poses | | | | |
| 58 | FashionAI (FAI) [111] | 2019 | 357,000 | 6* | 245 | Mixed | C, A, L |
| | | | Women's clothing, *41 Sub-categories, Hierarchical, 24 Landmarks | | | | |
| 59 | Fashionpedia [582] | 2019 | 48,827 | 46 | 294 | Wild | P, I, C, A |
| | | | Ontology, High resolution with 1710 × 2151, Localized attribute annotation | | | | |
| 60 | iMaterialist [58] | 2019 | 1,012,947 | 105 | 228* | Shop | A, C |
| | | | *In 8 groups, Multi-labeled, Fine-grained | | | | |
| 61 | Shop The Look (STL-Fashion) [486] | 2019 | 93,274* | 10 | - | Street/Item | CR, SR, C, I, FC, Scene |
| | | | *Scene-product pairs from 38,111 products and 47,739 scenes, Bbox, Style, Scene | | | | |
| 62 | Multi-Garment Network (MGN) [278] | 2019 | 356* | 5 | - | 3D garments | 3D |
| | | | *Pairs of image-digital garments, Real, Scan, Vertex Color, 3D body pose | | | | |
| 63 | GarNet [279] | 2019 | 600* | 3 | - | 3D | 3D |
| | | | *Models, Synthetic, Simulation, 40/23/31 poses for T-shirt/sweater/jeans | | | | |
| 64 | 3DPeople [280] | 2019 | 2M* | - | - | 3D | 3D |
| | | | *Photorealistic frames, 80 subjects, 70 actions, Body shapes, Skin tones, Clothing | | | | |

| No | Dataset | Year | #Images | #Category | #Attributes | Type | Suggested Applications |
|---|---|---|---|---|---|---|---|
| 65 | Amazon 2018 [583] | 2019 | 2,685,059 | Many | - | Shop | PR, C, A, DR, SR, OR, AT |
| | | | Clothing, Shoes, Jewelry, User-Item relations, Rating, 32M Reviews, Geotags, K-cores | | | | |
| 66 | Fashion IQ [402] | 2019 | 77,684 | 3* | 1,000† | Shop | CR, Captioning, S, AR |
| | | | Images with caption, *(Dress, Shirt, Top&Tee), †In 5 groups | | | | |
| 67 | MPV [238] | 2019 | 62,780* | - | - | Model/Item | 2D, T, CR, S |
| | | | *Triplets of two model images in different poses+corresponding item image | | | | |
| 68 | Fashion Culture DataBase (FCDB) [549] | 2019 | 25,707,690 | 16* | - | Wild | AT, Temporal trends |
| | | | World people, *Geo-tags (Cities), Person-tag, Person Bboxes, Time-stamp | | | | |
| 69 | FashionTryOn [239] | 2019 | 28,714* | - | - | Shop | 2D, T, CR, S |
| | | | *Triplets of two model images in different poses+corresponding item image | | | | |
| 70 | THUman [584] | 2019 | 7,000* | - | - | 3D | 3D |
| | | | *Real-world human textured surface mesh, Clothing, Shapes, Poses, ~28K images | | | | |
| 71 | Impersonator (iPER) [207] | 2019 | 206* | - | - | Video | 2D, Motion/Style Transfer |
| | | | *Video sequences, 241,564 frames, 30 subjects (Height, Shape, Gender), 103 clothes | | | | |
| 72 | Polyvore-T [497] | 2019 | 19,835* | 5 | - | Item | FC, OR, AT, C |
| | | | Cleaned Polyvore, *Outfits (3-8 items each), Name, Price, Likes | | | | |
| 73 | IQON3000 [498] | 2019 | 672,335* | 6 | Yes | Item | FC, A, PR, C, AT |
| | | | *Items forming 308,747 full outfits, 3,568 users, Attributes, Description, Price, Likes | | | | |
| 74 | BodyFashion [324] | 2019 | 116,532* | Yes | - | Shop | FC, PR, SF |
| | | | *User-item purchase records+Body shape+Size, Rating, 75,695 items, 11,784 users | | | | |
| 75 | Kaggle, Fashion Product Images [585] | 2019 | 44,000 | 52* | 199† | Shop | C, A, SR |
| | | | *7/45 master/sub, †143 Type+47 Color+9 Usage tags, Season, Year, 1800x2400 | | | | |
| 76 | Kaggle, Nitin Singh Fashion [586] | 2019 | 15,703 | 17* | - | Street | C (Dress type), I, SR |
| | | | Only Dress, *Dress types+Confidence level, Bboxes drawn on images | | | | |
| 77 | Deep Fashion3D [292] | 2020 | 2,078* | 10 | - | 3D | 3D |
| | | | *Models, Real, Multi-view stereo, 3D body pose & feature lines | | | | |
| 78 | Long-Term Cloth-Changing Person (LTCC) [587] | 2020 | 17,138 | - | - | Wild | DR |
| | | | Re-ID, 152 identities with 478 outfits, Person with same/different outfits & angles | | | | |
| 79 | SIZER [297] | 2020 | 2,000* | 10 | - | 3D | 3D, SF |
| | | | *Scans of 100 subjects, Same garment different sizes, Seg., SMPL+G, Body shape | | | | |
| 80 | TailorNet [298] | 2020 | 55,800* | 20† | - | 3D | 3D |
| | | | *Frames, †Aligned real static garments, 1782 poses, 9 body shapes, 1cm Res., RGB | | | | |
| 81 | Fashion32 [516] | 2020 | 40,667* | 32† | 152 | Item/Model | C, A, FC, CR, Theme-aware |
| | | | *Items, 51,415 models, 13,914 outfits, †Themes, Description, Style, Fit, Gender | | | | |
| 82 | VIBE [326] | 2020 | 1,957* | 2 | Tags | Item | SF, A |
| | | | *958 dresses+999 tops, Each front+back view, 68 models, Garment and Body sizes | | | | |
| 83 | Fashion Captioning Dataset (FACAD) [576] | 2020 | 993,000 | 78 | 990 | Shop | Captioning, C, A |
| | | | 130K descriptions, Each item 6~7 images, Colors, Poses, 1560×2392 | | | | |
| 84 | CAPE [303] | 2020 | 600* | 4 | | 3D | 3D |
| | | | *Motion sequences, 140K frames, 10 male+5 female models, 3D mesh scans, Pose | | | | |
| 85 | CLOTH3D [306] | 2020 | 8,000+* | 7 | - | 3D | 3D |
| | | | *Sequences, 2.1M frames of 11.3K 3D garments, Texture data, RGB, Pose, 1cm Res. | | | | |
| 86 | Attribute-Specific Embedding Network [388] | 2020 | 180,000+* | Yes | 8+ | Mixed | CR, DR, A |
| | | | Rebuilds DARN, FashionAI, and DeepFashion with attribute-specific annotations | | | | |

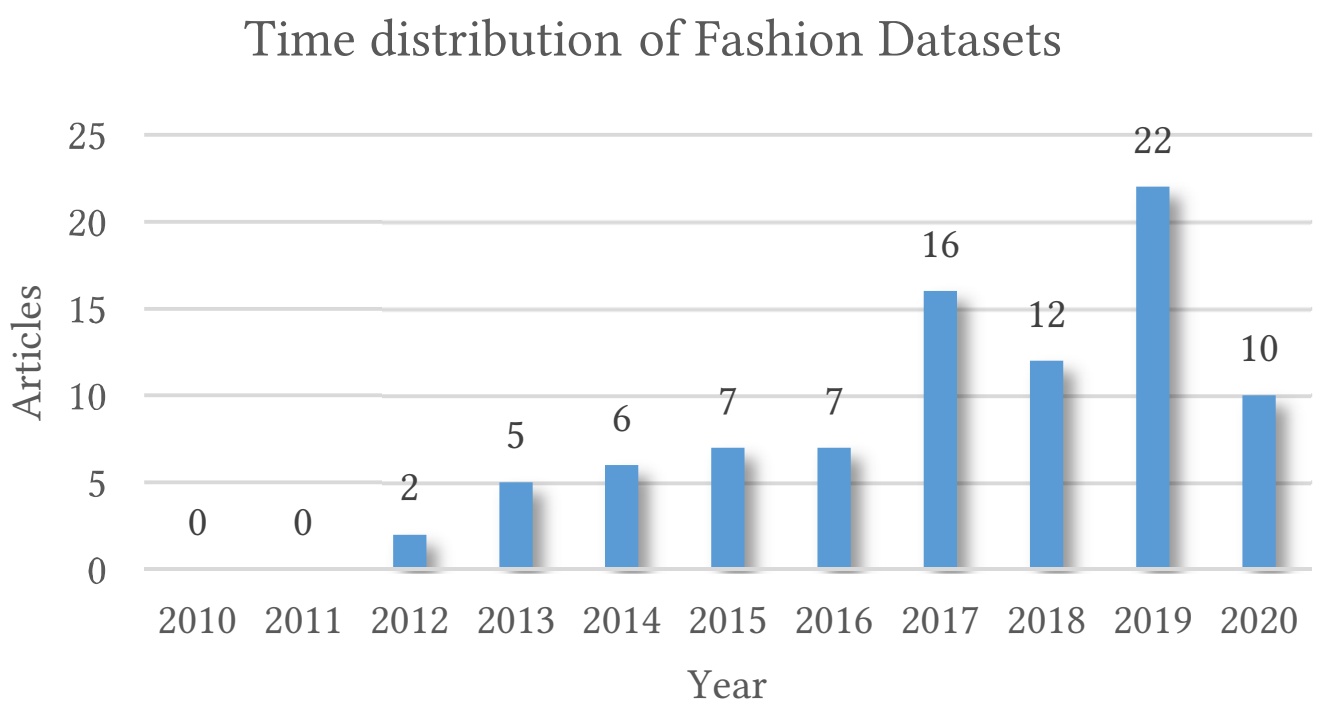

Figure 76: Time Analysis of Fashion Datasets

## 4 DISCUSSION AND FUTURE PATH

The implementation of computer vision and AI in the fashion industry is happening inevitably fast, but not fast enough. Although the past decade has witnessed a dramatic growth of research in this area (see Figure 77), the immense size of the area, including various applications and the increased need for online fashion retail shops throughout the world due to Covid-19 pandemic situations, show that still much work needs to be done.

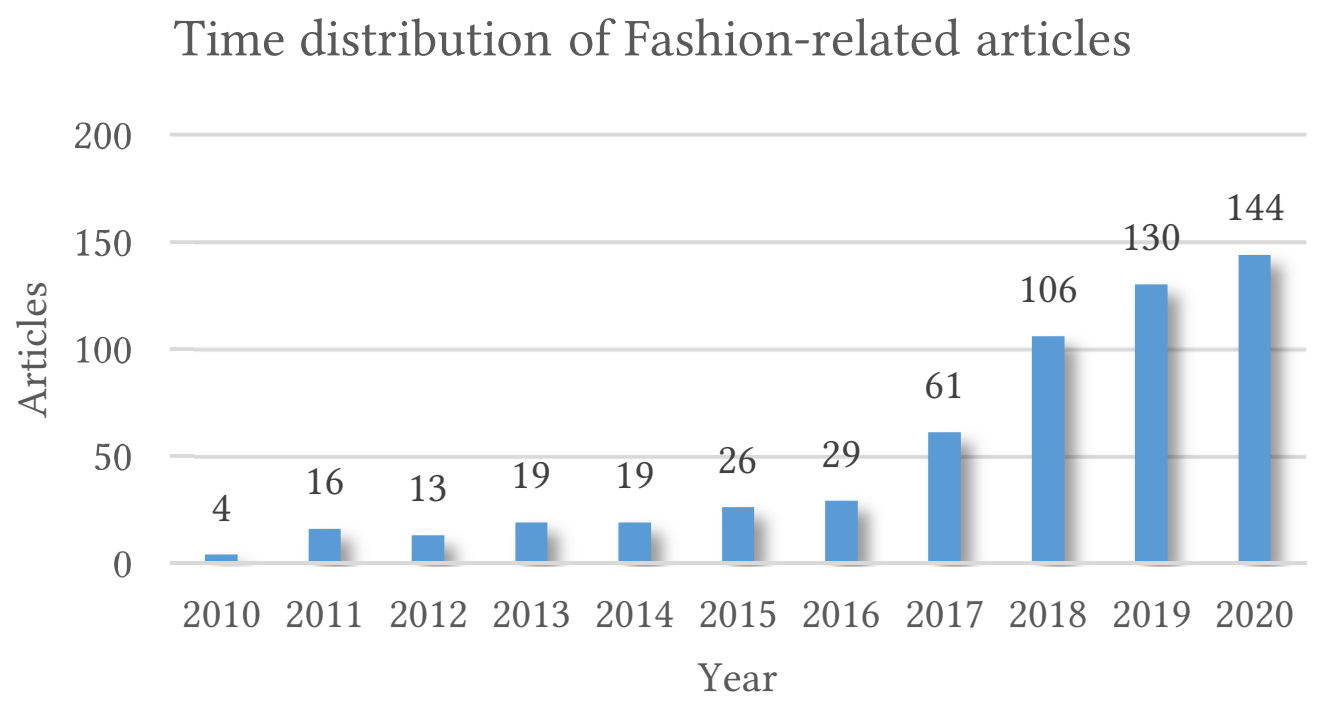

Figure 77: Time Analysis of Fashion-Related Articles Covered In This Study

A more thorough look at the fashion-related applications in Figure 78 helps us understand which areas need more attention. Needless to say, all of these fashion-related tasks (and many more we did not cover here) are incredibly useful in the fashion industry, and the proper implementation of each and every one of them can be highly profitable for companies. Therefore, Figure 78 is just a means to track which tasks are already hot topics, today's market needs and fast-growing, and which are neglected, thus have fantastic potential and are very promising in the coming years. We also did a keyword analysis on all the articles mentioned in this survey and the result, in Figure 79, is fascinating. We can see that the frequency of tags does not entirely reflect the frequency of tasks, as in Figure 78. Part of it is because each research article might contribute to more than one of these tasks or even study a higher-level task composed of several low-level tasks. It is clear that researchers need something more accurate than conventional keyword-based search engines to access the right resources, hence the need for this survey. We also present Table 24, a co-occurrence of different fashion-related tasks in articles, hoping to shine some light on the relationship between various tasks and how often they were analyzed simultaneously in different research articles. Each cell in the table is the rounded percentage of intersection over union, showing what ratio of articles related to every two tasks study these tasks simultaneously.

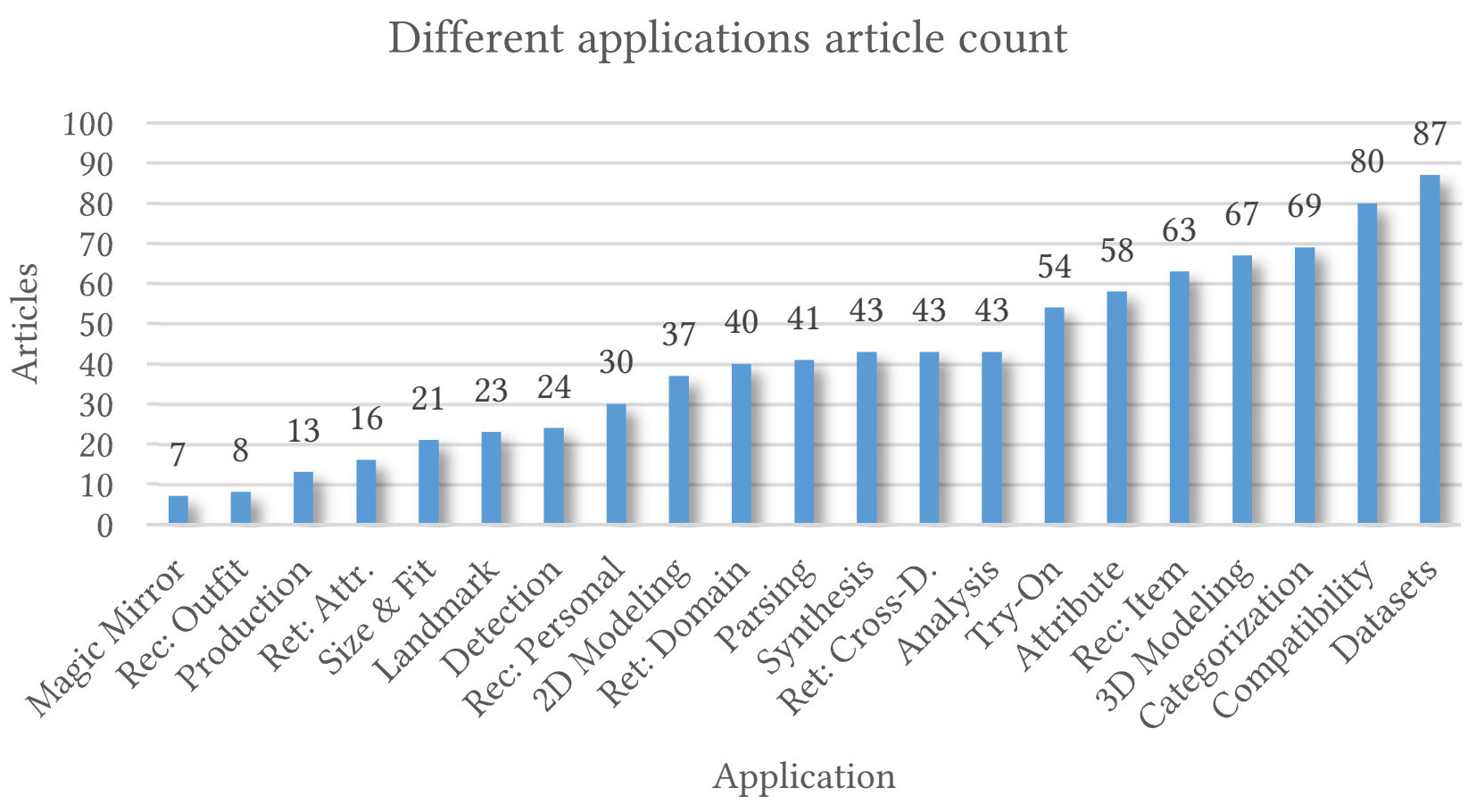

Figure 78: Article count based on different fashion tasks (One article may contribute to multiple applications)

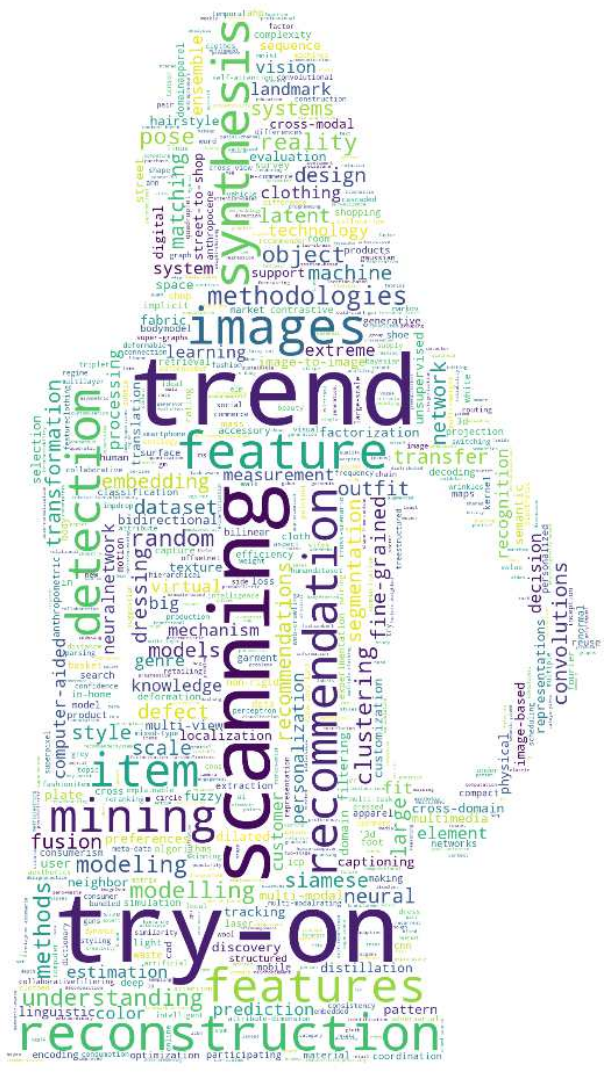

Figure 79: Frequency-based fashion keywords tag cloud

There are still multiple challenges along the way. One main challenge is the lack of a clean, large-scale fashion dataset from different sources. Fortunately, with the massive amount of data at hand and various ever-growing social media networks, the lack of data is no longer a problem. What we need is a good-enough annotation scheme to leverage this data. Many works in this area use small datasets tailored for their own needs, and even many of these datasets are never published. Although we introduce 86 different public datasets in this survey, it should be noted that almost none of them is a unified, universal fashion dataset. They are either small, task-sensitive or from a single or very few sources. Larger datasets are usually for general tasks, and more specific datasets are usually very small; thus, it would be fantastic to have it all in one dataset. It can actually be tough to find suitable and uniformly labeled datasets for some specific tasks.

Another problem is the lack of assessment techniques for some specific fashion tasks (e.g., recommendation, synthesis, and compatibility). It is hard to define objective metrics to reflect many notions in fashion like beauty, novelty, compatibility, fashionability, and many more. As a result, many tasks still use subjective assessments, which can be inaccurate and biased. Although one might introduce a metric that works in the same direction (e.g., use co-purchase as a sign of compatibility), the definition of a well-structured objective metric for many tasks is still an unsolved problem.

"Is smart fashion ready yet?" is the final question we need to answer. The performance of such systems is still of concern to fashion companies as many of these tasks still cannot compete with trained human assessors. Nevertheless, this should not stop them from using these technologies. Based on the remarkable improvements we have witnessed in such a short period, it will not take long before seeing smart fashion at its peak. Many researchers worldwide are contributing to the field to improve not only the performance of such systems but also the computational efficiency and cost-effectiveness of them as these features play an essential role in the usability of such systems and implementation on mobile phones and other smart devices.

Table 24: Co-Occurrence Table Of Fashion-Related Applications (Rounded Percentage Of Intersection Over Union)

| Total Count | Task | Categorization | Attribute | Detection | Parsing | Landmark | Try-On | 2D Modeling | 3D Modeling | Size & Fit | Magic Mirror | Synthesis | Ret: Domain | Ret: Cross-D. | Ret: Attr. | Rec: Item | Rec: Outfit | Compatibility | Rec: Personal | Analysis | Production | Datasets |
|---|---|---|---|---|---|---|---|---|---|---|---|---|---|---|---|---|---|---|---|---|---|---|
| 69 | Categorization | 100% | | | | | | | | | | | | | | | | | | | | |
| 58 | Attribute | 13% | 100% | | | | | | | | | | | | | | | | | | | |
| 24 | Detection | 5% | 11% | 100% | | | | | | | | | | | | | | | | | | |
| 41 | Parsing | 5% | 4% | 6% | 100% | | | | | | | | | | | | | | | | | |
| 23 | Landmark | 7% | 7% | 11% | 5% | 100% | | | | | | | | | | | | | | | | |
| 54 | Try-On | 0% | 0% | 0% | 1% | 5% | 100% | | | | | | | | | | | | | | | |
| 37 | 2D Modeling | 0% | 0% | 0% | 3% | 0% | 9% | 100% | | | | | | | | | | | | | | |
| 67 | 3D Modeling | 0% | 1% | 1% | 1% | 1% | 4% | 1% | 100% | | | | | | | | | | | | | |
| 21 | Size & Fit | 0% | 0% | 0% | 0% | 0% | 0% | 0% | 6% | 100% | | | | | | | | | | | | |
| 7 | Magic Mirror | 1% | 0% | 0% | 0% | 0% | 0% | 0% | 3% | 0% | 100% | | | | | | | | | | | |
| 43 | Synthesis | 1% | 1% | 1% | 1% | 0% | 4% | 4% | 1% | 2% | 0% | 100% | | | | | | | | | | |
| 40 | Ret: Domain | 7% | 8% | 5% | 4% | 3% | 0% | 0% | 0% | 0% | 0% | 1% | 100% | | | | | | | | | |
| 43 | Ret: Cross-D. | 6% | 6% | 7% | 5% | 6% | 0% | 0% | 0% | 0% | 0% | 2% | 16% | 100% | | | | | | | | |
| 16 | Ret: Attr. | 1% | 1% | 0% | 0% | 0% | 0% | 0% | 0% | 0% | 0% | 0% | 4% | 3% | 100% | | | | | | | |
| 63 | Rec: Item | 8% | 6% | 6% | 3% | 0% | 0% | 0% | 1% | 2% | 3% | 1% | 3% | 3% | 1% | 100% | | | | | | |
| 8 | Rec: Outfit | 1% | 5% | 3% | 0% | 0% | 0% | 0% | 0% | 0% | 0% | 0% | 0% | 0% | 0% | 1% | 100% | | | | | |
| 80 | Compatibility | 2% | 3% | 5% | 2% | 1% | 1% | 0% | 0% | 1% | 0% | 5% | 3% | 2% | 1% | 5% | 1% | 100% | | | | |
| 30 | Rec: Personal | 0% | 1% | 0% | 0% | 0% | 0% | 0% | 0% | 2% | 0% | 4% | 0% | 0% | 0% | 9% | 5% | 13% | 100% | | | |
| 43 | Analysis | 4% | 4% | 0% | 2% | 0% | 0% | 0% | 0% | 0% | 2% | 0% | 1% | 1% | 0% | 2% | 2% | 0% | 1% | 100% | | |
| 13 | Production | 1% | 0% | 0% | 0% | 0% | 0% | 0% | 0% | 0% | 0% | 0% | 4% | 0% | 0% | 0% | 0% | 0% | 0% | 0% | 100% | |
| 87 | Datasets | 6% | 6% | 3% | 12% | 3% | 2% | 2% | 8% | 5% | 0% | 4% | 4% | 7% | 3% | 3% | 2% | 5% | 3% | 8% | 0% | 100% |

## 5 CONCLUSION

All the research through the years led to the birth of these fantastic smart fashion technologies, and they still have a long way to fulfill their true potential. Leading fashion industry companies are beginning to see the many advantages of intelligent fashion and are focusing their attention on this research area; thus, the field is now so vast that a mere customary keyword search might not be enough to access related research articles. This fact highlights the importance of this unified fashion-related task-based survey to draw new researchers' attention to the subject and point them towards correct research directions and sources. This field is becoming enormous, we categorized more than 580 articles into multiple task-based groups, and there are still many more. The observed trends and growth speed guarantees that we will soon witness numerous significant improvements that close the human-machine gap.